\documentclass{article}

\usepackage[nonatbib,preprint]{neurips_data_2022}
\usepackage{geometry}
\usepackage{graphicx}
\usepackage{color}
\usepackage{amssymb}
\usepackage{tabularx}
\usepackage[dvipsnames,table]{xcolor}

\usepackage{url}
\usepackage{empheq}
\usepackage[utf8]{inputenc} % allow utf-8 input
\usepackage[T1]{fontenc}    % use 8-bit T1 fonts
\usepackage{hyperref}       % hyperlinks
\usepackage{url}            % simple URL typesetting
\usepackage{booktabs}       % professional-quality tables
\usepackage{amsfonts}       % blackboard math symbols
\usepackage{nicefrac}       % compact symbols for 1/2, etc.
\usepackage{microtype}      % microtypography
\usepackage{multirow}
\usepackage{multicol}
\usepackage{subfiles}
\usepackage{todonotes}

\usepackage{float}

\usepackage{hyperref}
\usepackage{url}

\usepackage[utf8]{inputenc} % allow utf-8 input
\usepackage[T1]{fontenc}    % use 8-bit T1 fonts
\usepackage{booktabs}       % professional-quality tables
\usepackage{amsfonts}       % blackboard math symbols
\usepackage{nicefrac}       % compact symbols for 1/2, etc.
\usepackage{microtype}      % microtypography

\usepackage{subfigure}
\usepackage{appendix}
\usepackage{mathtools}
\usepackage{subfiles}
\usepackage{times}
\usepackage{latexsym}
\usepackage{multirow}
\usepackage{empheq}
\usepackage{bm}
\usepackage{bbm}

\usepackage[normalem]{ulem}
\useunder{\uline}{\ul}{}
\usepackage{siunitx}    
\newcommand\blfootnote[1]{%
  \begingroup
  \renewcommand\thefootnote{}\footnote{#1}%
  \addtocounter{footnote}{-1}%
  \endgroup
}

\title{Shifts 2.0: Extending The Dataset of Real Distributional Shifts}

\author{Andrey Malinin \textsuperscript{1}  \and
Andreas Athanasopoulos \textsuperscript{3} \and 
Muhamed Barakovic \textsuperscript{4} \and 
Meritxell Bach Cuadra \textsuperscript{5} \and 
Mark J. F. Gales\textsuperscript{2} \and 
Cristina Granziera \textsuperscript{4} \and 
Mara Graziani \textsuperscript{6}  \and 
Nikolay Kartashev \textsuperscript{1} \and 
Konstantinos Kyriakopoulos \textsuperscript{3}   
\and Po-Jui Lu \textsuperscript{4} \and 
Nataliia Molchanova \textsuperscript{5,6} \and
Antonis Nikitakis \textsuperscript{3} \and 
Vatsal Raina \textsuperscript{2,6} \and 
Francesco La Rosa \textsuperscript{7} \and
Eli Sivena \textsuperscript{3} \and
Vasileios Tsarsitalidis \textsuperscript{3} \and 
Efi Tsompopoulou \textsuperscript{3}  \and 
Elena Volf \textsuperscript{1} \\
 \raggedright{{\tt \hspace{-300pt}\texttt{info@shifts.ai} }}
 }

\begin{document}

\maketitle

\blfootnote{$^1$ Shifts Project, $^2$ University of Cambridge, $^3$ DeepSea, $^4$ University of Basel,  $^5$ University of Lausanne, $^6$ University of Applied Sciences of Western Switzerland, $^7$ Icahn School of Medicine at Mount Sinai} %(HES-SO Valais)

\begin{abstract}

Distributional shift, or the mismatch between training and deployment data, is a significant obstacle to the usage of machine learning in high-stakes industrial applications, such as autonomous driving and medicine. This creates a need to be able to assess how robustly ML models generalize as well as the quality of their uncertainty estimates. Standard ML baseline datasets do not allow these properties to be assessed, as the training, validation and test data are often identically distributed. Recently, a range of dedicated benchmarks have appeared, featuring both distributionally matched and shifted data. Among these benchmarks, the Shifts dataset stands out in terms of the diversity of tasks as well as the data modalities it features. While most of the benchmarks are heavily dominated by 2D image classification tasks, Shifts contains tabular weather forecasting, machine translation, and vehicle motion prediction tasks. This enables the robustness properties of models to be assessed on a diverse set of industrial-scale tasks and either universal or directly applicable task-specific conclusions to be reached. In this paper, we extend the Shifts Dataset~\cite{malinin2021shifts} with two datasets sourced from industrial, high-risk applications of high societal importance. Specifically, we consider the tasks of segmentation of white matter Multiple Sclerosis lesions in 3D magnetic resonance brain images and the estimation of power consumption in marine cargo vessels. Both tasks feature ubiquitous distributional shifts and a strict safety requirement due to the high cost of errors. These new datasets will allow researchers to further explore robust generalization and uncertainty estimation in new situations. In this work, we provide a description of the dataset and baseline results for both tasks.
\end{abstract}
\section{Introduction}

In machine learning it is commonly assumed that training, validation, and test data are independent and identically distributed, implying that good testing performance is a strong predictor of model performance in deployment. This assumption seldom holds in real, "in the wild" applications. Real-world data are subject to a wide range of possible \emph{distributional shifts} -- mismatches between the training data and the test or deployment data \cite{Datasetshift,koh2020wilds,malinin2021shifts}. In general, the greater the degree of the shift in data, the poorer the model performance on it. The problem of distributional shift is relevant to the general machine learning community, as most ML practitioners have faced the issue of mismatched training and test data at some point. The issue is especially acute in high-risk industrial applications such as finance, medicine, and autonomous vehicles, where a mistake by the ML system may incur significant financial, reputational and/or human loss. 

Ideally, machine learning models should demonstrate robust generalization under a broad range of distributional shifts. However, it is impossible to be robust to \emph{all} forms of shifts due to the \emph{no free lunch theorem}~\cite{murphy}. ML models should therefore indicate when they fail to generalize via uncertainty estimates, which enables us to take actions to improve the safety and reliability of the ML system, deferring to human judgement~\cite{aisafety, malinin-thesis}, deploying active learning~\cite{active-learning,kirsch2019batchbald} or propagating the uncertainty through an ML pipeline~\cite{talarbel}. Unfortunately, standard machine learning benchmarks and evaluation setups, which contain i.i.d training, validation and test data, do not allow robustness to distributional shift and uncertainty quality to \emph{both} be assessed. Thus, there is an acute need for dedicated benchmarks and evaluation setups which are designed to assess both properties.

Until recently most work on uncertainty estimation\cite{malinin-endd-2019,malinin-thesis,Gal2016Dropout,deepensemble2017,baselinedetecting,ashukha2020pitfalls,trust-uncertainty} and robust generalisation has focused on small- and medium-scale image and text classification tasks, such as MNIST \cite{mnist}, SVHN \cite{svhn}, and CIFAR10/100 \cite{cifar}. More recent work has been evaluated on ImageNet and the associated A, R, C, and O versions of ImageNet~\cite{imagenet-r,imagenet-c,imagenet-ao}, which contain curated, but synthetic, distributional shifts. While a significant step forward, it is still limited, as the distributional shifts, the data modality, and the task are not representative of high-risk industrial applications. In the Natural Language Processing community, the Workshop on Machine Translation (WMT) holds a robustness track, where the goal is to evaluate the machine translation systems on data crawled from Reddit, which contains many examples of highly atypical usage of language~\cite{michel2018mtnt}. However, uncertainty estimation in not assessed in this benchmark.

Recently, however, the ML field has witnessed the appearance of dedicated benchmarks for assessing generalisation under distributional shift and uncertainty estimation. Specifically, the WILDS collection of datasets\cite{koh2020wilds,sagawa2021extending}, the Diabetic Retinopathy dataset~\cite{filos2019systematic}, the ML uncertainty benchmarks~\cite{nado2021uncertainty} and the Shifts dataset~\cite{malinin2021shifts}. WILDS is currently the most comprehensive dataset for the evaluation of domain generalisation - it is a collection of 10 datasets, of which 6 are image-based, 3 are text-based and 1 is molecule-based. WILDS considers a range of domain-generalisation tasks and features in-domain as well as shifted data. It was also extended~\cite{sagawa2021extending} with additional unsupervised data to enable investigation of domain adaptation. While the WILDS \emph{dataset} can be used to assess both uncertainty and robustness, the WILDS \emph{benchmark} has only examined robust domain generalisation and domain adaptation. Another limitation of WILDS is that it assumes access to a "domain label" at training and test time, which may not be a reasonable assumption in many real-world settings. In contrast, the Diabetic Retinopathy dataset and benchmark~\cite{filos2019systematic}, which is an image classification task, assess both model robustness and uncertainty estimation. Finally, the uncertainty-benchmarks~\cite{nado2021uncertainty} primarily assesses a range of uncertainty estimation techniques on the ImageNet suite of datasets. The Shifts Dataset~\cite{malinin2021shifts} is also a recent benchmark for jointly assessing the robustness of generalisation and uncertainty quality. Its principle difference to other benchmarks is that it contains large, industrially sourced data with examples of real distributional shifts, from three very different data modalities and four different predictive tasks - specifically a tabular weather forecasting task (classification and regression), a text-based translation task (discrete autoregressive prediction) and a vehicle motion-prediction task (continuous autoregressive prediction). It was constructed specifically to examine data modalities and predictive tasks which are not as well-studied as 2D image classification, which is the most represented in the other benchmarks described above. 

In this paper we extend the Shifts Dataset~\cite{malinin2021shifts} with two new datasets sourced from high-risk healthcare and industrial tasks of high societal importance. Specifically, 3D segmentation of white matter Multiple Sclerosis (MS) lesions in Magnetic Resonance Imaging (MRI) scans of the brain and the estimation of power consumption by marine cargo vessels. These two tasks constitute distinct examples of data modalities and predictive tasks that are still scarce in the field. The former represents a structured prediction task for 3D imaging data, which is novel to Shifts, and the latter a tabular regression task. Both tasks feature ubiquitous real-world distributional shifts and a strict requirement for robustness and reliability due to the high cost of erroneous predictions. For both datasets we assess ensemble-based baselines in terms of the robustness of generalisation and uncertainty quality.

\section{Benchmark Paradigm, Evaluation and Choice of Baselines}
\label{sec:paradigm}
% \begin{itemize}
%     \item bigger shifts higher error
%     \item use uncertainty as an evaluation of the risks
%     \item utility in downstream tasks
%     \item choice of baselines
% \end{itemize}

\paragraph{Paradigm} Similarly to the original Shifts paper~\cite{malinin2021shifts}, we view the problems of robustness and uncertainty estimation as having \emph{equal} importance. Models should be robust to as broad a range of distributional shifts as possible. However, through the \emph{no free lunch theorem}, we know that we can’t construct models which are guaranteed to be universally better on all shifted sub-distributions of a particular task than models natively trained on those sub-distributions. 
%The variability of the real world is so high that it is in impossible to uniformly cover the entire support of the “whole” distribution pertaining to a particular task. There will be regions (sub-distributions) which were either covered sparsely, or not at all, in the training data. For example, a self-driving car which has seen millions of hours of driving data may still be surprised and function poorly when it encounters a very rare “long-tail” event which happens once every million hours. Thus, even if a model is generally robust to a very broad set of shifts, it is conceivable that there will be a version of the task coming from a very distant and rare sub-distribution on which the 'broadly robust" model will perform worse than a natively trained model.
Thus, where models fail to robustly generalise, they should yield high estimates of uncertainty, enabling risk-mitigating actions to be taken (e.g., transferring control of a self-driving vehicle to a human operator). Thus, it is necessary to \emph{jointly} assess robustness and uncertainty estimation, in order to see whether uncertainty estimates at the level of a single prediction correlate well with the likelihood or degree of error. 

% However, we speculate that it should be possible to define a bounded set of shifts on which the model will yield performance comparable to that of natively trained models. Specifically, it should be possible to build in invariances and/or equivariances to certain types of shifts, such as rotations or flips of the inputs, for example.

The Shifts Dataset~\cite{malinin2021shifts} was originally constructed with the following attributes. First, the data is structured to have a `canonical partitioning' such that there are in-domain, or `matched' training, development and evaluation datasets, as well as a shifted development and evaluation dataset. The latter two datasets are also shifted relative to each other. Models are assessed on the joint in-domain and shifted development or evaluation datasets. This is because a model may be robust to certain examples of distributional shifts and yield accurate, low uncertainty predictions, and also perform poorly and yield high estimates of uncertainty on underrepresented data matched to the training set. Providing a dataset which contains both matched and shifted data enables better evaluation of this scenario. Second, it is assumed that at training or test time \emph{it is not known a priori} about whether or how the data is shifted. This emulates real-world deployments in which the variation of conditions cannot be sufficiently covered with data and is a more challenging scenario than one in which information about nature of shift is available~\cite{koh2020wilds}. In this work we maintain these attributes.

\paragraph{Evaluation} Robustness and uncertainty quality are jointly assessed via \emph{error-retention curves}~\cite{malinin-thesis,deepensemble2017,malinin2021shifts}. Given an error metric, error-retention curves depict the error over a dataset as a model's predictions are replaced by ground-truth labels in order of decreasing uncertainty. The area under this curve can be decreased either by improving the predictive performance of the model, such that it has lower overall error, or by providing uncertainty estimates which are better correlated with error. Thus, the area under the error retention curves (R-AUC) is a metric which jointly assesses robustness to distributional shift and uncertainty quality. More details are provided in appendix~\ref{apn:metrics}.

\paragraph{Choice of Baselines} Similarly to the original Shifts paper~\cite{malinin2021shifts}, we consider ensemble-based baselines in this work for three reasons. First, ensemble-based approaches are a standard way to \emph{both} improve robustness \emph{and} obtain interpretable uncertainties~\cite{malinin-thesis,deepensemble2017,Gal2016Dropout,galthesis,notinprincipled}. Second, ensembles are straightforward to apply to any task with little adaptation~\cite{malinin2021structured,malinin2021shifts,talarbel,talarbelpropagating,xiao2019wat,filos2020can,fomicheva2020unsupervised}. Third, they do not require information about the nature of distributional shift at training time, unlike other robust learning or domain adaptation methods, such as IRM~\cite{arjovsky2019invariant,koh2020wilds,sagawa2021extending}. The main downside of ensembles is their computational and memory cost, but there has been work on overcoming this limitation~\cite{malinin-endd-2019,ryabinin2021scaling,havasi2020training}. To our knowledge, there are no other approaches which have all three properties.

\section{White Matter Multiple Sclerosis Lesion Segmentation} \label{sec:wml}
The first dataset focuses on the segmentation of white matter lesions (WML) in 3D Magnetic Resonance (MR) brain images that are due to Multiple Sclerosis (MS). MS is a debilitating, incurable and progressive disorder of the central nervous system that negatively impacts an individual's quality of life. Estimates claim that every five minutes a person is diagnosed with MS, reaching 2.8 million cases in 2020 and that MS is two-to-four times more prevalent in women than in men~\cite{walton2020rising}. MRI plays a crucial role in the disease diagnosis and follow-up, as it allows physicians to manually track the lesion extension, dissemination, and progress over time~\cite{thompsonDiag2017}. However, manual annotations are expensive, time-consuming, and prone to inter- and intra-observer variations. Automatic, ML-based methods may introduce objectivity and labor efficiency in the tracking of MS lesions and have already showed promising results for the cross-sectional and longitudinal analysis of WML~\cite{zengReview2020}.

Patient data are rarely shared across medical centers and the availability of training images for machine learning methods is limited. No publicly available dataset fully describes the heterogeneity that the pathology presents in terms of disease severity and progression, reducing the applicability and robustness of automated models in real-world conditions. Furthermore, changes in the MRI scanner vendors, configurations of the magnetic field, or imaging software can lead brain scans to differ in terms of voxel resolution, signal-to-noise ratio, contrast parameters, slice thickness, non-linearity corrections, etc. Changes in the medical personnel using the devices also lead to variability in the imaging acquisition and annotation process. These differences, which are exacerbated when considering image acquisitions collected from multiple medical centers, represent a significant distributional shift for ML-based MS detection models. Models developed in one medical center (or set of centers) may transfer poorly to a different medical center, show little robustness to technical and pathological variability and thus yield poor performance. The development of robust MS lesion segmentation models which can indicate when and \emph{where} they are wrong would bring improvements in the quality and throughput of the medical care available to the growing number of MS patients. Ideally, this would allow patients to receive care and treatment tailored to their unique situations.

\paragraph{Task Description}
White matter MS lesion segmentation involves the generation of a 3D per-voxel segmentation mask of brain lesions in multi-modal MR images \cite{roviraEvidence2015}. Given an input 3D MRI scan, a model classifies each voxel into a lesion or non-lesion tissue. Two standard modalities for MS diagnosis are T1-weighted and, more commonly, Fluid-Attenuated Inversion Recovery (FLAIR) \footnote{Such modalities represent information captured by differing configurations of the scanner's magnetic field. In particular, FLAIR highlights the MS lesions as high-contrast regions within the gray-scale image~\cite{WATTJES2021653}.}.  %Voxel-level predictions are often combined into lesion- and patient-level predictions for use by the clinician.

\paragraph{Data}
The Shifts MS segmentation dataset is a combination of several publicly available and one unpublished datasets (see Table~\ref{tab:wml_meta}). Specifically, ISBI\cite{CARASS201777, carass2017}, MSSEG-1 \cite{commowick2018}, PubMRI \cite{Lesjak2017ANP} and a dataset provided by the university of Lausanne. The latter has not been released for privacy reasons and will be kept as a hidden evaluation set. However, we will set up a permanent leaderboard on Grand-Challenge, and it will be possible to evaluate models on it via dockers. Patient scans come from multiple clinical centers (locations in Table~\ref{tab:wml_meta}): Rennes, Bordeaux and Lyon (France), Ljubljana (Slovenia), Best (Netherlands) and Lausanne (Switzerland). The data from the locations Rennes, Bordeaux and Lyon originate from MSSEG-1; Best from ISBI; Ljubljana from PubMRI.

Each sample in the Shifts MS dataset consists of a 3D brain scan taken using both the FLAIR and T1 contrasts. Each sample has undergone pre-processing including denoising \cite{coupe2008}, skull stripping \cite{isensee2019} (brain mask is learned from the T1 modality image registered \cite{commowick2012} to the FLAIR space), bias field correction \cite{tustison2010} and interpolation to a 1mm isovoxel space. The ground-truth segmentation mask, also interpolated to the 1mm isovoxel space, is obtained as a consensus of multiple expert annotators and as a single mask for Best and Lausanne. Patient scans from different locations vary in terms of scanner models, local annotation (rater) guidelines, 
%for generating the consensus ground-truth segmentations,
scanner strengths (1.5T vs 3T) and resolution of the raw FLAIR scans. Table \ref{tab:wml_meta} details the main shifts that exist across different locations. 
\begin{table}[ht]
\fontsize{8}{9}\selectfont
\centering
\begin{small}
    \begin{tabular}{l|lclc|ccccc}
    \toprule
Location & Scanner & Field & Resolution (mm$^{3}$) & Raters &Trn & Dev$_{\text{in}}$ &Evl$_{\text{in}}$ & Dev$_{\text{out}}$ & Evl$_{\text{out}}$\\
\midrule
Rennes & S Verio & 3.0 T & \footnotesize $0.50 \times 0.50 \times 1.10$ & 7  &8 &2 &5 & 0 & 0\\
Bordeaux & GE Disc & 3.0 T & \footnotesize $0.47 \times 0.47 \times 0.90$ & 7 &5  &1 &2  & 0 & 0\\
\multirow{2}*{Lyon} 
& S Aera & 1.5 T & \footnotesize $1.03 \times 1.03 \times 1.25$ & \multirow{2}*{7} & \multirow{2}*{10} & \multirow{2}*{2} & \multirow{2}*{17} & \multirow{2}*{0} & \multirow{2}*{0}  \\
& P Ingenia & 3.0 T & \footnotesize $0.74 \times 0.74 \times 0.70$ & & & & & & \\
Best & P Medical & 3.0 T & \footnotesize $0.82 \times 0.82 \times 2.20$ & 2 & 10 & 2 & 9  & 0 & 0\\
\midrule
Ljubljana & S Mag & 3.0 T & \footnotesize $0.47 \times 0.47 \times 0.80$ & 3  & 0 & 0 & 0  & 25 & 0\\
Lausanne & S Mag& 3.0 T & \footnotesize $1.00 \times 1.00 \times 1.20$ & 2 & 0 & 0 & 0  & 0 & 74\\
  \bottomrule
    \end{tabular}
    \end{small}
\caption{Meta information and canonical splits for the WML dataset. Scanner models are: Siemens Verio, GE Discovery, Siemens Aera, Philips Ingenia, Philips Medical, Siemens Magnetom Trio.}
\label{tab:wml_meta}
\end{table}

For standardized benchmarking we have created a \emph{canonical partitioning} of the data into in-domain train, development (Dev) and evaluation (Evl) as well as shifted Dev and Evl datasets, described in Table~\ref{tab:wml_meta}. Rennes, Bordeaux, Lyon and Best are treated as the in-domain data. Ljubljana and Lausanne are treated as publicly available and heldout shifted development and evaluation sets, respectively. For locations containing multiple scans per patient, we ensure that all scans for a particular patient appear only in one dataset. This partitioning was selected to create a clear shift between the in-domain and shifted data. Refer to Appendix \ref{app:lesion_canon} for details regarding the choice of the splits.

\paragraph{License and Distribution} 
The component datasets of our benchmark are publicly available under a range of licenses. PubMRI and ISBI are available under creative commons licenses. MSSEG-1 is distributed via credentialized access on Shanoir. However, we have jointly re-released these datasets on Zenodo with uniform pre-processing - data is available here \url{https://zenodo.org/record/7051658} and \url{https://zenodo.org/record/7051692}. Due to privacy concerns and legal limitations, it is not possible to provide download access to the Lausanne dataset. However, it will be possible to freely evaluate dockerized models on this dataset by submitting to a permanent public leaderboard hosted on Grand-Challenge. Further details on the licensing and distribution are provided in appendices~\ref{apn:datasheet} and~\ref{apn:med}.

\paragraph{Assessment} 
Segmentation of 3D MRI images is typically assessed via the Dice Similarity Coefficient (DSC)\cite{Dice1945MeasuresOT, Srensen1948AMO} between manual lesion annotations and the model's prediction. However, DSC is strongly correlated with \emph{lesion load} - patients with higher lesion load (volume occupied by lesion) will have a higher DSC~\cite{reinke2021common}. Thus, we consider an adapted (normalized) DSC (nDSC) that de-correlates model performance and lesion load. %The normalization is achieved by scaling the false positives in each patient.
A description of this metric is provided in Appendix~\ref{app:adaptedDSC}. Both DSC and nDSC assess the ability of the model to perform the exact delineation of lesions in the input image as the metric is voxel-based. The performance metric reported for a model will be the average nDSC across all patients in a dataset. Given the nDSC scores, we construct an error-retention curve and calculate the area under the curve~\footnote{Technically, we calculate the above between the curve and a horizontal line at 1, as nDSC is `accuracy' metric -- ie, higher is better.}. %with only foreground voxels (similar approach to \cite{mehtaBrats2021}).
We additionally assess the lesion detection ability of the models with F1-score that only rewards the ability of models to detect white matter lesions with less emphasis on the exact shape or boundary of the predicted lesions. Refer to Appendix~\ref{appendix:lesionscalef1} for further details and results on lesion-level detection.

\paragraph{Methods} 
The baseline segmentation models are based on the 3D UNET architecture\cite{iek20163DUL} with hyperparameters tuned according to \cite{la2020multiple}. Specifically, the model is trained for a maximum of 300 epochs with early-stopping based on Dev-in performance. The model relies on splitting the volume into 3D patches of $96\times 96\times 96$ voxels; at training time 32 patches are sampled from each input volume with a central lesion voxel while; at inference  patches overlapping by 25\% are selected across the whole 3D volume with Gaussian weighted averaging for the final prediction of each voxel belonging to multiple patches. The model probabilities for each voxel are thresholded to generate the per-voxel segmentation map. The threshold is tuned on the Dev-in split. A deep ensemble~\cite{deepensemble2017} is formed by averaging the output probabilities from 5 distinct single UNET models. Monte Carlo dropout~\cite{Gal2016Dropout} (MCDP) ensembles are also considered a baseline. Here, 5 UNET-DP models are trained with 50\% dropout in each model. For MCDP, a single model is taken and dropout is turned on at inference time with an ensemble formed from 5 separate runs of the model (as the dropout introduces stochasticity). The process is repeated for each of the single models with dropout to get averaged results. Finally, we also consider deep ensembles of UNETR \cite{unetr2021} models, which feature a transformer-based encoder and a convolutional decoder. The training and inference regime for the UNETR is identical to the UNET. As each single model yields a per-voxel probabilities, ensemble-based uncertainty measures\cite{malinin-thesis,malinin2021structured} are available for uncertainty quantification. In this work, all ensemble models use reverse mutual information \cite{malinin2021structured} as the choice of uncertainty measure. Single models use the entropy of their output probability distribution at each voxel to capture the uncertainty. All results reported for single models are the mean of the individual model performances.

\paragraph{Baseline Results.} 
Table \ref{tab:performance_nDSC} presents voxel-level predictive performance and joint robustness and uncertainty performance of the considered baselines in terms of nDSC and R-AUC, respectively.\footnote{Please refer to Table \ref{tab:app_performance_F1} in the appendix for a lesion-level assessment of performance}. Several trends are evident in the results. Firstly, comparing the in-domain predictive performance against the shifted performance, it is clear that the shift in the location leads to severe degradation in performance at the voxel-scale with drops exceeding 10\% nDSC. This clearly shows that out benchmark allows discriminating between robust and non-robust models. Secondly, the transformer-based architecture, UNETR, is able to outperform the fully convolutional architecture, UNET, for all models by about 2\% nDSC across the various splits. This demonstrates that transformer based approaches are promising for medical imaging, despite the low-data scenario. However, it is also valuable to highlight that even though UNETR yields better performance, the degree the performance degradation is about the same as for the UNET-based models. Third, dropout, as a regularisation technique, adversely affects the UNET architecture and leads to a severe performance drop. This seems to suggest that either the current learning procedure is not very stable and additional noise prevents good convergence, or that the 'standard' UNET model for the task is too small, and dropout over-regularizes it. Finally, comparing deep ensembles against single models, it is clear that ensembling, as expected, boosts predictive performance. Notably, the UNETR gain more performance on shifted data from ensembling than UNET models. At the same time, dropout ensembles yield a decrease in performance.
\begin{table}[ht]
% \fontsize{8}{9}\selectfont
\centering
%\begin{small}
    \begin{tabular}{ll|llll|llll}
    \toprule
\multirow{2}{*}{Type} & \multirow{2}{*}{Model} & \multicolumn{4}{c|}{nDSC (\%) $\left(\uparrow\right)$} & \multicolumn{4}{c}{R-AUC (\%) $\left(\downarrow\right)$}  \\
& & Dev$_{\text{in}}$  & Dev$_{\text{out}}$ & Evl$_{\text{in}}$ & Evl$_{\text{out}}$ & Dev$_{\text{in}}$  & Dev$_{\text{out}}$ & Evl$_{\text{in}}$ & Evl$_{\text{out}}$ \\
\midrule
\multirow{3}{*}{Single} & UNET & $68.54$ & $49.33$ & $67.59$  & $55.79$ & $2.51$  & $7.84$ & $2.77$ & $9.87$ \\
 & UNET-DP &  $59.73$  & $48.35$ & $63.93$ & $54.43$ & $2.62$  & $8.76$ & $2.66$ & $9.71$\\
 & UNETR  & $71.21$ & $51.60$ & $69.27$ & $56.76$ & $1.89$ & $6.17$  & $1.95$& $6.47$ \\
\midrule
 \multirow{3}{*}{Ensemble}  & UNET & $69.70$  & $50.85$ & $68.89$ & $57.53$ & $1.17$  & $4.66$ & $1.76$ & $7.40$ \\
& UNET-DP & $60.65$  &  $44.70$ & $61.78$ & $50.06$ & $1.92$  & $6.77$ & $2.52$ & $7.89$ \\
 & UNETR & $\textbf{72.51}$ &  $\textbf{53.46}$ & $\textbf{71.41}$ & $\textbf{59.49}$ & $\textbf{0.34}$ & $\textbf{1.52}$ & $\textbf{0.63}$ & $\textbf{2.88}$ \\
  \bottomrule
    \end{tabular}
%    \end{small}
\caption{Segmentation performance (nDSC) and joint eval of robustness and uncertainty (R-AUC).}
\label{tab:performance_nDSC}
\end{table}

Now let's examine the baselines in terms of joint assessment of robustness and uncertainty. Again, there are a number of observations to be made. Firstly, there is again a clear degradation of performance between the shifted and in-domain data. Secondly, UNETR models, both single and ensembled, yield by far the best performance, which shows that they are both more robust and yield better uncertainties. Thirdly, what is especially notable is that despite inferior predictive performance relative to the single-model counterpart on shifted data (48.35 vs 44.7), the MCDP ensemble yields improved performance in terms of R-AUC (8.76 vs 6.77). This highlights the value in quantifying knowledge (epistemic) uncertainty, and that systems which are less robust can still be competitive by providing informative estimates of uncertainty.

\section{Vessel Power Estimation}\label{sec:ship}

The second dataset involves predicting the energy consumption of cargo-carrying vessels in different weather and operating conditions. Such models are used, for example by DeepSea, to optimize route of cargo vessels for minimum fuel consumption. Maritime transport delivers around 90\% of the world's traded goods \cite{christodoulou2019sustainable}, emitting almost a billion tonnes of CO$_2$ annually and increasing \cite{hilakari2019carbon}. Energy consumption varies greatly depending on the chosen routes, speeds, operation and maintenance of ships, but the complex underlying relationships are not fully known or taken into account at the time these decisions are made, leading to significant fuel waste. Lack of predictability of fuel needs also leads to vessels carrying more fuel than necessary, costing even more fuel to carry. Training accurate consumption models, both for use on their own and for downstream route optimisation, can therefore help significantly reduce costs and emissions~\cite{gkerekos2019machine, zhu2020predicting}. 

While performance data is increasingly collected from vessels, data is still scarce and sensors are prone to noise. The weather and sea conditions that affect vessel power are highly variable based on seasonality and geographical location and cannot all be fully measured. Further, phenomena such as the accumulation of marine growth on the vessel's hull (hull fouling) cause the relationship between conditions and power to shift over time in unpredictable ways. The result of the above is that significant distributional shifts can be expected to occur between the real use cases of models and the data used to train and evaluate them. Inaccurate power prediction and the resultant errors in fuel planning and route optimisation can be considerably costly, hazardous and place the vessel, its crew and cargo at high risk. For example, in the context of routing and autonomous navigation, inaccurate modeling of speed-power relation can lead to instructed speeds that cause the engine to enter unsafe barred RPM and power zones or the adoption of excessive speeds in extreme weather conditions. In the context of automated bunker planning - if a vessel incorrectly predicts the fuel requirements for a voyage it could run out of fuel in the middle of the ocean. Thus, the development of uncertainty-aware and robust power consumption models is essential to enable the safe and effective deployment of this technology to reduce the carbon footprint of global supply chains.

\paragraph{Task Description} 
This is a scalar regression task which involves predicting the current power consumption of a merchant vessel at a particular timestep based on tabular features describing vessel and weather conditions. A probabilistic regression model would yield a probability density over the power consumption. The prediction of the output is mostly attributed to the current timestep, but due to transient effects (e.g inertia) and hull fouling previous timesteps can also affect target power.

\paragraph{Assessment} 
Predictive performance is assessed using the standard metrics: Root Mean Square Error (RMSE), Mean Absolute Error (MAE) and Mean Absolute Percentage Error (MAPE). Area under Mean Square error (MSE) and F1 retention curves~\cite{malinin-thesis,malinin2021shifts} is used to assess jointly the robustness to distributional shift and uncertainty quality. The respective performance metrics are named R-AUC and F1-AUC. Following the methodology proposed by~\cite{malinin2021shifts}, we use the MSE as the error metric and for F1 scores we consider acceptable predictions those with $MSE < (500 kW)^2$. As the uncertainty measure, we use the total variance (i.e. the sum of data and knowledge uncertainty~\cite{malinin-thesis}). A good model should have a small R-AUC and large F1-AUC. 

% \paragraph{Data} 
% \begin{table}
%   \begin{minipage}{.5\linewidth}
%     \centering
%      \begin{small}
%       \begin{tabular}{l|ccc}
%         \toprule
%         Wind speed & Train & Validation & Test \\
%         \midrule
%         0-3 Beaufort & 231626 & 8017 & 20355 \\
%         3-4  Beaufort & 118698 & 4108 & 10448 \\
%         4-5  Beaufort & 172866 & 5983 & 15218 \\
%         > 5 Beaufort & 0 & 18108 & 46021 \\
%         \midrule
%         \textbf{Total}  & 523190 & 36216 & 92042 \\
%         \bottomrule
%       \end{tabular}
%     \end{small}
%   \caption{Wind speed partitioning}\label{tab:second}
%   \end{minipage}
%   \begin{minipage}{.5\linewidth}
%     \centering
%      \begin{small}
%       \begin{tabular}{l|ccc}
%         \toprule
%         Time & Train & Validation & Test \\
%         \midrule
%         interval 1 & 523190 & 18108 & 46021 \\
%         interval 2 & 0 & 18108 & 46021  \\
%         \midrule
%         \textbf{Total}  & 523190 & 36216 & 92042 \\
%         \bottomrule
%       \end{tabular}
%     \end{small}
%   \caption{Time partitioning}\label{tab:second}
%   \end{minipage}
%   \caption{Canonical Splits for Ship Power Consumption dataset. - REPLACE WITH DIAGRAMS}
%   \label{tab:real_ds_partitioning}
% \end{table}

The Shifts vessel power estimation dataset consists of measurements sampled every minute from sensors on-board a merchant ship over a span of 4 years, cleaned and augmented with weather data from a third-party provider. The task is to predict the ships main engine shaft power, which can be used to predict fuel consumption given an engine model, from the vessel's speed, draft, time since last dry dock cleaning and various weather and sea conditions. Noise in the data arises due to sensor noise, measurement and transmission errors, and noise in historical weather. Distributional shift arises from hull performance degradation over time due to fouling, sensor calibration drift, and variations in non-measured sea conditions such as water temperature and salinity, which vary across regions and times of year. The features are detailed in Appendix \ref{deepsea_features}.

\begin{figure}[h]
\includegraphics[scale=0.39]{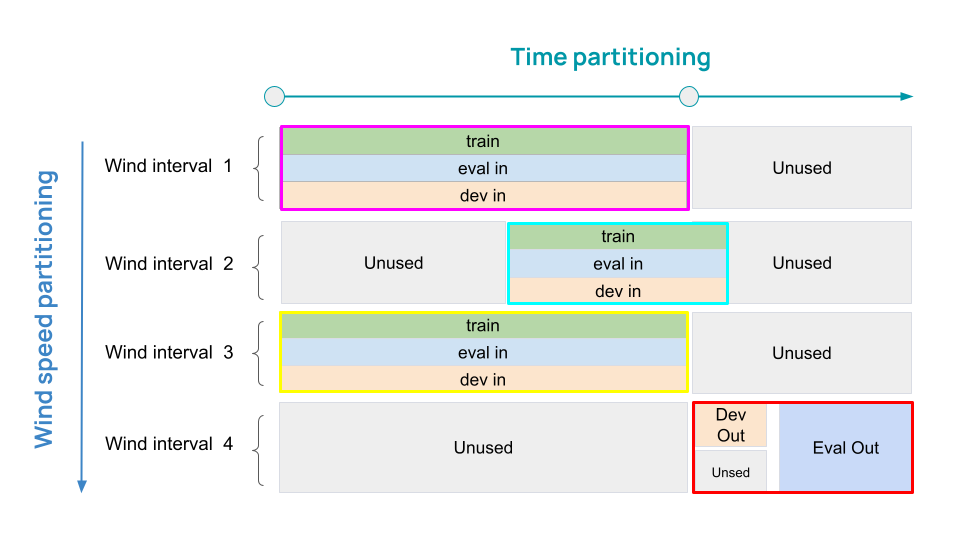}
\centering
\caption{Canonical partitioning for vessel power dataset. Wind intervals represent 0-3, 3-4, 4-5 and >5 on the Beaufort scale. Train, dev and eval sets contains 530706, 18368 and 47227 records.}
\label{fig:ds_partitioning}
\end{figure}

To provide a standardized benchmark, we have created a \emph{canonical partition} on the dataset into in-domain train, development and evaluation as well as distributionally shifted development and evaluation splits, as is the standard in Shifts. The dataset is partitioned along two dimensions: wind speed and time, as illustrated in Figure \ref{fig:ds_partitioning}. Wind speed is a proxy for unmeasured components of the sea state, while partitioning in time aims to capture effects such as hull fouling and sensor drift.

\begin{figure}[htp!]
    \begin{minipage}[b]{.52\linewidth}
        \centering
        \includegraphics[trim={0 11.9cm 9cm 0.2cm}, clip, scale=0.28]{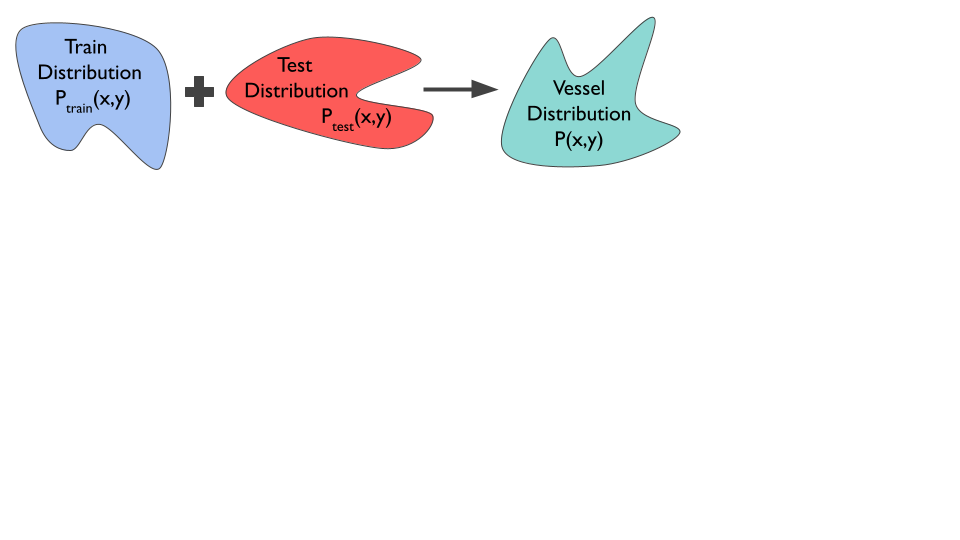}
        \caption{Regular split of data.}
        \label{fig:ds_regular_data_split}
    \end{minipage}
    \begin{minipage}[b]{.42\linewidth}
        \centering
        \includegraphics[trim={0 11.5cm 13cm 0}, clip, scale=0.28]{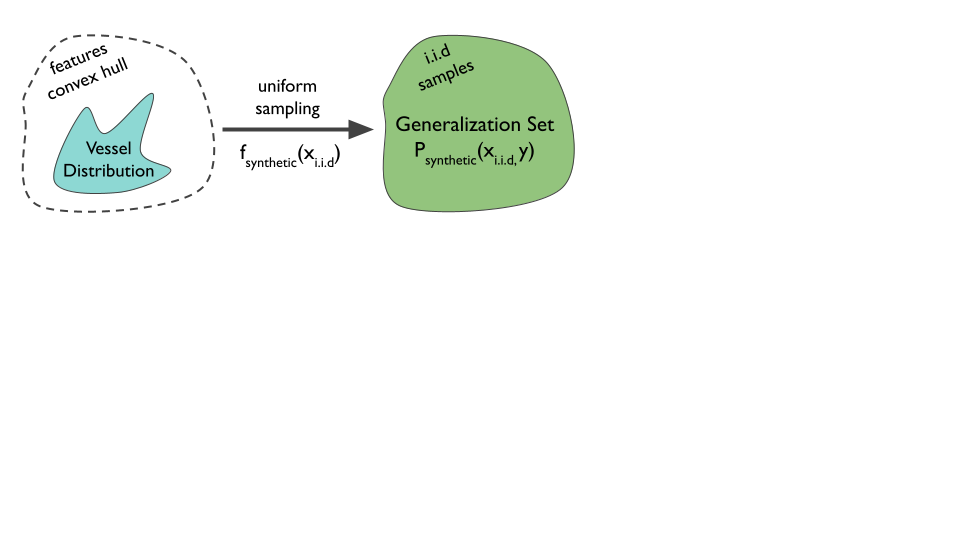}
        \caption{Generalization set.}
        \label{fig:ds_gen_set_idea}
    \end{minipage}
\end{figure}

In addition to the standard canonical benchmark we provide a synthetic benchmark dataset created using an analytical physics-based vessel model. The synthetic data contains the input features real data, but the target power labels are replaced with the predictions of a physics model. Given that vessel physics are well understood, it is possible to create a model of reality which captures most relevant factors of variation and model them robustly. However, this physics model is still a simplified version of reality and therefore is an easier task with fewer factors of variation than the real dataset. Crucially, it assumes that the dataset features constitute a \emph{sufficient} description of all relevant factors of variation, which may not be the case in reality. A significant advantage of the physics model is that it allows generating a \emph{generalization dataset} which covers the \emph{convex hull} of possible feature combinations (Figure \ref{fig:ds_gen_set_idea}). Here, data is generated by applying the model to input features independently and uniformly sampled from a predefined range. Thus, models can be assessed on both rare and common combinations of conditions, which are now equally represented. For example, vessels are unlikely to adopt high speeds in severe weather, which could lead ML models to learn a spurious correlation. This bias would not be detected during evaluation on real data as the same correlation would be present. Evaluating a candidate model on this generalization set tests its ability to properly disentangle causal factors and generalise to unseen conditions. This generalisation set enables assessing model robustness with greater coverage, even if on a simplified version of reality.

We provide both the synthetic generalisation of 2.5 millions samples as well as a synthetic version of the real data. The latter has input features sampled from the same vessel and split into the same canonical partitions as the real dataset. However, the target labels are provided by the physics models. This dataset allows establishing common ground between the real and synthetic tasks. The real and synthetic datasets can be used together. The real-world performance and robustness to unseen latent factors is assessed using the real dataset, while the synthetic generalisation set broad assessment of generalisation and causal disentanglement. Indeed, the best models have high performance on both. Models which perform well on the the generalisation set and poorly on real data are not robust to unseen latent factors. Conversely, models that perform poorly on the generalisation set and well on real data are strongly affected by spurious correlations in the measures features.

\paragraph{License and Distribution} 
This data is provided by DeepSea under a creative commons CC BY NC SA 4.0 license. The data is available at \url{https://zenodo.org/record/7057666}. Further details on the licensing and distribution are provided in appendices~\ref{apn:datasheet} and~\ref{apn:ship}.

\paragraph{Methods} 
We examine the following range of baseline models: Deep Neural Networks (DNNs), Monte-Carlo Dropout ensembles of DNN, variational DNNs models and a proprietary DeepSea symbolic model. Additionally, we also consider deep ensembles of 10 of each of the aforementioned models. In all cases, each ensemble member predicts the parameters of the conditional normal distribution over the target (power) given the input features. %The models are optimized to minimise the negative log likelihood loss function. 
As a measure of uncertainty use the total variance - the sum of the variance of the predicted mean and the mean of predicted variance across the ensemble\cite{malinin-thesis}. All baselines methods are detailed in Appendix~\ref{apn:ship-baselines}. 

\paragraph{Baseline Results} 
Table~\ref{tab:predictive_singel_ens} presents the results of evaluating baseline models on both the real and synthetic versions of the power estimation data. Several trends can be observed. First, the results show that on the proposed data split the shifted data is more challenging for the models to handle - both the error rates are higher on the shifted partitions. Secondly, it is clear that the real dataset is overall more challenging than the synthetic dataset for all models, which is expected, as reality contains far more unknown factors of variation. Third, of the single model approaches, the variational inference (VI) model consistently yields both the best predictive performance and the best retention performance on all of the real and synthetic canonical partitions. However, on the synthetic generalisation set, which uniformly covers the convex hull of possible inputs, the proprietary DeepSea symbolic model does best in terms of predictive quality and second-best in terms of R-AUC. Note, the symbolic model yields worst performance RMSE and R-AUC on real data. This highlights the value of the generalisation set - to show which models are overall more robust, rather than just performing well on more typical events in standard train/dev/eval splits. Fourth, the ensembles consistently outperform the mean performance of their single seed counterparts. The ensemble-based results show similar a similar story to those of single models. Here, the ensemble VI model is broadly comparable to or better than the other ensemble models on all splits except for the generalisation set, where the ensemble of symbolic models does best. Overall, the results show that the ensemble VI models achieves the best overall balance of robustness and uncertainty across all splits.
\begin{table}[H]
\centering
    \begin{small}
        \begin{tabular}{l|l|lll|ll||ll|r}
            \toprule
            \multirow{3}{*}{ Method } & \multirow{3}{*}{ Model } & \multicolumn{5}{c||}{RMSE (kW) $\downarrow$} & \multicolumn{3}{c}{R-AUC ($10^5 kW^2$) $\downarrow$} \\ 
            \cmidrule{3-10}
             & & \multicolumn{3}{c|}{Synthetic} & \multicolumn{2}{c||}{Real} & \multicolumn{2}{c|}{Synthetic} & \multicolumn{1}{c}{Real} \\
             & & In & Out & Gen & In & Out &  Full & Gen & \multicolumn{1}{c}{Full} \\
            \midrule
            \multirow{4}{*}{Single} &\multirow{1}{*}{DNN} & 1084  &  1116  &  1487  &  1296  &  1985  &  4.49  &  5.27  &  10.97  \\
            &\multirow{1}{*}{MC dropout}  &  1078  &  1122  &  1526 &   1271  &  1954  &    4.54  &  5.39  &  10.00 \\
            &\multirow{1}{*}{VI} & \textbf{1072}  &  \textbf{1109}  &  1458  &   \textbf{1255}  &  \textbf{1916}  &   \textbf{4.33}  &  \textbf{4.53}  &  \textbf{9.57}  \\
            &\multirow{1}{*}{Symbolic}&  1120  &  1137   &  \textcolor{red}{\textbf{1213}}  &  1403  &  2366  &    5.13  &  4.55  &  17.51  \\
            \midrule
             \multirow{4}{*}{Ens.} &\multirow{1}{*}{DNN} & 1076 & \textbf{1099}  & 1427  & 1264 & 1928  & 4.32 & \textcolor{red}{\textbf{4.20}} & 9.52  \\
            & \multirow{1}{*}{MC dropout} & \textbf{1069} & 1111 & 1498   & 1248 & 1925 & 4.47 & 4.97  & 9.28 \\
            &\multirow{1}{*}{VI} &  \textbf{1069} & 1104 &  1446 & \textbf{1243} & \textbf{1895} &  \textbf{4.29} & 4.32 & \textbf{9.13} \\
            &\multirow{1}{*}{Symbolic} &  1117 & 1133  & \textcolor{red}{\textbf{1204}}& 1393 & 2341 &  5.09 & 4.41  & 13.56 \\
            \bottomrule
        \end{tabular}
        \end{small}
\caption{Results on the real and synthetic canonical eval partitions and on the generalization set.}
\label{tab:predictive_singel_ens}
\end{table}

\begin{figure}[htp!]\label{fig:ship_correlation}
\centering
    \begin{tabular} {ccc}
        \subfigure[%Correlation of retention performance for the real and synthetic full sets.
        ]{
            \includegraphics[trim={9.cm, 1.2cm, 10.1cm, 0.03cm}, clip, width=0.29\textwidth]{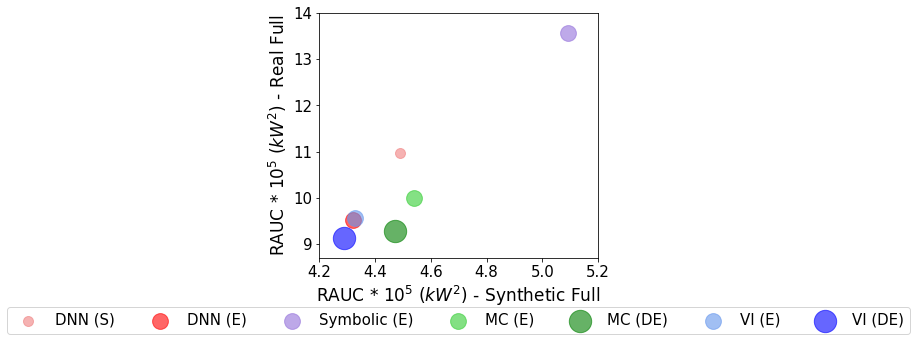}
            \label{fig:ds_retention_correlation}
        } &
        \subfigure[%Predictive performance of the real full set versus the generalization set.
        ]{
            \includegraphics[trim={8.5cm, 1.2cm, 10.6cm, 0.03cm}, clip,width=0.29\textwidth]{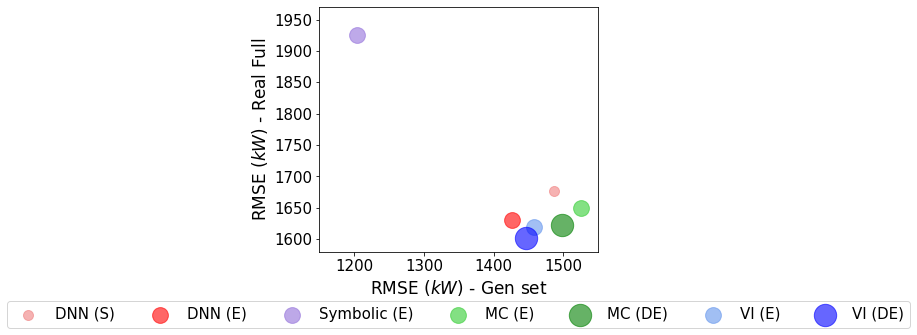}
            \label{fig:robustness}
        } &
        \subfigure[%Retention vs predictive performance of the generalization set.
        ]{
            \includegraphics[trim={9cm, 1.2cm, 10.1cm, 0.03cm}, clip,width=0.29\textwidth]{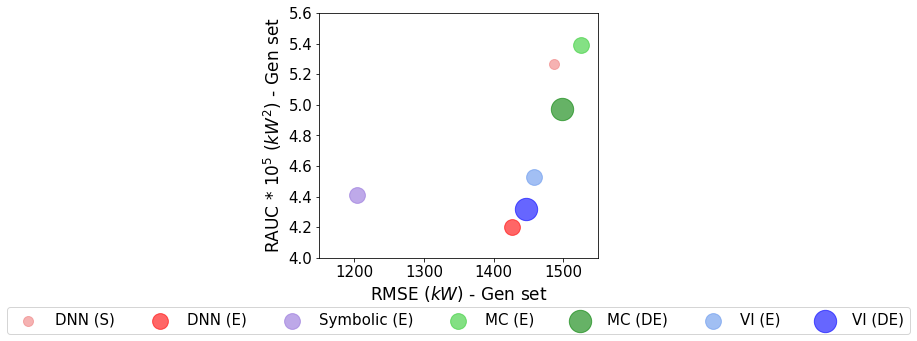}
            \label{fig:ds_retention_robustness}
        }\\
        \multicolumn{3}{c}{
            \includegraphics[trim={0cm, 0.2cm, 0cm, 10.8cm}, clip,width=0.85\textwidth]{figures/retention.png}
        }\\
    \end{tabular}
\caption{Power estimation key performance trends. Circles' size increases as we go from single models (S) $\rightarrow$ ensembles (E) $\rightarrow$ deep ensembles (DE).} %Color notation: Red shades for DNN, purple for Symbolic, green shades for MC dropout and blue shades for VI methods.}
\label{fig:ds_key_trends}
\end{figure}

\vspace{-0.4cm}
Figure~\ref{fig:ship_correlation} reveals additional insights. Figure~\ref{fig:ship_correlation}a shows that joint uncertainty and robustness performance of all models on the full (in+out) real and synthetic evaluation sets are strongly correlated. Figure~\ref{fig:ship_correlation}b shows that there is a trade-off between broad robustness and high performance on the real data. The symbolic model, a low variance, high bias models, isn't great overall, but neither does it fail as strongly as the neural models in unfamiliar situations. Conversely, neural models (low bias, high variance models) are better able to exploit the correlations within the real data, but feature more brittle generalisation. Finally, Figure~\ref{fig:ship_correlation}c shows the benefits of uncertainty quantification. While the neural models are consistently worse than the symbolic model on the generalisation set, they show comparable or superior joint robustness and uncertainty (R-AUC) to the symbolic models. Thus, using uncertainty estimates to detect errors, the neural models can achieve superior operating points to the symbolic models.
Additional results are provided in Appendix~\ref{apn:ship-results}.

\section{Discussion}\label{sec:discussion}

The two new benchmark datasets described in this paper will extend Shifts and enable additional insights to be drawn. The WML segmentation dataset brings both a new modality, a new predictive task and a very low-data regime, where models are almost always operating under some degree of distributional shift even on nominally matched data. This is quite different from the three datasets present in the original release of Shifts, which all operated in a large-data regime. The marine cargo-vessel dataset bring another tabular regression task to the table. While similar to the shifts weather forecasting dataset~\cite{malinin2021shifts}, which is also a tabular regression task, there are key differences. First, while the features are `tabular', the data described is entirely different - effectively a modality unto itself. Secondly, the fact that naval engineering and ocean physics are sufficiently well understood that a reliable physics simulation can be created which can then be used to assess model robustness and generalisation.  The real data tests robustness to distributional shift in the presence of real noise as in the other tasks while the synthetic data enables evaluation of the ability of the model to disentangle causal factors over the convex hull of the input features. This is a novel feature of the Shifts benchmark which does not appear in other uncertainty and robustness benchmarks. 

It must be stated that regardless of how well made, the current benchmark still has limitations. First, our benchmark, like any fixed benchmark, assess robustness and uncertainty quality on a particular set of samples, rather than across all possible shifts. Thus, even if an evaluation set is large, a model still may get 'lucky', achieving good robustness or uncertainty quality on the benchmark, but fail in deployment. Thus, insights which are consistent on all Shifts benchmarks are more reliable than those which are observed on only one task.

Another limitation of our benchmark is how uncertainty is assessed. In this work we assess how well uncertainty estimates correlate with the degree or likelihood of an error. Theoretically, uncertainty estimates can enable errors to be detected and risk mitigating actions taken. However, our benchmark does not actually assess using uncertainty to take risk mitigating actions or convey critical information in a downstream application. The main difficulty is that there is limited consensus on \emph{how} uncertainty can be used in any particular application. For example, we believe that the communication of the uncertainty level of an AI system in health care applications is crucial. Uncertainty can, on one side, support the rapid trust calibration of the system and, on the other side, speed up the MRI assessment by attracting the attention of experts to the most uncertain areas~\footnote{See appendix~\ref{apn:med-uncertainty} for an example}. However, it is unclear what is the best way to convey voxel-level uncertainty to a clinician such that it is an asset, rather than a distraction. Furthermore, even if a clinical trial were run, it is unclear how it should be assessed. Alternatively, cargo vessel power estimation models are typically used for route optimisation. While power consumption uncertainty estimates are conceptually useful, it is unclear how a route optimisation algorithm would make use of them and weather the result should be assessed via real-life fuel consumption or some other metric. While we do not have answer to these questions, we believe that there is value in making them explicit - answering them will be the next step once models which are robust to distributional shift and which yield uncertainty estimates correlated with the likelihood of errors become widely available.

\section{Conclusion}\label{sec:conclusion}

In this paper we extended the Shifts Dataset~\cite{malinin2021shifts} with two datasets sourced from industrial, high-risk applications of high societal importance which feature ubiquitous distributional shift. Specifically, we provide a white matter Multiple Sclerosis lesion segmentation in 3D MR brain images and a marine cargo vessels power consumption estimation dataset. The MS segmentation task features an extremely low-data regime, ubiquitous distributional shift due to heterogeneity of the underlying pathology as well as changes in scanner model, configuration and personnel across multiple medical centers. This presents an extremely challenging scenario in which models are rarely reliable, motivating the need for improving robustness and estimating uncertainty. A further conceptual challenge is that predictions and uncertainty estimates are produced as the individual voxel level and additional work needs to be done in order to provide relevant lesion and patient level information to clinicians. Ultimately, reliable ML systems for this application can enable a better understanding on the pathology, an increase in throughput to meet the rising incidence of MS and a more individual treatment plan. The cargo vessel power estimation task features power consumption data taken from a real ship in operation. While not operating in the low-data regime, the task still features extensive distribution shift due to changing weather, ship and route conditions. A further novelty of this dataset is the construction of a synthetic version of it using a physics model. This enables the creation of a \emph{generalisation} set which covers the full convex hull of the input features. While representing a simplified model of reality, this is nevertheless a valuable tool to assess which design features improve model performance on the generalisation set. Reliable power consumption estimation systems can help optimize fuel usage, decrease the amount of extra fuel transported and thereby reduce the climate impact of the shipping industry. Adding these datasets to the Shifts benchmark will enable researchers to further investigate generalization under distributional shift and uncertainty estimation and come up with new insights and solutions on how to create robust and reliable ML models for high-risk medical and industrial applications.

% \section{Medical Use Cases}
% \begin{itemize}
%     \item concrete examples of utility ... 
%     \item dataset specific results
% \end{itemize}

% \section{Ship Use Case}
% \begin{itemize}
%     \item 
% \end{itemize}

\newpage
\bibliographystyle{paper/IEEEbib}
\bibliography{paper/bibliography}

\begin{thebibliography}{10}

\bibitem{malinin2021shifts}
Andrey Malinin, Neil Band, Yarin Gal, Mark Gales, Alexander Ganshin, German
  Chesnokov, Alexey Noskov, Andrey Ploskonosov, Liudmila Prokhorenkova, Ivan
  Provilkov, Vatsal Raina, Vyas Raina, Denis Roginskiy, Mariya Shmatova,
  Panagiotis Tigas, and Boris Yangel,
\newblock ``Shifts: A dataset of real distributional shift across multiple
  large-scale tasks,''
\newblock in {\em Thirty-fifth Conference on Neural Information Processing
  Systems Datasets and Benchmarks Track (Round 2)}, 2021.

\bibitem{Datasetshift}
Joaquin Quiñonero-Candela,
\newblock {\em {Dataset Shift in Machine Learning}},
\newblock The MIT Press, 2009.

\bibitem{koh2020wilds}
Pang~Wei Koh, Shiori Sagawa, Henrik Marklund, Sang~Michael Xie, Marvin Zhang,
  Akshay Balsubramani, Weihua Hu, Michihiro Yasunaga, Richard~Lanas Phillips,
  Sara Beery, Jure Leskovec, Anshul Kundaje, Emma Pierson, Sergey Levine,
  Chelsea Finn, and Percy Liang,
\newblock ``Wilds: A benchmark of in-the-wild distribution shifts,'' 2020.

\bibitem{murphy}
Kevin~P. Murphy,
\newblock {\em {Machine Learning}},
\newblock The MIT Press, 2012.

\bibitem{aisafety}
Dario Amodei, Chris Olah, Jacob Steinhardt, Paul~F. Christiano, John Schulman,
  and Dan Man{\'{e}},
\newblock ``Concrete problems in {AI} safety,''
  \url{http://arxiv.org/abs/1606.06565}, 2016,
\newblock arXiv: 1606.06565.

\bibitem{malinin-thesis}
Andrey Malinin,
\newblock {\em Uncertainty Estimation in Deep Learning with application to
  Spoken Language Assessment},
\newblock Ph.D. thesis, University of Cambridge, 2019.

\bibitem{active-learning}
Burr Settles,
\newblock ``Active learning literature survey,''
\newblock Tech. {R}ep., University of Wisconsin-Madison Department of Computer
  Sciences, 2009.

\bibitem{kirsch2019batchbald}
Andreas Kirsch, Joost Van~Amersfoort, and Yarin Gal,
\newblock ``Batchbald: Efficient and diverse batch acquisition for deep
  bayesian active learning,''
\newblock {\em Advances in neural information processing systems}, vol. 32, pp.
  7026--7037, 2019.

\bibitem{talarbel}
Tanya Nair, Doina Precup, Douglas~L Arnold, and Tal Arbel,
\newblock ``Exploring uncertainty measures in deep networks for multiple
  sclerosis lesion detection and segmentation,''
\newblock {\em Medical image analysis}, vol. 59, pp. 101557, 2020.

\bibitem{malinin-endd-2019}
Andrey Malinin, Bruno Mlodozeniec, and Mark~JF Gales,
\newblock ``Ensemble distribution distillation,''
\newblock in {\em International Conference on Learning Representations}, 2020.

\bibitem{Gal2016Dropout}
Yarin Gal and Zoubin Ghahramani,
\newblock ``{Dropout as a Bayesian Approximation: Representing Model
  Uncertainty in Deep Learning},''
\newblock in {\em Proc. 33rd International Conference on Machine Learning
  (ICML-16)}, 2016.

\bibitem{deepensemble2017}
B.~Lakshminarayanan, A.~Pritzel, and C.~Blundell,
\newblock ``{Simple and Scalable Predictive Uncertainty Estimation using Deep
  Ensembles},''
\newblock in {\em Proc. Conference on Neural Information Processing Systems
  (NIPS)}, 2017.

\bibitem{baselinedetecting}
Dan Hendrycks and Kevin Gimpel,
\newblock ``{A Baseline for Detecting Misclassified and Out-of-Distribution
  Examples in Neural Networks},'' \url {http://arxiv.org/abs/1610.02136}, 2016,
\newblock arXiv:1610.02136.

\bibitem{ashukha2020pitfalls}
Arsenii Ashukha, Alexander Lyzhov, Dmitry Molchanov, and Dmitry Vetrov,
\newblock ``Pitfalls of in-domain uncertainty estimation and ensembling in deep
  learning,''
\newblock in {\em International Conference on Learning Representations}, 2020.

\bibitem{trust-uncertainty}
Yaniv Ovadia, Emily Fertig, Jie Ren, Zachary Nado, D~Sculley, Sebastian
  Nowozin, Joshua~V Dillon, Balaji Lakshminarayanan, and Jasper Snoek,
\newblock ``Can you trust your model's uncertainty? evaluating predictive
  uncertainty under dataset shift,''
\newblock {\em Advances in Neural Information Processing Systems}, 2019.

\bibitem{mnist}
Y.~LeCun, L.~Bottou, Y.~Bengio, and P.~Haffner,
\newblock ``Gradient-based learning applied to document recognition,''
\newblock {\em Proceedings of the \textsc{ieee}}, vol. 86, pp. 2278–2324,
  1998.

\bibitem{svhn}
Ian~J. Goodfellow, Yaroslav Bulatov, Julian Ibarz, Sacha Arnoud, and Vinay~D.
  Shet,
\newblock ``Multi-digit number recognition from street view imagery using deep
  convolutional neural networks,'' 2013,
\newblock arXiv:1312.6082.

\bibitem{cifar}
Alex Krizhevsky,
\newblock ``Learning multiple layers of features from tiny images,''
\newblock 2009.

\bibitem{imagenet-r}
Dan Hendrycks, Steven Basart, Norman Mu, Saurav Kadavath, Frank Wang, Evan
  Dorundo, Rahul Desai, Tyler Zhu, Samyak Parajuli, Mike Guo, Dawn Song, Jacob
  Steinhardt, and Justin Gilmer,
\newblock ``The many faces of robustness: A critical analysis of
  out-of-distribution generalization,'' 2020.

\bibitem{imagenet-c}
Dan Hendrycks and Thomas Dietterich,
\newblock ``Benchmarking neural network robustness to common corruptions and
  perturbations,'' 2019.

\bibitem{imagenet-ao}
Dan Hendrycks, Kevin Zhao, Steven Basart, Jacob Steinhardt, and Dawn Song,
\newblock ``Natural adversarial examples,'' 2021.

\bibitem{michel2018mtnt}
Paul Michel and Graham Neubig,
\newblock ``{MTNT}: A testbed for {M}achine {T}ranslation of {N}oisy {T}ext,''
\newblock in {\em Proceedings of the 2018 Conference on Empirical Methods in
  Natural Language Processing (EMNLP)}, 2018.

\bibitem{sagawa2021extending}
Shiori Sagawa, Pang~Wei Koh, Tony Lee, Irena Gao, Sang~Michael Xie, Kendrick
  Shen, Ananya Kumar, Weihua Hu, Michihiro Yasunaga, Henrik Marklund, et~al.,
\newblock ``Extending the wilds benchmark for unsupervised adaptation,''
\newblock {\em arXiv preprint arXiv:2112.05090}, 2021.

\bibitem{filos2019systematic}
Angelos Filos, Sebastian Farquhar, Aidan~N. Gomez, Tim G.~J. Rudner, Zachary
  Kenton, Lewis Smith, Milad Alizadeh, Arnoud de~Kroon, and Yarin Gal,
\newblock ``A systematic comparison of bayesian deep learning robustness in
  diabetic retinopathy tasks,'' 2019.

\bibitem{nado2021uncertainty}
Zachary Nado, Neil Band, Mark Collier, Josip Djolonga, Michael~W Dusenberry,
  Sebastian Farquhar, Qixuan Feng, Angelos Filos, Marton Havasi, Rodolphe
  Jenatton, et~al.,
\newblock ``Uncertainty baselines: Benchmarks for uncertainty \& robustness in
  deep learning,''
\newblock {\em arXiv preprint arXiv:2106.04015}, 2021.

\bibitem{galthesis}
Yarin Gal,
\newblock {\em Uncertainty in Deep Learning},
\newblock Ph.D. thesis, University of Cambridge, 2016.

\bibitem{notinprincipled}
Pascal Notin, Jos{\'e}~Miguel Hern{\'a}ndez-Lobato, and Yarin Gal,
\newblock ``Principled uncertainty estimation for high dimensional data,''
\newblock in {\em Uncertainty \& Robustness in Deep Learning Workshop, ICML},
  2020.

\bibitem{malinin2021structured}
Andrey Malinin and Mark Gales,
\newblock ``Uncertainty estimation in autoregressive structured prediction,''
\newblock in {\em International Conference on Learning Representations}, 2021.

\bibitem{talarbelpropagating}
Raghav Mehta, Thomas Christinck, Tanya Nair, Aurlie Bussy, Swapna Premasiri,
  Manuela Costantino, M.~Mallar Chakravarthy, Douglas~L. Arnold, Yarin Gal, and
  Tal Arbel,
\newblock ``Propagating uncertainty across cascaded medical imaging tasks for
  improved deep learning inference,''
\newblock {\em IEEE Transactions on Medical Imaging}, vol. 41, no. 2, pp.
  360--373, 2022.

\bibitem{xiao2019wat}
Tim~Z Xiao, Aidan~N Gomez, and Yarin Gal,
\newblock ``Wat heb je gezegd? detecting out-of-distribution translations with
  variational transformers,''
\newblock in {\em Bayesian Deep Learning Workshop (NeurIPS)}, 2019.

\bibitem{filos2020can}
Angelos Filos, Panagiotis Tigkas, Rowan McAllister, Nicholas Rhinehart, Sergey
  Levine, and Yarin Gal,
\newblock ``Can autonomous vehicles identify, recover from, and adapt to
  distribution shifts?,''
\newblock in {\em International Conference on Machine Learning}. PMLR, 2020,
  pp. 3145--3153.

\bibitem{fomicheva2020unsupervised}
Marina Fomicheva, Shuo Sun, Lisa Yankovskaya, Fr{\'e}d{\'e}ric Blain, Francisco
  Guzm{\'a}n, Mark Fishel, Nikolaos Aletras, Vishrav Chaudhary, and Lucia
  Specia,
\newblock ``Unsupervised quality estimation for neural machine translation,''
\newblock {\em arXiv preprint arXiv:2005.10608}, 2020.

\bibitem{arjovsky2019invariant}
Martin Arjovsky, L{\'e}on Bottou, Ishaan Gulrajani, and David Lopez-Paz,
\newblock ``Invariant risk minimization,''
\newblock {\em arXiv preprint arXiv:1907.02893}, 2019.

\bibitem{ryabinin2021scaling}
Max Ryabinin, Andrey Malinin, and Mark Gales,
\newblock ``Scaling ensemble distribution distillation to many classes with
  proxy targets,''
\newblock {\em arXiv preprint arXiv:2105.06987}, 2021.

\bibitem{havasi2020training}
Marton Havasi, Rodolphe Jenatton, Stanislav Fort, Jeremiah~Zhe Liu, Jasper
  Snoek, Balaji Lakshminarayanan, Andrew~M. Dai, and Dustin Tran,
\newblock ``Training independent subnetworks for robust prediction,'' 2020.

\bibitem{walton2020rising}
Clare Walton, Rachel King, Lindsay Rechtman, Wendy Kaye, Emmanuelle Leray,
  Ruth~Ann Marrie, Neil Robertson, Nicholas La~Rocca, Bernard Uitdehaag, Ingrid
  van~der Mei, et~al.,
\newblock ``Rising prevalence of multiple sclerosis worldwide: Insights from
  the atlas of ms,''
\newblock {\em Multiple Sclerosis Journal}, vol. 26, no. 14, pp. 1816--1821,
  2020.

\bibitem{thompsonDiag2017}
Alan Thompson, Brenda Banwell, Frederik Barkhof, William Carroll, Timothy
  Coetzee, Giancarlo Comi, Jorge Correale, Franz Fazekas, Massimo Filippi, Mark
  Freedman, Kazuo Fujihara, Steven Galetta, Hans-Peter Hartung, Ludwig Kappos,
  Fred Lublin, Ruth Marrie, Aaron Miller, Debbie Miller, Xavier Montalban, and
  Jeffrey Cohen,
\newblock ``Diagnosis of multiple sclerosis: 2017 revisions of the mcdonald
  criteria,''
\newblock {\em The Lancet Neurology}, vol. 17, 12 2017.

\bibitem{zengReview2020}
Chenyi Zeng, Lin Gu, Zhenzhong Liu, and Shen Zhao,
\newblock ``Review of deep learning approaches for the segmentation of multiple
  sclerosis lesions on brain mri,''
\newblock {\em Frontiers in Neuroinformatics}, vol. 14, 2020.

\bibitem{roviraEvidence2015}
Alex Rovira, Mike Wattjes, Mar Tintorè, Carmen Tur, Tarek Yousry, Maria~Pia
  Sormani, Nicola De~Stefano, Massimo Filippi, Cristina Auger, Mara Rocca,
  Frederik Barkhof, Franz Fazekas, Ludwig Kappos, Chris Polman, David Miller,
  Xavier Montalban, and Jette Frederiksen,
\newblock ``Evidence-based guidelines: Magnims consensus guidelines on the use
  of mri in multiple sclerosis - clinical implementation in the diagnostic
  process,''
\newblock {\em Nature reviews. Neurology}, vol. 11, 07 2015.

\bibitem{WATTJES2021653}
Mike~P Wattjes, Olga Ciccarelli, Daniel~S Reich, Brenda Banwell, Nicola {de
  Stefano}, Christian Enzinger, Franz Fazekas, Massimo Filippi, Jette
  Frederiksen, Claudio Gasperini, Yael Hacohen, Ludwig Kappos, David K~B Li,
  Kshitij Mankad, Xavier Montalban, Scott~D Newsome, Jiwon Oh, Jacqueline
  Palace, Maria~A Rocca, Jaume Sastre-Garriga, Mar Tintoré, Anthony
  Traboulsee, Hugo Vrenken, Tarek Yousry, Frederik Barkhof, Àlex Rovira,
  Mike~P Wattjes, Olga Ciccarelli, Nicola {de Stefano}, Christian Enzinger,
  Franz Fazekas, Massimo Filippi, Jette Frederiksen, Claudio Gasperini, Yael
  Hacohen, Ludwig Kappos, Kshitij Mankad, Xavier Montalban, Jacqueline Palace,
  María~A Rocca, Jaume Sastre-Garriga, Mar Tintore, Hugo Vrenken, Tarek
  Yousry, Frederik Barkhof, Alex Rovira, David K~B Li, Anthony Traboulsee,
  Scott~D Newsome, Brenda Banwell, Jiwon Oh, Daniel~S Reich, Daniel~S Reich,
  and Jiwon Oh,
\newblock ``2021 magnims–cmsc–naims consensus recommendations on the use of
  mri in patients with multiple sclerosis,''
\newblock {\em The Lancet Neurology}, vol. 20, no. 8, pp. 653--670, 2021.

\bibitem{CARASS201777}
Aaron Carass, Snehashis Roy, Amod Jog, Jennifer~L. Cuzzocreo, Elizabeth
  Magrath, Adrian Gherman, Julia Button, James Nguyen, Ferran Prados, Carole~H.
  Sudre, Manuel {Jorge Cardoso}, Niamh Cawley, Olga Ciccarelli, Claudia~A.M.
  Wheeler-Kingshott, Sébastien Ourselin, Laurence Catanese, Hrishikesh
  Deshpande, Pierre Maurel, Olivier Commowick, Christian Barillot, Xavier
  Tomas-Fernandez, Simon~K. Warfield, Suthirth Vaidya, Abhijith Chunduru,
  Ramanathan Muthuganapathy, Ganapathy Krishnamurthi, Andrew Jesson, Tal Arbel,
  Oskar Maier, Heinz Handels, Leonardo~O. Iheme, Devrim Unay, Saurabh Jain,
  Diana~M. Sima, Dirk Smeets, Mohsen Ghafoorian, Bram Platel, Ariel Birenbaum,
  Hayit Greenspan, Pierre-Louis Bazin, Peter~A. Calabresi, Ciprian~M.
  Crainiceanu, Lotta~M. Ellingsen, Daniel~S. Reich, Jerry~L. Prince, and
  Dzung~L. Pham,
\newblock ``Longitudinal multiple sclerosis lesion segmentation: Resource and
  challenge,''
\newblock {\em NeuroImage}, vol. 148, pp. 77--102, 2017.

\bibitem{carass2017}
Aaron Carass, Snehashis Roy, Amod Jog, Jennifer~L. Cuzzocreo, Elizabeth
  Magrath, Adrian Gherman, Julia Button, James Nguyen, Pierre-Louis Bazin,
  Peter~A. Calabresi, and et~al.,
\newblock ``Longitudinal multiple sclerosis lesion segmentation data
  resource,''
\newblock {\em Data in Brief}, vol. 12, pp. 346–350, 2017.

\bibitem{commowick2018}
Olivier Commowick, Audrey Istace, Michaël Kain, Baptiste Laurent, Florent
  Leray, Mathieu Simon, Sorina Pop, Pascal Girard, Roxana Ameli,
  Jean-Christophe Ferré, Anne Kerbrat, Thomas Tourdias, Frederic Cervenansky,
  Tristan Glatard, Jeremy Beaumont, Senan Doyle, Florence Forbes, Jesse Knight,
  April Khademi, and Christian Barillot,
\newblock ``Objective evaluation of multiple sclerosis lesion segmentation
  using a data management and processing infrastructure,''
\newblock {\em Scientific Reports}, vol. 8, pp. 13650--13666, 09 2018.

\bibitem{Lesjak2017ANP}
Ziga Lesjak, Alfiia Galimzianova, Ales Koren, Matej Lukin, Franjo Pernus,
  Bostjan Likar, and Žiga piclin,
\newblock ``A novel public mr image dataset of multiple sclerosis patients with
  lesion segmentations based on multi-rater consensus,''
\newblock {\em Neuroinformatics}, vol. 16, pp. 51--63, 2017.

\bibitem{coupe2008}
Pierrick Coupé, Pierre Yger, Sylvain Prima, Pierre Hellier, Charles Kervrann,
  and Christian Barillot,
\newblock ``An optimized blockwise nonlocal means denoising filter for 3-d
  magnetic resonance images,''
\newblock {\em IEEE Transactions on Medical Imaging}, vol. 27, 05 2008.

\bibitem{isensee2019}
Fabian Isensee, Marianne Schell, Irada Pflueger, Gianluca Brugnara, David
  Bonekamp, Ulf Neuberger, Antje Wick, Heinz-Peter Schlemmer, Sabine Heiland,
  Wolfgang Wick, Martin Bendszus, Klaus Maier-Hein, and Philipp Kickingereder,
\newblock ``Automated brain extraction of multisequence mri using artificial
  neural networks,''
\newblock {\em Human Brain Mapping}, vol. 40, 08 2019.

\bibitem{commowick2012}
Olivier Commowick, Nicolas Wiest-Daesslé, and Sylvain Prima,
\newblock ``Block-matching strategies for rigid registration of multimodal
  medical images,''
\newblock in {\em 2012 9th IEEE International Symposium on Biomedical Imaging
  (ISBI)}, 2012, pp. 700--703.

\bibitem{tustison2010}
Nicholas~J Tustison, Brian~B Avants, Philip~A Cook, Yuanjie Zheng, Alexander
  Egan, Paul~A Yushkevich, and James~C Gee,
\newblock ``N4itk: Improved n3 bias correction,''
\newblock {\em IEEE Transactions on Medical Imaging}, vol. 29, no. 6, pp.
  1310–1320, 2010.

\bibitem{Dice1945MeasuresOT}
Lee~Raymond Dice,
\newblock ``Measures of the amount of ecologic association between species,''
\newblock {\em Ecology}, vol. 26, pp. 297--302, 1945.

\bibitem{Srensen1948AMO}
Tage S{\o}rensen, Tage S{\o}rensen, Tor Biering-S{\o}rensen, Tia S{\o}rensen,
  and John~T. Sorensen,
\newblock ``A method of establishing group of equal amplitude in plant
  sociobiology based on similarity of species content and its application to
  analyses of the vegetation on danish commons,''
\newblock 1948.

\bibitem{reinke2021common}
Annika Reinke, Matthias Eisenmann, Minu~D Tizabi, Carole~H Sudre, Tim
  R{\"a}dsch, Michela Antonelli, Tal Arbel, Spyridon Bakas, M~Jorge Cardoso,
  Veronika Cheplygina, et~al.,
\newblock ``Common limitations of image processing metrics: A picture story,''
\newblock {\em arXiv preprint arXiv:2104.05642}, 2021.

\bibitem{iek20163DUL}
{\"O}zg{\"u}n Çiçek, Ahmed Abdulkadir, Soeren~S. Lienkamp, Thomas Brox, and
  Olaf Ronneberger,
\newblock ``3d u-net: Learning dense volumetric segmentation from sparse
  annotation,''
\newblock {\em ArXiv}, vol. abs/1606.06650, 2016.

\bibitem{la2020multiple}
Francesco La~Rosa, Ahmed Abdulkadir, M{\'a}rio~Jo{\~a}o Fartaria, Reza
  Rahmanzadeh, Po-Jui Lu, Riccardo Galbusera, Muhamed Barakovic, Jean-Philippe
  Thiran, Cristina Granziera, and Merixtell~Bach Cuadra,
\newblock ``Multiple sclerosis cortical and wm lesion segmentation at 3t mri: a
  deep learning method based on flair and mp2rage,''
\newblock {\em NeuroImage: Clinical}, vol. 27, pp. 102335, 2020.

\bibitem{unetr2021}
Ali Hatamizadeh, Dong Yang, Holger Roth, and Daguang Xu,
\newblock ``Unetr: Transformers for 3d medical image segmentation,'' 03 2021.

\bibitem{christodoulou2019sustainable}
Anastasia Christodoulou and Johan Woxenius,
\newblock ``Sustainable short sea shipping,'' 2019.

\bibitem{hilakari2019carbon}
Marianna Hilakari,
\newblock ``Carbon footprint calculation of shipbuilding,''
\newblock 2019.

\bibitem{gkerekos2019machine}
Christos Gkerekos, Iraklis Lazakis, and Gerasimos Theotokatos,
\newblock ``Machine learning models for predicting ship main engine fuel oil
  consumption: A comparative study,''
\newblock {\em Ocean Engineering}, vol. 188, pp. 106282, 2019.

\bibitem{zhu2020predicting}
Yongjie Zhu, Yi~Zuo, and Tieshan Li,
\newblock ``Predicting ship fuel consumption based on lstm neural network,''
\newblock in {\em 2020 7th International Conference on Information,
  Cybernetics, and Computational Social Systems (ICCSS)}. IEEE, 2020, pp.
  310--313.

\bibitem{gebru2018datasheets}
Timnit Gebru, Jamie Morgenstern, Briana Vecchione, Jennifer~Wortman Vaughan,
  Hanna Wallach, Hal Daum{\'e}~III, and Kate Crawford,
\newblock ``Datasheets for datasets,''
\newblock {\em arXiv preprint arXiv:1803.09010}, 2018.

\bibitem{py06nimg}
Paul Yushkevich, Joseph Piven, Heather~Cody Hazlett, Rachel~Gimpel Smith, Sean
  Ho, James~C. Gee, and Guido Gerig,
\newblock ``User-guided {3D} active contour segmentation of anatomical
  structures: Significantly improved efficiency and reliability,''
\newblock {\em Neuroimage}, vol. 31, no. 3, pp. 1116--1128, 2006.

\bibitem{hullpic2022}
Efthymia Tsompopoulou, Andreas Athanassopoulos, Elli Sivena, Kyriakos
  Polymenakos, Vasileios Tsarsitalidis, Antonis Nikitakis, and Konstantinos
  Kyriakopoulos,
\newblock ``{On the Evaluation of Uncertainty of AI models for Ship Powering
  and its Effect on Power Estimates for Non-ideal Conditions},''
\newblock in {\em HullPIC}, 2022.

\bibitem{bose2008marine}
Neil Bose,
\newblock ``Marine powering prediction and propulsors,''
\newblock {\em Published by: The Society of Naval Architects and Marine
  Engineers, SNAME, ISBN: 0-939773-65-1}, 2008.

\bibitem{van1969wageningen}
WPA Van~Lammeren, JD~van van Manen, and MWC Oosterveld,
\newblock ``The wageningen b-screw series,''
\newblock 1969.

\bibitem{holtrop1982approximate}
J~Holtrop and GGJ Mennen,
\newblock ``An approximate power prediction method,''
\newblock {\em International Shipbuilding Progress}, vol. 29, no. 335, pp.
  166--170, 1982.

\bibitem{nikolopoulos2019study}
Lampros Nikolopoulos and Evangelos Boulougouris,
\newblock ``A study on the statistical calibration of the holtrop and mennen
  approximate power prediction method for full hull form, low froude number
  vessels,''
\newblock {\em Journal of Ship Production and Design}, vol. 35, no. 01, pp.
  41--68, 2019.

\bibitem{ships2015marine}
ISO,
\newblock ``Marine technology—guidelines for the assessment of speed and
  power performance by analysis of speed trial data,''
\newblock {\em ISO: Geneva, Switzerland}, 2015.

\bibitem{fujiwara2006cruising}
Toshifumi Fujiwara, Michio Ueno, and Yoshiho Ikeda,
\newblock ``Cruising performance of a large passenger ship in heavy sea,''
\newblock in {\em The Sixteenth International Offshore and Polar Engineering
  Conference}. OnePetro, 2006.

\bibitem{tsujimoto2008practical}
Masaru Tsujimoto, Kazuya Shibata, Mariko Kuroda, and Ken Takagi,
\newblock ``A practical correction method for added resistance in waves,''
\newblock {\em Journal of the Japan Society of Naval Architects and Ocean
  Engineers}, vol. 8, pp. 177--184, 2008.

\bibitem{carlton2018marine}
John Carlton,
\newblock {\em Marine propellers and propulsion},
\newblock Butterworth-Heinemann, 2018.

\bibitem{bertram2012chapter}
Volker Bertram,
\newblock ``Practical ship hydrodynamics,''
\newblock 2012.

\bibitem{townsin1981estimating}
RL~Townsin, D~Byrne, TE~Svensen, and A~Milne,
\newblock ``Estimating the technical and economic penalties of hull and
  propeller roughness,''
\newblock {\em Trans. SNAME}, vol. 89, pp. 295--318, 1981.

\bibitem{seo2016study}
Kwang-Cheol Seo, Mehmet Atlar, and Bonguk Goo,
\newblock ``A study on the hydrodynamic effect of biofouling on marine
  propeller,''
\newblock {\em Journal of the Korean Society of Marine Environment \& Safety},
  vol. 22, no. 1, pp. 123--128, 2016.

\bibitem{farkas2020impact}
Andrea Farkas, Nastia Degiuli, Ivana Marti{\'c}, and Roko Dejhalla,
\newblock ``Impact of hard fouling on the ship performance of different ship
  forms,''
\newblock {\em Journal of Marine Science and Engineering}, vol. 8, no. 10, pp.
  748, 2020.

\bibitem{tsamoura2021neural}
Efthymia Tsamoura, Timothy Hospedales, and Loizos Michael,
\newblock ``Neural-symbolic integration: A compositional perspective,''
\newblock in {\em Proceedings of the AAAI Conference on Artificial
  Intelligence}, 2021, vol.~35, pp. 5051--5060.

\bibitem{wen2018flipout}
Yeming Wen, Paul Vicol, Jimmy Ba, Dustin Tran, and Roger Grosse,
\newblock ``{Flipout: Efficient Pseudo-Independent Weight Perturbations on
  Mini-Batches},''
\newblock in {\em International Conference on Learning Representations}, 2018.

\end{thebibliography}

\newpage
\appendix
\appendixpage
\section{Shifts Dataset General Datasheet}\label{apn:datasheet}

Here we describe the motivation, uses, distribution as well as the maintenance and support plan for the Shifts 2.0 Dataset in the \emph{datasheet for datasets} format~\cite{gebru2018datasheets}. The details of the composition, collection and pre-prossessing of each component dataset are provided in appendices~\ref{apn:med}-\ref{apn:ship}

\paragraph{Motivation} The primary goal for the creation of the Shifts 2.0 Dataset was the evaluation of uncertainty quantification models and robustness to distributional shift on industrial and medical tasks of large practical and societal importance. These datasets span multiple modalities and feature real examples of distributional shift. Making these datasets available allows models' robust generalisation and uncertainty quality to be assessed - something not possible with standard in-domain benchmarks. Furthermore, by construction a dataset using real medical or industrial tasks, the any insights reached can be directly applied without the need for adaptation. This is an important feature, as most novel ML methods fail at the stage of adaptation and scaling to actual applications. 

\paragraph{Uses} The dataset is used as part of the Shifts Challenge 2.0, which is organized around this dataset\footnote{ \url{https://shifts.ai}}. The Shifts Challenge consists of two tracks organized around each of the constituent datasets within Shifts 2.0 . The dataset, baseline models and code to reproduce it all is provided in a GitHub repository\footnote{\url{https://github.com/Shifts-Project/shifts}}. Other than uncertainty and robustness research the dataset could be used for developing better models for each of the separate tasks - WM MS lesion segmentation and margine cargo vessel power estimation.

\paragraph{Distribution}
It is our intention that the Shifts 2.0 dataset be freely available for research purposes. All the code is available under an open-source Apache 2.0 licence. 

The Shifts cargo vessel power estimation datasets is distributed by DeepSea under an open-source CC BY NC SA 4.0 license. The training and in-domain/shifted devevelopment sets, both with real and synthetic targets, will be freely distributed via the Zenodo platform. The evaluation sets will not be released, but will be hosted on permanent leaderboards on the Grand-Challenge platform~\footnote{\url{https://grand-challenge.org}}. Should the leaderboards close for any reason, the evaluation sets will be similarly released via Zenodo. The reason for keeping the evaluation sets private is to ensure a truly clean 'out-of-domain generalisation scenario' and avoid any possible, even unintentional, data leakage.

The MS lesion segmentation dataset has a more complex structure. Part of the dataset (ISBI train set and PubMRI) is shared under a permissive CC BY NC SA 4.0 license. These components will be hosted on Zenodo. However, the MSSEG-1 component~\cite{commowick2018} was only available via credentialized access via the Shanoir Platform~\footnote{\url{https://project.inria.fr/shanoir/}} under an OFSEP DUA. Getting this access to some time. However, we have reached an agreement with OFSEP to allow us to host our copy of the MSSEG-1 data on Zenodo under their DUA to facilitate faster and simpler credentialized access within a consistent, pre-processed data format. Thus, the in-domain training, dev and eval as well as the shifted dev set will be available for download from Zenodo. The data will be split into two archives - the MSSEG archive, which will require credentialized access which will be fast to achieve, and the remaining data, which will be freely hosted under a permissive CC BY NC SA 4.0 license. Researchers wishing to use the dataset will need to download both archives and then follow the included instructions to combine the two archives into the canonical splits we have defined. 

Finally, the dataset sourced at Lausanne, which is used as the Shifted evaluation set, was collected in such a way that sharing the dataset itself is not possible, even via credentialized access. Specifically, patients have the right to withdraw their data from the dataset at any time - the only way to ensure this is for the data collectors to maintain both ownership and control over the dataset. However, the data owners (who are also authors on this paper) are happy to freely allow researchers to evaluate their models on this data via dockers on a public leaderboard, which will be hosted in Grand-Challenge.

\paragraph{Maintenance}
The dataset is being actively maintained by the Shifts Project\footnote{This is an international collaboration of researchers studying distributional shift. Website will be launched shortly}. The team can be contacted by raising an issue on GitHub and by writing to the first author of this paper. The dataset will be hosted on the Zenodo\footnote{\url{https://zenodo.org}} storage platform and will be hosted there permanently for the foreseeable future. The dataset can be updated at the discretion of the dataset creators, though regular updates are not planned. Updates which expand the evaluation sets or add new ones will mean that the previous dev/eval sets are supported. Updates which fix errors in dev/eval sets mean that the prior ones are obsolete and unsupported. If any update is to occur, we will make an announcement via GitHub, twitter, and the Shifts Project mailing list.

We do not allow other parties to update the Shifts Dataset. However, any issues found can be logged by raising an issue on GitHub or contacting the first author of this paper so that they can be addressed. For the Shifts datasets that are released under an open-source CC BY NC SA 4.0 license which allows modifications, we are happy for people to create derivative datasets using ours, provided the modifications are documented and the original dataset referenced. For the medical data, where licensing is more complicated, users need to be aware of the exact license on the data component being used and what is allowed. 

\paragraph{Societal Consequences} Research on uncertainty estimation and robustness aims to make AI safer and more reliable, and therefore has limited negative societal consequences overall. As discussed in sections~\ref{sec:wml} and~\ref{sec:ship}, both tasks considered for Shifts 2.0 have high societal importance. High-quality automatic segmentation of MS lesions can enable greater patient throughput, more regular checkups, and in the long term, a more personalised treatment plan. Similarly, accurate and reliable cargo vessel power consumption estimation can help optimize fuel usage, carry less surplus fuel and thereby decrease both the cost of marine cargo transport as well as it's climate impact.

\paragraph{Guidelines for Ethical Use}
Users of this dataset are encouraged to use it for the purpose of improving the reliability and safety of large-scale applications of machine learning. Furthermore, we encourage users of our dataset to develop compute and memory efficient methods for improving safety and reliability. 

As part of this data features 3D MRI brain scans taken from MS patients, users of this dataset should not attempt to establish or retrieve the identity of the patients. Furthermore, users should not link this data to any other database in a way that could provide identifying information. Users similarly should not request the pseudonymisation key that would link this data to an individual's personal information. When sharing secondary or derivative data (e.g. group statistical maps, learnt models, etc...), users should only do so if they are on a group level, and information from individual participants cannot be deduced.

\paragraph{Responsibility} The authors confirm that, to the best of our knowledge, the released dataset does not violate any prior licenses or rights. However, if such a violation were to exist, we are responsible for resolving this issue.

\begin{itemize}
    \item Yandex, DeepSea, CUED Speech Group (Cambridge), ThINk Basel, Medical Image Analysis Laboratory (MIAL) Unil, MedGIFT (HES-SO) employees can participate in the competition, but are ineligible for prizes and will be excluded from the final leaderboard.
    \item In order to participate in the competition, participants must register by accepting these Terms as described on the Competition Website.
    \item Participants can work individually or in teams. Participants who work together on the same method are considered to be the same team. It is responsibility of participants to declare their team.
    \item Participants can work on both competition tracks. Participation in any track is \textbf{not} mutually exclusive with participation the other track.
    \item Participants should submit docker files containing their solution to evaluated on development and evaluation data, which will be used to update the respective leader boards. Participants' models should, in addition to making prediction, also yield a measure of uncertainty to accompany each prediction. Details of what to submit and how it should be submitted will be provided on the competition website.
    \item Participants should download the datasets from either the Data Provider’s website or from the Competition Website, and accept the applicable terms and conditions. The Organizer will specify the links to the data on the Competition Website.
    \item The participants should build models only the on the data provided for this competition by the Organizers. The use of any external data for any of the three competition tracks is \textbf{forbidden}. However, intelligent data augmentation is allowed and encouraged.
    \item Models must run on, at most, 1 RTX 2080 Ti GPU with 12 GB of graphics memory and yield predictions within 800ms per 1 input sample. Submitted solutions which break these limitations will not be considered in the final scoring. 
    \item To be eligible for prizes, the top-3 participants in each track are required to submit:
    \begin{enumerate}
        \item A detailed technical report on their solution.
        \item Their code used to train the model -- code \textbf{must be reproducible by organizers}.
        \item The model which corresponds to the submission.
        \item Performance on evaluation data will be verified using the provided code, models, and technical report after the end of the competition during a verification stage. Only submissions that are verified to be consistent with submitted scores will eligible for prizes. Organizers will reach out to participants in case difficulties arise.
    \end{enumerate}
    \item During the development phase, we will limit the number of submissions on the development data to 20, per track, per team.
    \item During the evaluation phase, we will limit the number of submissions on the evaluation data to 3, per track, per team.
    \item Organizers have the right to update the rules in unknown situations (e.g., a tie among participants ranking) or lead the competition in a way that best suits the goals.
    \item Sponsor reserves the right to modify the leaderboard at its discretion for reasons of fairness, proper play, and/or compliance with these Official Rules, including, for example, if Sponsor believes that participants or entries shown on the leaderboard do not meet any eligibility or other requirements in these Official Rules.
\end{itemize}

\section{Assessment Metrics}\label{apn:metrics}

As discussed in Section \ref{sec:paradigm}, in this work we consider robustness and uncertainty estimation to be two equally important factors in assessing the reliability of a model. We assume that as the degree of distributional shift increases, so should a model's errors; in other words, a model's uncertainty estimates should be correlated with the degree of its error. This informs our choice of assessment metrics, which must \emph{jointly} assess robustness and uncertainty estimation. 

One standard approach to jointly assess robustness and uncertainty are \emph{error-retention curves}~\cite{malinin-thesis,deepensemble2017}, which plot a model's mean error over a dataset, as measured using a metric such as error-rate, MSE, or nDSC, with respect to the fraction of the dataset for which the model's predictions are used. These retention curves are traced by replacing a model's predictions with ground-truth labels obtained from an oracle in order of \emph{decreasing uncertainty}, thereby decreasing error. Ideally, a model's uncertainty is correlated with its error, and therefore the most errorful predictions would be replaced first, which would yield the greatest reduction in mean error as more predictions are replaced. This represents a hybrid human-AI scenario, where a model can consult an oracle (human) for assistance in difficult situations and obtain from the oracle a perfect prediction on those examples.

The area under the retention curve (R-AUC) is a metric for jointly assessing robustness to distributional shift and the quality of the uncertainty estimates. R-AUC can be reduced either by improving the predictions of the model, such that it has lower overall error at any given retention rate, or by providing estimates of uncertainty which better correlate with error, such that the most incorrect predictions are rejected first. It is important that the dataset in question contains both a subset ``matched'' to the training data, and a distributionally shifted subset.

\begin{figure}[htbp!]
     \centering
     \subfigure[Example]{\includegraphics[width=0.48\textwidth]{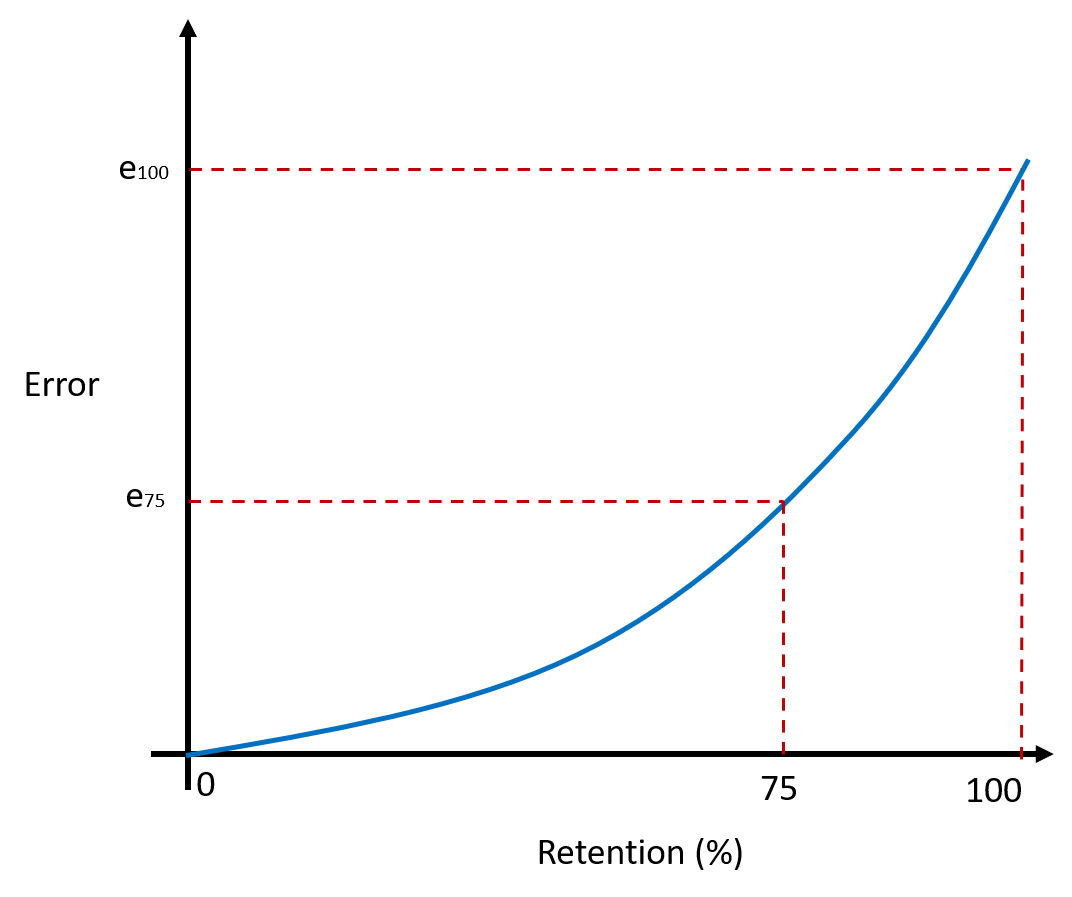}}
     \subfigure[Robustness]{ \includegraphics[width=0.48\textwidth]{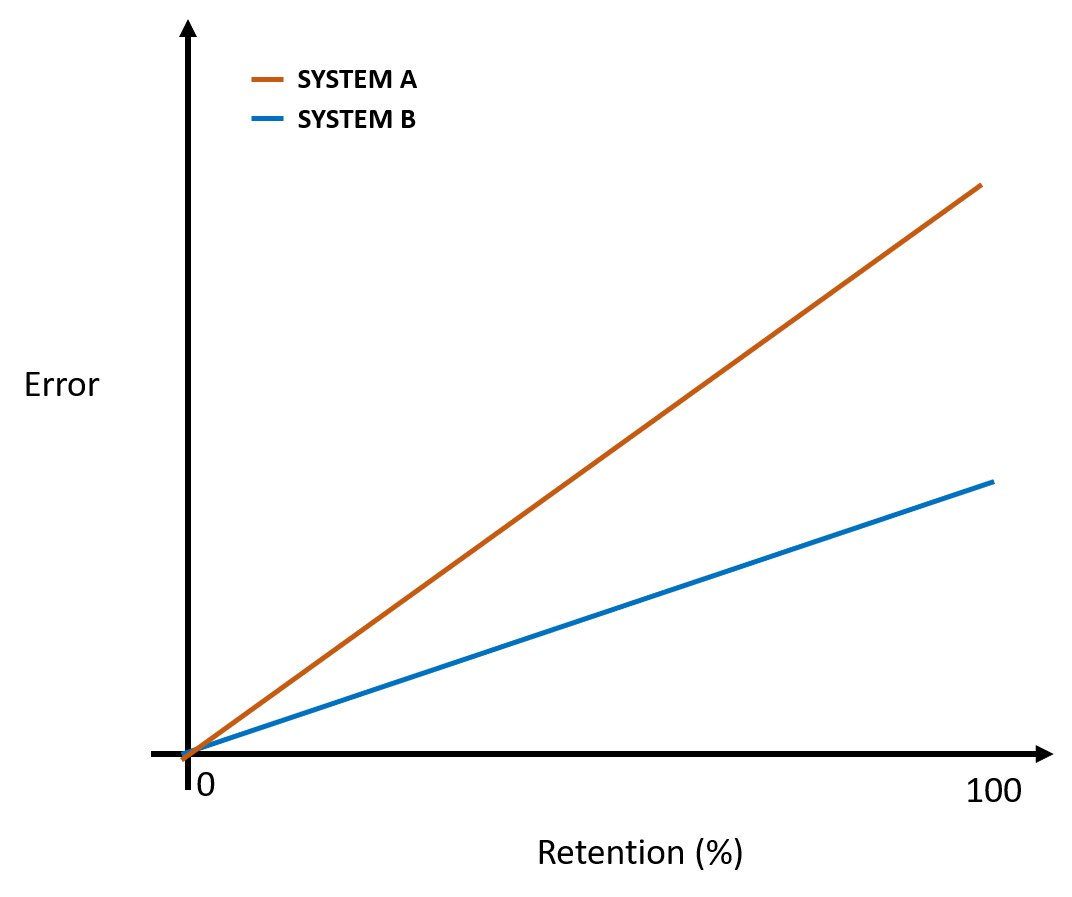}}
     \\
     \subfigure[Uncertainty]{ \includegraphics[width=0.48\textwidth]{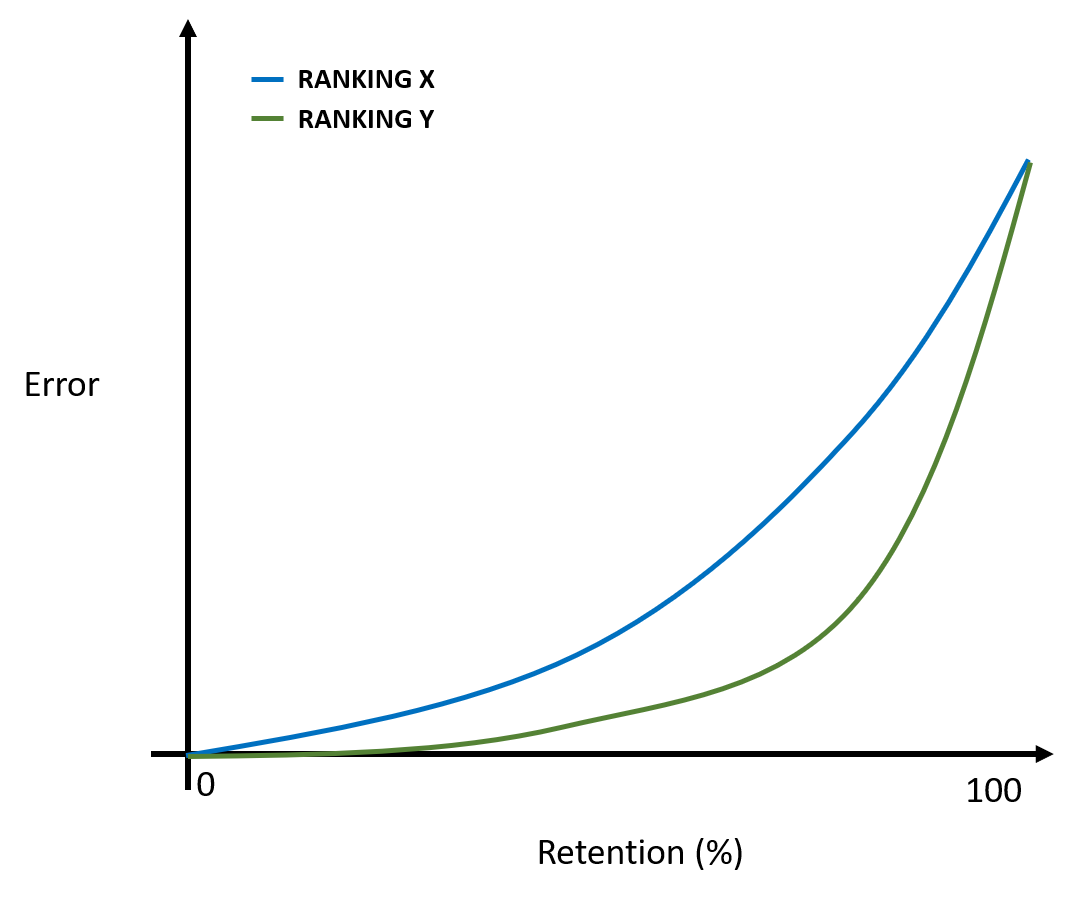}}
     \subfigure[Mixture]{ \includegraphics[width=0.48\textwidth]{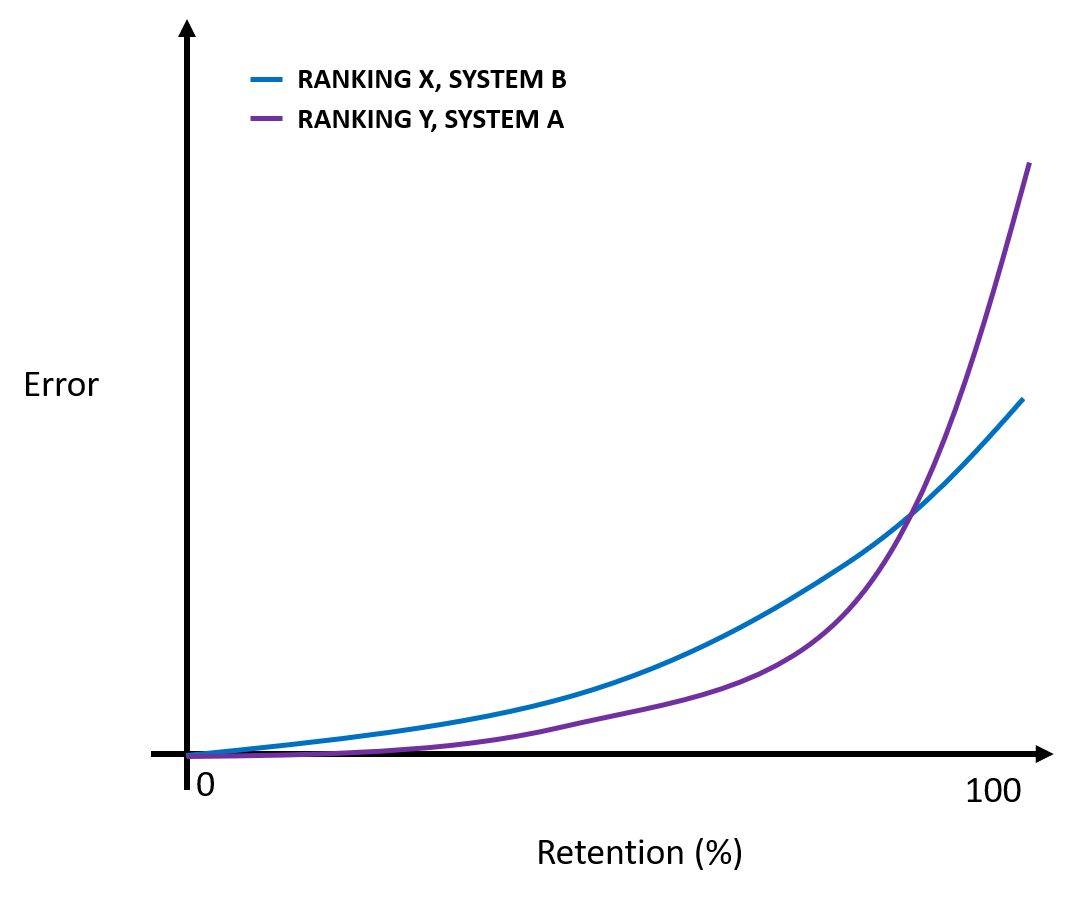}}
     \caption{Schematic explanation of error retention curves.}
     \label{fig:schematic_retention_curves}
\end{figure}
Schematic explanations of error-retention curves are given in Figure \ref{fig:schematic_retention_curves}, which demonstrates how these curves jointly assess robustness and uncertainty by measuring the area under such curves. Consider Figure \ref{fig:schematic_retention_curves}a. Here we can see that replacing a certain percentage of the models predictions with ground truth labels will decrease the error rate. Specifically, $e_{100}$ is the performance of the system using all the data while $e_{75}$ is the error of the system using the top 75\% of the data with the rejected data set to the ground-truth. Now consider Figure \ref{fig:schematic_retention_curves}b, where we demonstrate robustness. Here we plot retention curves for two systems, where one system is broadly more robust than the other. Predictions are rejected in a random, uninformative order, yielding a straight line. Here we can see that the more robust system (System B) will have a lower area under the retention curve (R-AUC) than the less robust system A. Now consider Figure \ref{fig:schematic_retention_curves}c, where we demonstrate uncertainty quality. For the same system, two different uncertainty approaches are consider where the uncertainty measure Y produces a better ranking which is more strongly correlated with the degree of error than uncertainty measure X. As a result, the largest errors are rejected first. Thus, the area under the retention curve constructed using the ranking defined by measure Y is smaller than under the retention curve defined using measure X. Finally, let's consider Figure \ref{fig:schematic_retention_curves}d, where we show a \emph{joint} assessment of robustness and uncertainty. Here, despite having worse predictive robustness, system A has a better uncertainty ranking measure, leading the a smaller R-AUC. Thus, this model is capable of achieving more operating points where it has lower error than system A, and is therefore better in terms of joint assessment of robustness and uncertainty. The converse scenario can also occur - we could have a more which is so robust, despite uninformative uncertainty, that is achieves superior performance at all retention percentages than a less robust model with informative uncertainty estimates. 
\begin{figure}[htbp!]
     \centering
     \subfigure[MS Segmentation]{\includegraphics[width=0.49\textwidth]{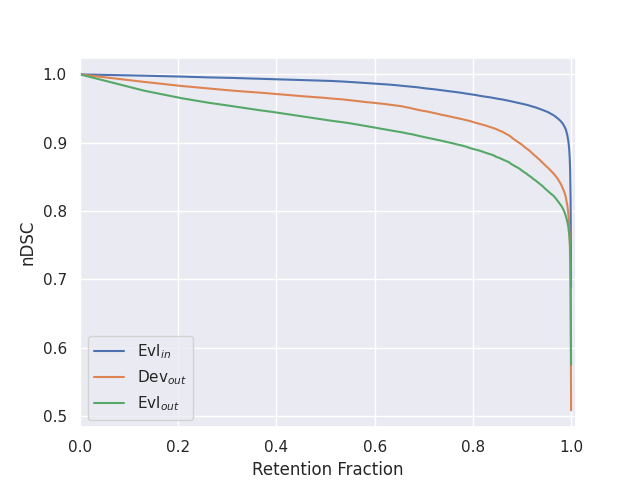}}
     \subfigure[Power Estimation]{ \includegraphics[width=0.49\textwidth]{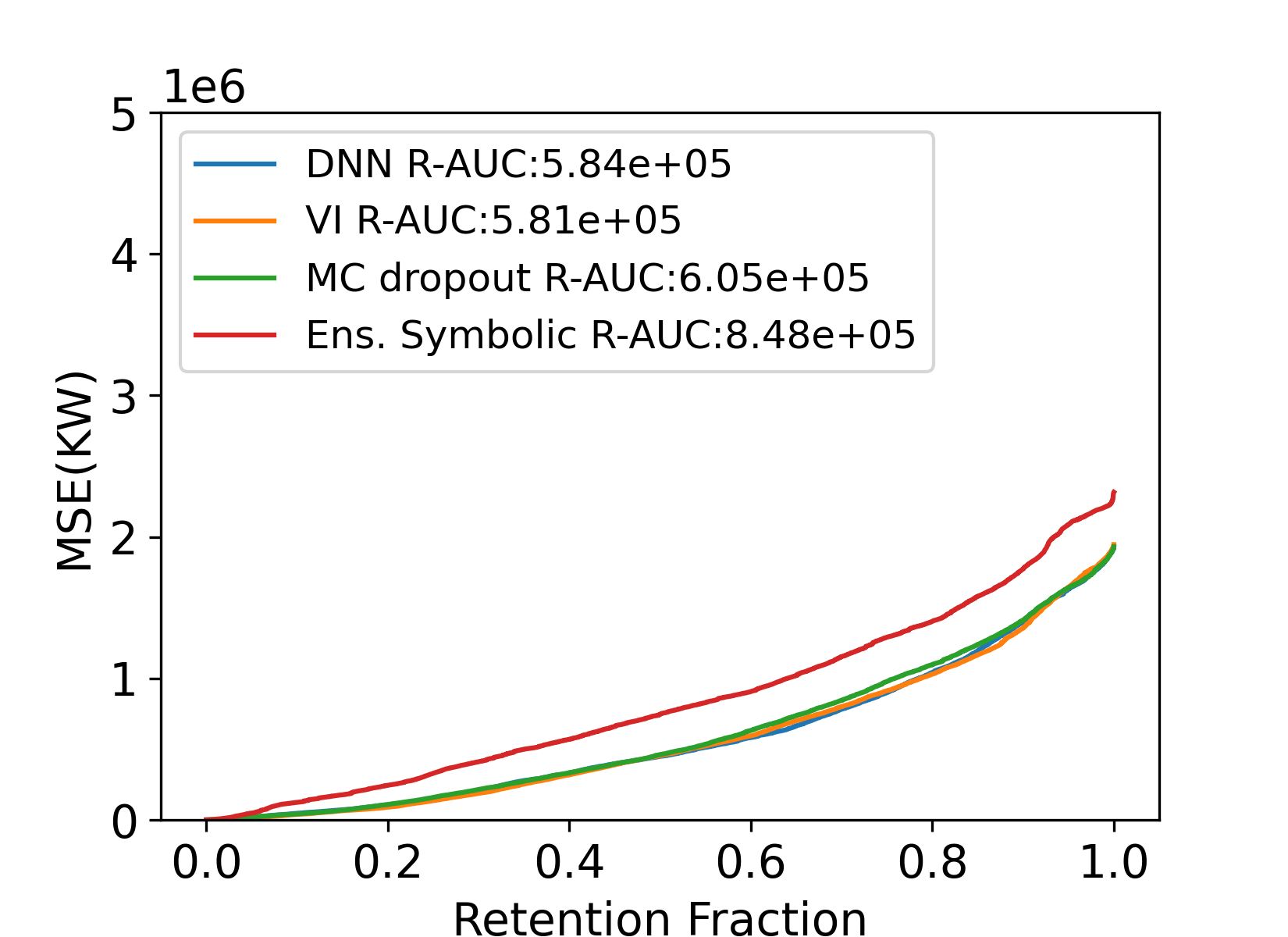}}
     \caption{Example error retention curves for the two tasks of the Shifts 2.0 Dataset.}\label{fig:example_retention_curves}
\end{figure}

In addition to area under an error-retention curve, we also consider an F1-retention curve, which is broadly similar, but uses the notion of `acceptable' error to assess whether uncertainty estimates can be used to detect `un-acceptable errors'. The metric is less susceptible to errors at the level of noise, but it is not always possible to define what is an 'acceptable error'. Thus, this metric is only used to assess the power estimation tasks, but not the segmentation task. For a detailed descriptions of the F1-retention curve, please see~\cite{malinin2021shifts}. 

The area under the error-retention curve and F1-retention curve is a \emph{summary statistic} which describes possible \emph{operating points}. We can specify a particular operating point, such as 95\% retention, and evaluate the error or F1 at that point for comparison. This is also an important figure, as all models work at a particular operating point which satisfies task-specific desiderata.

\section{Lesion Segmentation}\label{apn:med}

The current appendix provides further details on the Medical data collected as well as more complete set of baseline results.

\subsection{Dataset Description}\label{app:lesion_canon}

\paragraph{Canonical Dataset Construction} Here, we detail additional experiments using a UNET model that were run in order to select the canonical partitioning of the data. For both Tables \ref{tab:results_individualLocations} and \ref{tab:nfold}, ensembles of size 5 were used. First, in Table \ref{tab:results_individualLocations}, models are trained on data from each location and evaluated on all other location to identify the greatest shifts. Three different approaches to choosing the classification threshold were examined - in-domain on the corresponding dev set of the location, on the train and dev sets of the different location (not used for training), and finally on the actual test-set of each location. These different threshold tuning strategies allow us to examine the range of expected and upper bound performance on each location.
\begin{table}[ht]
\centering
\begin{small}
    \begin{tabular}{ll|cccccc}
    \toprule
Thresholding & Train & Rennes &  Bordeaux & Lyon & Ljubljana & Best & Lausanne \\
\midrule
% \multirow{5}*{$\tau=0.3$}
% & Rennes & 62.93 & 65.37 & 56.08 & 45.95 & 57.23 & 51.35 \\
% & Bordeaux & 45.04 & 71.68 & 49.69 & 25.46 & 40.41 & 41.45 \\
% & Lyon & 62.67 & 71.83 & 64.70 & 46.81 & 56.81 & 55.66 \\
% & Ljubljana & 68.23 & 68.95 & 62.30 & 59.45 & 61.71 & - \\
% & Best & 61.19 & 70.36 & 63.67 & 48.58 & 64.43 & - \\
% \midrule
\multirow{5}*{In-domain Dev}
& Rennes & 50.51 & \textbf{72.95} & 54.81 & 35.78 & 47.05 & 40.63 \\
& Bordeaux & 49.46 & 68.18 & 55.12 & 34.70 & 50.13 & 46.71 \\
& Lyon & 58.73 & 69.75 & \textbf{66.68} & 42.51 & 54.84 &  52.00 \\
& Ljubljana & \textbf{66.18} & 70.29 & 65.98 & \textbf{57.03} & \textbf{63.45} & 62.12 \\
& Best & 59.03 & 71.28 & 63.93 & 46.95 & 63.27 & 55.74 \\
\midrule
\multirow{5}*{Out-domain Train + Dev}
& Rennes & 57.70 & 67.91 & 59.38 & 47.37 & 56.26 & - \\
& Bordeaux & 50.90 & \textbf{71.80} & 56.96 & 34.70 & 50.13 & - \\
& Lyon & 65.23 & 71.65 & \textbf{69.00} & 52.44 & 55.18 & - \\
& Ljubljana & \textbf{66.91} & 70.00 & 66.67 & \textbf{59.03} & 60.85 & -  \\
& Best & 60.54 & 71.07 & 64.17 & 48.09 & \textbf{61.97} & - \\
\midrule
\multirow{5}*{Out-domain Test}
& Rennes & 65.11 & \textbf{73.13} & 60.19 & 47.40 & 57.71 & 54.26 \\
& Bordeaux & 50.90 & 71.87 & 56.96 & 34.70 & 50.13 & 46.71\\
& Lyon & 65.79 & 72.50 & \textbf{69.01} & 52.44 & 57.29 & 60.13 \\
& Ljubljana & \textbf{68.37} & 70.25 & 66.73 & \textbf{59.85} & 64.17 & 66.30 \\
& Best & 61.19 & 71.35 & 64.17 & 48.58 & \textbf{64.43} & 58.34 \\
\bottomrule
    \end{tabular}
    \end{small}
\caption{Cross-performance results using nDSC $(\uparrow)$ (\%) for selected splits. Ensembles of 5 models are always used. Threshold is searched in increments of 0.01. The following threshold were used respectively for the in-domain dev threshold tuning: [0.8, 0.1, 0.47, 0.66, 0.50]. Here, R-AUC is calculated over all voxels in each image.}
\label{tab:results_individualLocations}
\end{table}

We performed N-fold cross-validation in Table \ref{tab:nfold} to determine which location should be considered as the shifted set as training on single locations may lead to unreliable conclusions due to the small size of the training sets. Train 5 systems (each one an ensemble) using all training data apart from one site at a time. Hyperparameters are tuned using all the dev sets apart from the site excluded. We evaluate this system on all the data (train + dev + test) from the excluded site (out) and in-domain test sets too. From the results, Ljubjana is an appropriate choice for the shifted development set as it faces the greatest degradation compared to in-domain performance.

\definecolor{LightCyan}{rgb}{0.88,1,1}

\begin{table}[ht]
\centering
\begin{small}
    \begin{tabular}{p{1.5cm}l|lll|lll}
    \toprule
    & & \multicolumn{3}{c|}{nDSC (\%) $\left(\uparrow\right)$} & \multicolumn{3}{c}{R-AUC (\%) $\left(\downarrow\right)$} \\
Excluded \newline Location & Model & In & Out & Lausanne & In & Out & Lausanne \\
\midrule
\multirow{2}*{Rennes}
& Single & $66.43_{\pm 0.50}$ & $70.86_{\pm 0.42}$ & $64.50_{\pm 0.83}$& $3.10_{\pm 0.21}$ & $2.81_{\pm 0.55}$ & $6.54_{\pm 0.96}$\\
& Ensemble & 68.01 & 72.48 & 66.46 & 2.02 & 1.69 & 4.14 \\
\midrule
\multirow{2}*{Bordeaux} 
& Single & $65.66_{\pm 0.74}$ & $72.14_{\pm 1.10}$ & $63.21_{\pm 1.26}$ & $3.09_{\pm 0.19}$ & $2.48_{\pm 0.36}$ & $6.61_{\pm 0.58}$\\
& Ensemble & 66.33 & 72.73 & 63.25 & 1.87 & 1.33 & 4.05\\
\midrule
\multirow{2}*{Lyon}  
& Single & $63.51_{\pm 0.18}$ & $69.27_{\pm 0.69}$ & $61.85_{\pm 1.69}$ & $3.68_{\pm 0.76}$ & $2.33_{\pm 0.62}$ & $6.69_{\pm 1.54}$\\
& Ensemble & 65.21 & 70.69 & 64.46 & 2.54 & 1.81 & 4.70 \\
\midrule
\rowcolor{LightCyan}
& Single & $67.59_{\pm 0.63}$ & $49.33_{\pm 1.52}$ & $55.70_{\pm 1.04}$& $2.77_{\pm 0.98}$ & $7.84_{\pm 2.21}$ & $9.87_{\pm 1.40}$\\
\rowcolor{LightCyan}
\multirow{-2}*{Ljubljana} & Ensemble & 68.89 & 50.85 & 57.53 & 1.76 & 4.66 & 7.40 \\

\midrule
\multirow{2}*{Best}
& Single & $65.87_{\pm 1.62}$ & $57.37_{\pm 0.79}$ & $61.78_{\pm 2.21}$ & $2.65_{\pm 0.48}$ & $3.05_{\pm 0.69}$ & $5.70_{\pm 1.22}$\\
& Ensemble & 66.68 & 58.38 & 61.93 & 1.54 & 1.69 & 3.15\\
   \bottomrule
    \end{tabular}
    \end{small}
\caption{N-fold cross-validation with nDSC $(\uparrow)$ (\%) as the performance metric. The threshold is selected based on the (in-domain) development set. The following thresholds are used: [0.25, 0.55, 0.25, 0.35, 0.55]. Entropy is used as the uncertainty measure for single models and reverse mutual information for ensembles.}
\label{tab:nfold}
\end{table}

\begin{figure}[h!]
    \centering
    \includegraphics[width=0.7\textwidth]{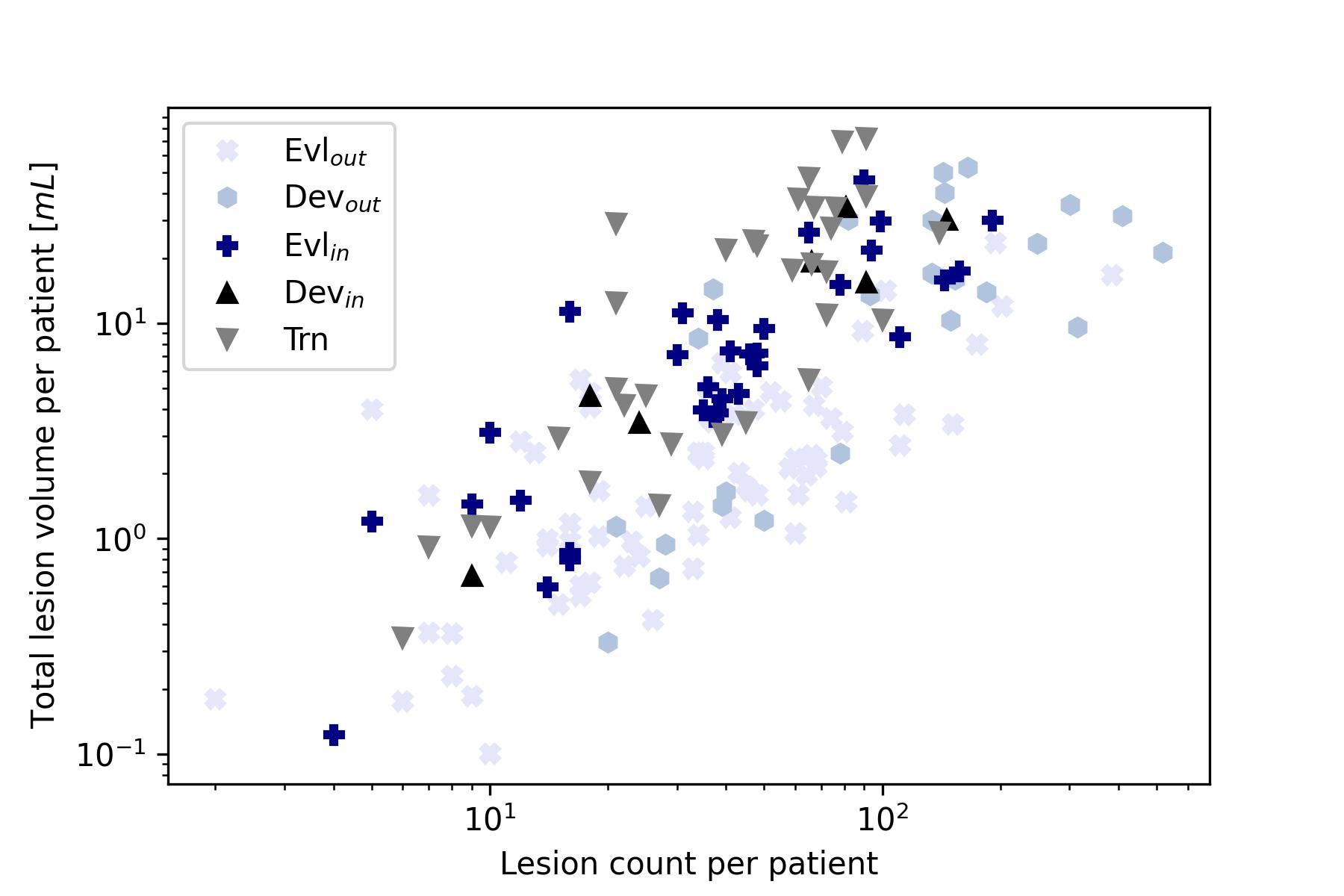}
    \caption{Log–log plot of white matter lesions characteristics in terms of per patient total lesion volume (TLV) and number of lesions for different datasets.}
    \label{fig:datadistribution}
\end{figure}

\paragraph{Data distributions} Here, a more detailed characterisation of the datasets (Trn, Dev$_{\text{in}}$, Evl$_{\text{in}}$, Dev$_{\text{out}}$ and Evl$_{\text{out}}$) described in Section \ref{sec:wml} is given. Distributions of total lesion volumes and number of lesions across patients are shown in Figure \ref{fig:datadistribution}. General characteristics of the datasets are given in Table \ref{tab:datadistribution}. It can be seen that the difference in datasets comes not only from the location of the medical center or the scanner type, but also from the the sizes of lesions. Out-of-domain datasets have more subjects with smaller lesions. Per patient lesion counts, however, do not vary significantly across the datasets.
Additionally, it was mentioned in the main paper that the component datasets of our benchmark are based on ISBI\cite{CARASS201777, carass2017}, MSSEG-1 \cite{commowick2018} and PubMRI \cite{Lesjak2017ANP}. Table \ref{tab:sourcedatameta} offers additional meta-information on these source datasets with regard to age and gender ratio of the patient scans from each of these datasets.

\begin{table}[h!]
    \centering
    {\small
    \begin{tabular}{p{3cm}|*{5}{|c}} 
        \toprule
         Parameters & Trn & Dev$_{\text{in}}$ & Evl$_{\text{in}}$ & Dev$_{\text{out}}$ & Evl$_{\text{out}}$ \\ 
        \midrule
        Total lesion count & 1628 & 435 & 1738 & 3544 & 3826 \\
        \midrule
        MS stages & \multicolumn{3}{c|}{RR, PP, SP$^*$} & RR, SP, PR, CIS & RR \\
        \midrule
        Average across scans \newline TLV, $mL$ & $18.58_{\pm 18.75}$ & $15.49_{\pm 12.42}$ & $10.03_{\pm 10.28}$ & $17.10_{\pm 15.56}$ & $3.34_{\pm 4.13}$ \\
        \bottomrule
    \end{tabular}
    }
    \caption{Additional characteristics of the datasets, such as total amount of lesions in a dataset, MS stages and average across scans total lesion volume (TLV) in milliliters. MS stages abbreviations: RR - relapsing remitting, PP - primary progressive, SP - secondary progressive, CIS - clinically isolated syndrome. \\ {\small$^*$Information about MS stages in MSSEG-1 was not found.}}
    \label{tab:datadistribution}
\end{table}

\begin{table}[h!]
    \centering
    {\small
    \begin{tabular}{l|ccc} 
        \toprule
& ISBI & MSSEG-1 & PubMRI \\
\midrule
Age (years) & $40.4_{\pm 9.3}$ & $45.3_{\pm 10.3}$ & 39 (median) \\
Gender ratio (M:F) & 0.21 & 0.40 & 0.23 \\
Inter-rater agreement (DSC) & 0.63 & 0.71 & 0.78 \\
        \bottomrule
    \end{tabular}
    }
    \caption{Age and gender meta-information for source datasets. Additionally, inter-rater agreement is reported as DSC.}
    \label{tab:sourcedatameta}
\end{table}

\paragraph{Format} The data will be shared as a series of compressed .nii files, all the data within will be pre-processed, interpolated to the 1mm iso-voxel space and skull-stripped for additional anonymisation. We will share both the T1 weighted and FLAIR modalities.

\subsection{Performance metrics}

We now detail performance metrics used to assess lesion segmentation models.

\subsubsection{Normalized Dice Similarity Coefficient (nDSC)} \label{app:adaptedDSC}

Typically, the Dice Similarity Coefficient (DSC) is used as the performance metric between the ground-truth $Y$ and its corresponding prediction $\hat{Y}$:
$$\text{DSC} = \frac{2|Y\cap \hat{Y}|}{|Y| + |\hat{Y}|} = \frac{2TP}{FP+2TP+FN} = 2\frac{\texttt{ precision } * \texttt{ recall}}{\texttt{precision } + \texttt{ recall}}$$
The reported score is usually the DSC averaged across all patient scans.
However, DSC is biased to yield greater values for patients that have a greater lesion load i.e. a greater probability of the event occurring, where the event here is described as identifying a voxel as a lesion. To de-correlated DSC with lesion-load and obtain an unbiased metric of permormance, we consider a normalised DSC (nDSC). The following steps explain and justify how and why we calculate the proposed nDSC:
\begin{enumerate}
    \item The probability of a successful event (identifying a lesion) influences the DSC score as the precision at 100\% recall varies across the patients (the precision at 100\% recall is simply the percentage of lesion voxels for the patient - i.e. the lesion load).
    \item The DSC score is calculated as a geometric ratio of the precision, $\text{Pr}_{\tau}$, and recall, $\text{Re}_{\tau}$ values at a selected threshold, $\tau$ (ML models typically have a probabilistic prediction for each voxel which must be compared against a threshold to classify as either a positive class or a negative class).
    \item Here, the recall is held fixed and the precision for each patient is adjusted ($\text{Pr}_{\tau} \rightarrow \overline{\text{Pr}}_{\tau} $) by a different amount such that the cross-patient performance can be fairly evaluated.
    \item The new value of the precision is determined by the scaling applied to the FP (false positives) which is scaled by a factor, $k_{p}$ that is different for each patient, $p$.
    \item $k_p$ for each patient is determined by using the 100\% recall rate point as this point is not influenced by model performance. 
    \item Hence, $k_p$ for patient $p$ is the factor the FP at 100\% recall must be scaled by in order to ensure the precision achieved is a chosen reference value, $r$.
    Derivation of deducing $k_p$ is given. The subscript 100\% denotes operating at 100\% recall.
    $$ \text{Pr}_{100\%} = \frac{\text{TP}_{100\%}}{\text{TP}_{100\%} + \text{FP}_{100\%}},\quad r = \overline{\text{Pr}}_{100\%} = \frac{\text{TP}_{100\%}}{\text{TP}_{100\%} + k_p\text{FP}_{100\%}}, \quad  k_p = \frac{(1-r)\text{TP}_{100\%}}{r\text{FP}_{100\%}}$$
    % $$r = \overline{\text{Pr}}_{100\%} = \frac{\text{TP}_{100\%}}{\text{TP}_{100\%} + k_p\text{FP}_{100\%}} $$
    % $$k_p = \frac{(1-r)\text{TP}_{100\%}}{r\text{FP}_{100\%}}$$
    \item Here, $r$ is selected as 0.1\% because this is approximately the average precision across the patients at 100\% recall (i.e. the average fraction of lesion voxels).
    \item The recall is not influenced by scaling the FP by $k_p$.
    \item The precision is directly affected as the new precision at our selected operating point (threshold to form the segmentation mask), $\tau^*$, is given by:
    $$\overline{\text{Pr}}_{\tau^*} = \frac{\text{TP}_{\tau^*}}{\text{TP}_{\tau^*} + k_p\text{FP}_{\tau^*}}$$
    Recall, $k_p$ is given in step 6.
    \item Thus, nDSC is calculated as the geometric mean of $\overline{\text{Pr}}_{\tau^*}$ and $\text{Re}_{\tau^*}$ for each patient.
\end{enumerate}
The averaged nDSC is used as the predictive performance metric.

\begin{figure}[htbp!]
     \centering
     \subfigure[DSC]{\includegraphics[width=0.49\textwidth]{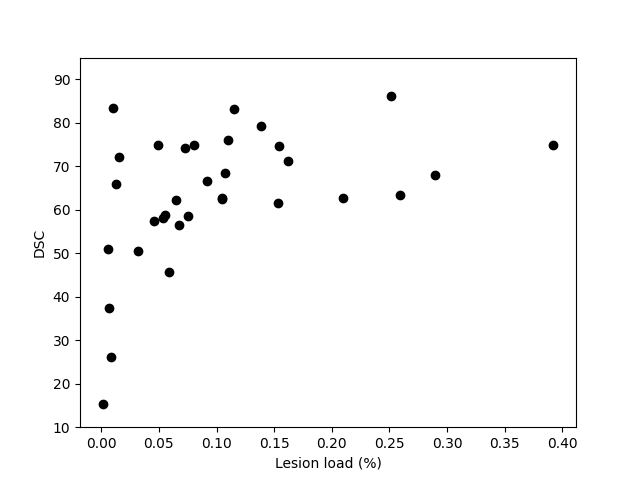}}
     \subfigure[nDSC]{ \includegraphics[width=0.49\textwidth]{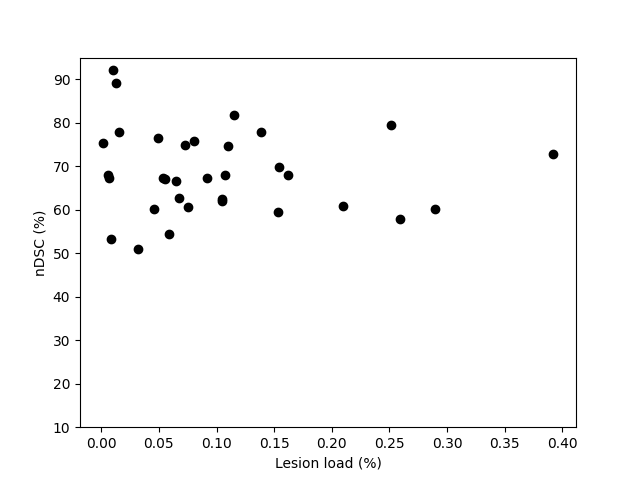}}
     \caption{Empirical relationship of each metric with lesion load on Evl$_{\text{in}}$ using UNET ensemble.}\label{fig:ndsc_correlations}
\end{figure}

We empirically demonstrate that the nDSC metric is less dependent on the lesion load compared to DSC via Figure \ref{fig:ndsc_correlations} and table\ref{tab:transition}. Recall, lesion load is defined as the fraction of voxels that are lesion voxels for a given subject. Figure \ref{fig:ndsc_correlations} plots the performance in terms of both DSC and nDSC against the lesion load for each subject for Evl$_{\text{in}}$. It is clear that DSC is dependent on the lesion load while nDSC decorrelates this relationship by flat line average. Table \ref{tab:transition} presents the transition table between DSC and nDSC as well as providing the Spearman's rank correlation coefficients between either DSC or nDSC with the lesion load. Notably, the nDSC metric is less correlated with the lesion load than DSC for each of the splits.
\begin{table}[htbp!]
\centering
    \begin{tabular}{l|cc|cc}
    \toprule
    \multirow{2}*{Split} & \multicolumn{2}{c|}{Performance} & \multicolumn{2}{c}{Correlation}  \\
     & DSC & nDSC & DSC & nDSC \\
    \midrule
dev-in & 71.71 & 68.54 & 0.63 & -0.09 \\
dev-out & 49.85 & 49.33 & 0.57 & 0.46 \\
eval-in & 63.16 & 67.59 & 0.44 & -0.10 \\
eval-out & 48.48 & 55.79 & 0.40 & 0.18 \\
   \bottomrule
    \end{tabular}
\caption{Performance and Pearson's rank correlation coefficients between metric and the lesion load for the canonical white matter lesion segmentation splits using the baseline UNET model ensemble.}
    \label{tab:transition}
\end{table}

\subsubsection{Lesion-scale F1 score} \label{appendix:lesionscalef1}

For MS lesion segmentation task it is important to assess not only the overall voxel-level segmentation quality, but also the lesion detection quality. Therefore, in addition to the nDSC we calculate the lesion-scale F1 score.

A general formula for computation of the F1-score:

\begin{equation}
    F_1 = \frac{TP}{TP + 0.5(FP + FN)}
\end{equation}

can be adapted for the assessment of lesions detection quality given a proper definition of true positive, false positive and false negative lesions. 

We use the intersection over union (IoU) between lesions on a ground truth map and connected components on a corresponding prediction map to derive these definitions. In particular, the following condition were used:
\begin{itemize}
    \item[TP:] If the maximum IoU between a connected component on the prediction map with lesions on the ground truth is greater than $0.5$.
    \item[FP:] If the maximum IoU between a connected component on the prediction map with lesions on the ground truth is less than $0.5$.
    \item[FN:] If the maximum IoU between a lesion on the ground truth map with connected components on the prediction map is less than $0.5$.
\end{itemize}

\subsection{Additional Results}

For completeness, Monte Carlo dropout~\cite{Gal2016Dropout} based ensembles are considered here too using the UNET architecture. The baseline single models considered here have no dropout (as this gives best performance on Dev-in) and the deep ensembles are built using these single models. The deep ensemble is formed by averaging the output probabilities from 5 distinct single models. A separate set of 5 models are trained with 50\% dropout in each model in order to be able to perform Monte Carlo Dropout (MCDP) as an additional comparison. The single models here have dropout usually turned off at inference time. For MCDP, a single model is taken and dropout is turned on at inference time with an ensemble formed from 5 separate runs of the model (as the dropout introduces stochasticity). The process is repeated for each of the single models with dropout to get averaged results. 
As each single model yields a per-voxel probabilistic prediction, ensemble-based uncertainty measures\cite{malinin-thesis,malinin2021structured} are available for uncertainty quantification. Our ensembled models (Deep Ensemble and MCDP) use reverse mutual information \cite{malinin2021structured} as the choice of uncertainty measure. Single models use the entropy of the discrete binary probability distribution at each voxel to capture the uncertainties. All results reported for single models are the mean of the individual model performances with one standard deviation indicated.

Tables \ref{tab:app_performance_nDSC} and \ref{tab:app_performance_F1} present the performance ability of various baseline models. Table \ref{tab:app_performance_nDSC} focuses on the ability of the models to identify the exact delineations of lesions through nDSC (voxel-scale) while Table \ref{tab:app_performance_F1} compares the lesion detection ability of the models with F1 (lesion-scale). Comparing the in-domain performance against the out-of-domain performance, it is clear that the shift in the location naturally leads to severe degradation in performance at both the voxel-scale and the lesion-scale with drops exceeding 10\% nDSC and F1. Comparing the deep ensembles against the single models, it is clear that ensembling such models boosts performance by about 1\% nDSC and 1\% F1 for each of the test sets. In particular, the transformer based architecture, UNETR, is able to outperform the fully convolutional architecture, UNET, for both the single and ensembled performance in terms of delineation and lesion detection of about 2\% nDSC and 5\% F1 respectively across the various splits. Introducing dropout in the models at training time costs the single model in performance at both voxel and lesion-scales with greater degradation observed in the in-domain splits. Consequently, the detrimental effect of dropout at training time seriously harms the performance of the MCDP systems that keep the dropout on at training time.
\begin{table}[ht]
% \fontsize{8}{9}\selectfont
\centering
\begin{small}
    \begin{tabular}{lll|llll}
    \toprule
\multirow{2}{*}{Arch} & \multirow{2}{*}{DP} & \multirow{2}{*}{Model} & \multicolumn{4}{c}{nDSC (\%) $\left(\uparrow\right)$}  \\
& & & Dev$_{\text{in}}$  & Dev$_{\text{out}}$ & Evl$_{\text{in}}$ & Evl$_{\text{out}}$  \\
\midrule
\multirow{4}{*}{UNET} & \multirow{2}{*}{0.0} & Single & $68.54_{\pm 0.68}$  & $49.33_{\pm 1.52}$ & $67.59_{\pm 0.63}$ & $55.79_{\pm 1.04}$  \\
& & Deep Ensemble & $69.70$  & $50.85$ & $68.89$ & $57.53$  \\
\cmidrule{2-7}
& \multirow{2}{*}{0.5} & Single &  $59.73_{\pm 1.17}$  & $48.35_{\pm 1.73}$ & $63.93_{\pm 0.45}$ & $54.43_{\pm 1.41}$  \\
& & MCDP & $60.65_{\pm 0.91}$  &  $44.70_{\pm 1.35}$ & $61.78_{\pm 0.90}$ & $50.06_{\pm 1.67}$ \\
\midrule
\multirow{2}{*}{UNETR} & \multirow{2}{*}{0.0} & Single  & $71.21_{\pm 0.96}$ & $51.60_{\pm 1.66}$ & $69.27_{\pm 0.94}$ & $56.76_{\pm 2.63}$ \\
& & Deep Ensemble & $72.51$ &  $53.46$ & $71.41$ & $59.49$ \\
  \bottomrule
    \end{tabular}
    \end{small}
\caption{Lesion segmentation: Performance at voxel-level with nDSC with 1 standard deviation quoted for single results.}
\label{tab:app_performance_nDSC}
\end{table}

% \begin{table}[ht]
% % \fontsize{8}{9}\selectfont
% \centering
% \begin{small}
%     \begin{tabular}{lll|llll}
%     \toprule
% \multirow{2}{*}{Arch} & \multirow{2}{*}{DP} & \multirow{2}{*}{Model} & \multicolumn{4}{c}{nDSC (\%) $\left(\uparrow\right)$}  \\
% & & & Dev$_{\text{in}}$  & Dev$_{\text{out}}$ & Evl$_{\text{in}}$ & Evl$_{\text{out}}$  \\
% \midrule
% \multirow{4}{*}{UNET} & \multirow{2}{*}{0.0} & Single & $67.47_{\pm 0.54}$  & $47.21_{\pm 1.68}$ & $67.27_{\pm 0.48}$ & $54.96_{\pm 1.14}$  \\
% & & Deep Ensemble & $67.97$  & $48.81$ & $68.58$ & $56.60$  \\
% \cmidrule{2-7}
% & \multirow{2}{*}{0.5} & Single &  $59.73_{\pm 1.17}$  & $48.35_{\pm 1.73}$ & $63.93_{\pm 0.45}$ & $54.43_{\pm 1.41}$  \\
% & & MCDP & $61.12_{\pm 0.91}$  &  $46.37_{\pm 1.35}$ & $61.90_{\pm 0.90}$ & $51.76_{\pm 1.67}$ \\
% \midrule
% \multirow{2}{*}{UNETR} & \multirow{2}{*}{0.0} & Single  & $71.85_{\pm 0.60}$ & $51.95_{\pm 0.78}$ & $69.71_{\pm 0.63}$ & $58.62_{\pm 1.19}$ \\
% & & Deep Ensemble & $72.87$ &  $53.78$ & $71.32$ & $60.41$ \\
%   \bottomrule
%     \end{tabular}
%     \end{small}
% \caption{Lesion segmentation: Performance at voxel-level with nDSC with 1 standard deviation quoted for single results.}
% \label{tab:app_performance_nDSC}
% \end{table}

\begin{table}[ht]
% \fontsize{8}{9}\selectfont
\centering
\begin{small}
    \begin{tabular}{lll|llll}
    \toprule
\multirow{2}{*}{Arch} & \multirow{2}{*}{DP} & \multirow{2}{*}{Model} &   \multicolumn{4}{c}{F1 (\%) $\left(\uparrow\right)$}   \\
& & & Dev$_{\text{in}}$  & Dev$_{\text{out}}$ & Evl$_{\text{in}}$ & Evl$_{\text{out}}$  \\
\midrule
\multirow{4}{*}{UNET} & \multirow{2}{*}{0.0} & Single &  $25.02_{\pm 2.51}$  & $8.17_{\pm 0.73}$ & $25.46_{\pm 1.51}$ & $14.79_{\pm 0.71}$  \\
& & Deep Ensemble &  $28.07$  & $9.04$ & $27.74$ & $16.74$ \\
\cmidrule{2-7}
& \multirow{2}{*}{0.5} & Single &   $14.42_{\pm 0.43}$  & $6.75_{\pm 0.70}$ & $18.66_{\pm 0.51}$ & $11.85_{\pm 0.47}$ \\
& & MCDP & $12.61_{\pm 0.89}$  & $4.59_{\pm 0.78}$ & $17.31_{\pm 0.95}$ & $10.70_{\pm 0.58}$ \\
\midrule
\multirow{2}{*}{UNETR} & \multirow{2}{*}{0.0} & Single  & $33.60_{\pm 1.36}$ & $15.03_{\pm 1.16}$ & $33.85_{\pm 0.43}$ & $17.19_{\pm 1.22}$ \\
& & Deep Ensemble & $35.22$ & $15.80$ & $35.61$ & $18.90$ \\
  \bottomrule
    \end{tabular}
    \end{small}
\caption{Lesion segmentation: Performance at lesion-level with F1 with 1 standard deviation quoted for single results.}
\label{tab:app_performance_F1}
\end{table}

% \begin{table}[ht]
% % \fontsize{8}{9}\selectfont
% \centering
% \begin{small}
%     \begin{tabular}{lll|llll}
%     \toprule
% \multirow{2}{*}{Arch} & \multirow{2}{*}{DP} & \multirow{2}{*}{Model} &   \multicolumn{4}{c}{F1 (\%) $\left(\uparrow\right)$}   \\
% & & & Dev$_{\text{in}}$  & Dev$_{\text{out}}$ & Evl$_{\text{in}}$ & Evl$_{\text{out}}$  \\
% \midrule
% \multirow{4}{*}{UNET} & \multirow{2}{*}{0.0} & Single &  $24.02_{\pm 0.66}$  & $7.75_{\pm 0.66}$ & $25.97_{\pm 0.52}$ & $14.05_{\pm 1.55}$  \\
% & & Deep Ensemble &  $25.56$  & $8.07$ & $26.78$ & $15.97$ \\
% \cmidrule{2-7}
% & \multirow{2}{*}{0.5} & Single &   $14.42_{\pm 0.43}$  & $6.75_{\pm 0.70}$ & $18.66_{\pm 0.51}$ & $11.85_{\pm 0.47}$ \\
% & & MCDP & $16.08_{\pm 0.89}$  & $5.32_{\pm 0.78}$ & $16.85_{\pm 0.95}$ & $11.17_{\pm 0.58}$ \\
% \midrule
% \multirow{2}{*}{UNETR} & \multirow{2}{*}{0.0} & Single  & $33.23_{\pm 1.26}$ & $15.11_{\pm 1.49}$ & $34.13_{\pm 0.31}$ & $17.63_{\pm 1.73}$ \\
% & & Deep Ensemble & $33.83$ & $16.02$ & $35.43$ & $18.93$ \\
%   \bottomrule
%     \end{tabular}
%     \end{small}
% \caption{Lesion segmentation: Performance at lesion-level with F1 with 1 standard deviation quoted for single results.}
% \label{tab:app_performance_F1}
% \end{table}

\subsection{Uncertainty estimation}\label{apn:med-uncertainty}

\begin{figure}
    \centering
    \includegraphics[width=0.7\textwidth]{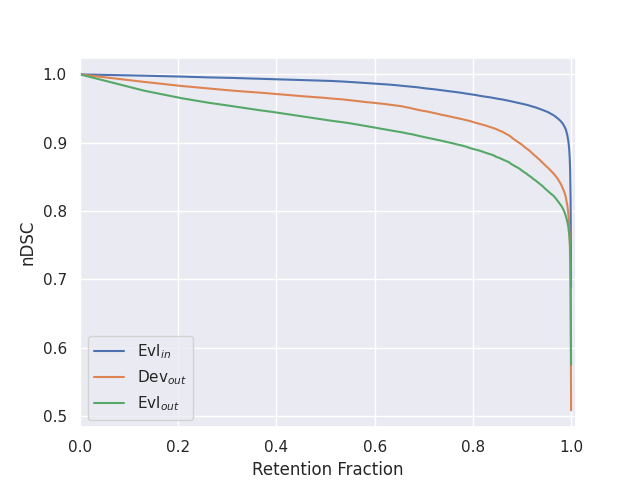}
    \caption{nDSC retention curves using the ensembled UNET on various canonincal splits.}
    \label{fig:ndsc_retention}
\end{figure}

Table \ref{tab:app_unc} explores the joint robustness and uncertainty quantification performance using the R-AUC metric. Here, the deep ensemble of the UNETR outperforms all other systems, achieving R-AUC scores as low as 0.63 on Evl$_{\text{in}}$ and 2.88 on Evl$_{\text{out}}$. It is interesting to note that despite performing worse at voxel-scale identification of lesions, the MCDP system does better than its equivalent single system when jointly assessing uncertainty and robustness. Therefore, it is clear that the quality of the uncertainty measures in the ensembled-based models (including both the deep ensemble and MCDP) allows the development of richer uncertainty quantification measures compared to single models. Figure \ref{fig:ndsc_retention} presents the corresponding retention curves (averaged across all the patients with one example model chosen for the single systems) using the deep ensembled UNET on the Evl$_{\text{in}}$, Dev$_{\text{out}}$ and Evl$_{\text{out}}$ splits. All systems substantially outperform a randomized ordering as a large volume of the input brain image is non white-matter tissue, for which the system is correctly certain that there are no white matter lesion voxels present in those regions. Particularly, the retention curve for the Evl$_{\text{in}}$ appears to be very close to ideal which demonstrates the high quality of its voxel-scale uncertainties at identifying regions where the model is not confident in its prediction.
\begin{table}[ht]
\fontsize{8}{9}\selectfont
\centering
\begin{small}
    \begin{tabular}{lll|llll}
    \toprule
\multirow{2}{*}{Arch} & \multirow{2}{*}{DP} & \multirow{2}{*}{Model} & \multicolumn{4}{c}{R-AUC (\%) $\left(\downarrow\right)$} \\
& & & Dev$_{\text{in}}$  & Dev$_{\text{out}}$ & Evl$_{\text{in}}$ & Evl$_{\text{out}}$ \\
\midrule
\multirow{4}{*}{UNET} & \multirow{2}{*}{0.0} & Single & $2.51_{\pm 0.59}$  & $7.84_{\pm 2.21}$ & $2.77_{\pm 0.98}$ & $9.87_{\pm 1.40}$ \\
& & Deep Ensemble & $1.17$  & $4.66$ & $1.76$ & $7.40$  \\
\cmidrule{2-7}
& \multirow{2}{*}{0.5} & Single & $2.62_{\pm 0.56}$  & $8.76_{\pm 1.08}$ & $2.66_{\pm 0.56}$ & $9.71_{\pm 1.53}$\\
& & MCDP & $1.92_{\pm 0.26}$  & $6.77_{\pm 0.79}$ & $2.52_{\pm 0.41}$ & $7.89_{\pm 1.04}$\\
\midrule
\multirow{2}{*}{UNETR} & \multirow{2}{*}{0.0} & Single & $1.89_{\pm 0.84}$ & $6.17_{\pm 1.99}$  & $1.95_{\pm 0.70}$& $6.47_{\pm 2.08}$ \\
& & Deep Ensemble & $0.34$ & $1.52$ & $0.63$ & $2.88$ \\
  \bottomrule
    \end{tabular}
    \end{small}
\caption{Lesion segmentation: Joint robustness and uncertainty assessment (using reverse mutual information for ensembled models and entropy for single models) at voxel-level with R-AUC. 1 standard deviation is quoted for single results.}
\label{tab:app_unc}
\end{table}

% \begin{table}[ht]
% \fontsize{8}{9}\selectfont
% \centering
% \begin{small}
%     \begin{tabular}{lll|llll}
%     \toprule
% \multirow{2}{*}{Arch} & \multirow{2}{*}{DP} & \multirow{2}{*}{Model} & \multicolumn{4}{c}{R-AUC (\%) $\left(\downarrow\right)$} \\
% & & & Dev$_{\text{in}}$  & Dev$_{\text{out}}$ & Evl$_{\text{in}}$ & Evl$_{\text{out}}$ \\
% \midrule
% \multirow{4}{*}{UNET} & \multirow{2}{*}{0.0} & Single & $2.11_{\pm 0.18}$  & $3.53_{\pm 0.80}$ & $1.68_{\pm 0.08}$ & $3.75_{\pm 0.22}$ \\
% & & Deep Ensemble & $1.93$  & $2.91$ & $1.48$ & $3.43$  \\
% \cmidrule{2-7}
% & \multirow{2}{*}{0.5} & Single & $2.56_{\pm 0.56}$  & $3.13_{\pm 1.08}$ & $2.20_{\pm 0.56}$ & $3.73_{\pm 1.53}$\\
% & & MCDP & $2.21_{\pm 0.26}$  & $3.70_{\pm 0.79}$ & $1.79_{\pm 0.41}$ & $3.77_{\pm 1.04}$\\
% \midrule
% \multirow{2}{*}{UNETR} & \multirow{2}{*}{0.0} & Single & $2.42_{\pm 0.38}$ & $5.22_{\pm 0.48}$  & $2.95_{\pm 0.25}$& $5.43_{\pm 3.27}$ \\
% & & Deep Ensemble & $1.95$ & $4.60$ & $2.40$ & $5.21$ \\
%   \bottomrule
%     \end{tabular}
%     \end{small}
% \caption{Lesion segmentation: Joint robustness and uncertainty assessment (using reverse mutual information for ensembled models and entropy for single models) at voxel-level with R-AUC. 1 standard deviation is quoted for single results.}
% \label{tab:app_unc}
% \end{table}
% \begin{figure}
%     \centering
%     \includegraphics[width=0.7\textwidth]{figures/unc_ret_dsc_norm.png}
%     \caption{nDSC retention curves using the ensembled UNET on various canonincal splits.}
%     \label{fig:ndsc_retention}
% \end{figure}

\begin{figure}[htp]
     \centering
    \includegraphics[width=0.9\textwidth]{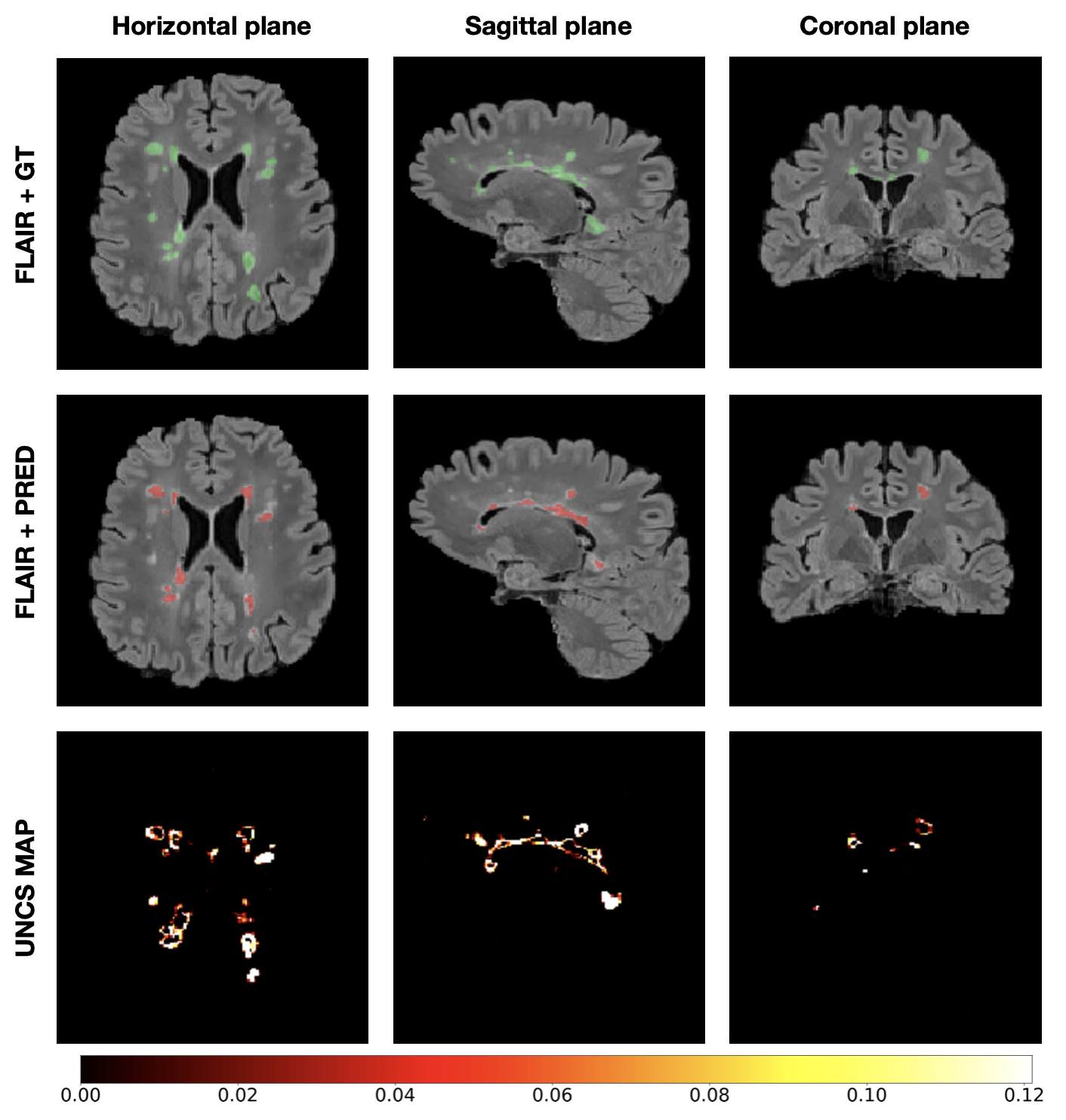}
    \caption{Examples on one subject of a FLAIR image with a ground truth (\textbf{FLAIR + GT}) and predicted (\textbf{FLAIR + PRED}) WML maps overlays and an uncertainty map (\textbf{UNCS MAP}). For each of the 3D maps single horizontal, sagittal and coronal slices are displayed. Predictions were obtained using an ensemble of 5 UNET models. Uncertainty map was computed as reversed mutual information from the probabilistic voxel-wise predictions of models in ensemble. Color bar corresponds to the uncertainty map, where outlying values above 0.121 are displayed in white. All images were displayed using ITK-SNAP software \cite{py06nimg}.}
    \label{apn:uncsmap}
\end{figure}

Figure \ref{apn:uncsmap} gives an idea about the spatial distribution of uncertainty. In particular, it can be seen that higher uncertainty regions are located around predicted lesions, therefore should be related to the quality of delineation. False negative lesions, however, can also have higher uncertainties in comparison to the background.
% \begin{figure}[htp]
% \begin{subfigure}
%      \centering
%     \includegraphics[width=0.3\textwidth]{paper-old/figures/flair_gt.png}
%     %\caption{Ground truth.}
% \end{subfigure}%
% \hfill
% \begin{subfigure}
%      \centering
%     \includegraphics[width=0.3\textwidth]{paper-old/figures/flair_pred.png}
%     %\caption{Prediction.}
% \end{subfigure}%
% \hfill
% \begin{subfigure}
%      \centering
%     \includegraphics[width=0.3\textwidth]{paper-old/figures/uncs_map.png}
%     %\caption{Uncertainty map.}
% \end{subfigure}
%   \caption{Example on one subject of a FLAIR image with ground truth and prediction map overlays and uncertainty map (from left to right). Predictions and uncertainty maps were obtained using a deep ensemble of 5 UNET models. Uncertainty maps were computed using voxel-wise probabilistic predictions with reverse mutual information.}
%     \label{apn:uncsmap}
% \end{figure}

\newpage

\section{Ship Power Consumption}\label{apn:ship}

\subsection{Dataset Description}\label{apn:ship-data}

\paragraph{Collection Process} Data is collected by DeepSea from a real vessel as it experiences various weather, loading and operational conditions. The sampling frequency is 1 min, and achieved by either interfacing directly with the vessel’s sensors or through already existing onboard signal aggregation systems like the Electronic Chart Display and Information System (ECDIS) or the vessel’s Alarm Monitoring System (AMS). The interfacing in either case is performed by dedicated IoT edge devices like Deepsea’s Neuro\footnote{https://www.deepsea.ai/the-neuro/} that gather all relevant data and transmit them via satellite link to Deepseas’ databases.

\paragraph{Preprocessing, Cleaning and Labeling} The available features are recorded by on-board sensors and the global positioning system (GPS) is being used to complement the acquired data with weather data from a global weather provider. The data is preprocessed to remove extreme outliers and stationary states, for example when a vessel is at port, by applying feature filters. Furthermore, we create a second dataset, the synthetic dataset, by combining the real samples with synthetic power labels generated by our synthetic model (detailed below).

\paragraph{Partitioning into train, development, and evaluation sets} We create a canonical partitioning of power estimation dataset so that it contains both in-domain and shifted components. In order to define the distributional shifts, the data split along two dimensions: time and true wind speed, as shown in Figure \ref{fig:ds_partitioning}, using the wind speed intervals from Table \ref{table:ds_wind_intervals}.

The time dimension is intended to capture the non-stationary effects of fouling (no cleaning events occur during the time period under study), whereas the wind speed dimension is intended to capture weather effects (by acting as a proxy since wind is correlated with wind-waves) and to better expose the model's performance in bad or uncertain weather. Partitioning the datasets in more dimensions would have added complexity without adding any practical benefits because the most important uncertainty factors (weather and fouling) are already represented. 

Given these shifts, three main subsets are created:
\begin{itemize}
    \item \textbf{Train set}: It covers the time range of 39.4 months starting after a dry docking cleaning event and includes data with true wind speed up to 19 kn.
    \item \textbf{Development set}: It consists of an in-domain partition dev\_in and an out-of-domain partition dev\_out, with equal representatives. Dev\_in is sampled from the same partitions as the train set while dev\_out includes more recent records (time period of 6.6 months) that correspond to wind speeds in the range [19, 26) kn.
    \item \textbf{Evaluation set}: Evaluation set, like development set, have an in-domain eval\_in and an out-of-domain split eval\_out with equal populations.Eval\_in is sampled from the same subsets as the train set. Eval\_out is the most shifted partition from the in-domain distribution, containing the most recent records spanning an 18 months period and the most severe wind conditions seen in the whole dataset, corresponding to wind speeds ranging between [19, 40] kn.
\end{itemize}

The number of records of the proposed partitions (rows) along with the respective populations in each 2D segmentation (columns with prefix group) of the synthetic and real datasets are reported in Tables \ref{table:ds_no_records_synthetic} and \ref{table:ds_no_records_real} respectively. 

\begin{table}[htp]
\centering
    \begin{small}
        \begin{tabular}{ccc}
        \toprule
        Wind interval & Range (kn) & Range in Beaufort \\
        \midrule
        1  & [0, 9) & Up to ~3 \\
        2 & [9, 14) & 3-4 \\
        3 & [14, 19) & 4-5 \\
        4 & $\geq19$ & $\geq5$ \\
        \bottomrule
        \end{tabular}
    \end{small}
\caption{Wind intervals considered for data partitioning. Beaufort ranges  are defined approximately.}
\label{table:ds_wind_intervals}
\end{table}

\begin{table}[htp]
\centering
    \begin{small}
        \begin{tabular}{l|cccccc}
        \toprule
        Data & pct (\%) & total & \textcolor{magenta}{Group 1} & \textcolor{cyan}{Group 2} &  \textcolor{yellow}{Group 3} &  \textcolor{red}{Group 4} \\
        \midrule
        train & 80.3 & 523190 & 231626 & 118698 & 172866 & 0 \\
        dev\_in & - & 18108 & 8017 & 4108 & 5983 & 0 \\
        dev\_out & - & 18108 & 0 & 0 & 0 & 18108 \\
        dev & 5.6 & 36216 & 8017 & 4108 & 5983 & 18108 \\
        eval\_in & - & 46021 & 20355 & 10448 & 15218 & 0 \\
        eval\_out & - & 46021 & 0 & 0 & 0 & 46021\\
        eval & 14.1 & 92042 & 20355 & 10448 & 15218 & 46021 \\
        \bottomrule
        \end{tabular}
    \end{small}
\caption{Number of records in the canonical partitioning of the synthetic dataset. The color notation is the same as in Figure \ref{fig:ds_partitioning} and indicates the data segments from which the partitions are sampled.}
\label{table:ds_no_records_synthetic}
\end{table}

\begin{table}[htp]
\centering
    \begin{small}
        \begin{tabular}{l|cccccc}
        \toprule
        Data & pct (\%) & total & \textcolor{magenta}{Group 1} & \textcolor{cyan}{Group 2} &  \textcolor{yellow}{Group 3} &  \textcolor{red}{Group 4} \\
        \midrule
        train    & 80.2 &  530706 &    236401 &    119084 &    175221 &         0 \\
        dev\_in   & -    &   18368 &      8182 &      4122 &      6064 &         0 \\
        dev\_out  & -    &   18368 &         0 &         0 &         0 &     18368 \\
        dev      & 5.6  &   36736 &      8182 &      4122 &      6064 &     18368 \\
        eval\_in  & -    &   47227 &     21037 &     10597 &     15593 &         0 \\
        eval\_out & -    &   47227 &         0 &         0 &         0 &     47227 \\
        eval     & 14.3 &   94454 &     21037 &     10597 &     15593 &     47227 \\
        \bottomrule
        \end{tabular}
    \end{small}
\caption{Number of records in the canonical partitioning of the real dataset. The color notation is the same as in Figure \ref{fig:ds_partitioning} and indicates the data segments from which the partitions are sampled.}
\label{table:ds_no_records_real}
\end{table}

\begin{figure}[htp]
 \centering
     \begin{subfigure}
         \centering
         \includegraphics[width=0.3\textwidth]{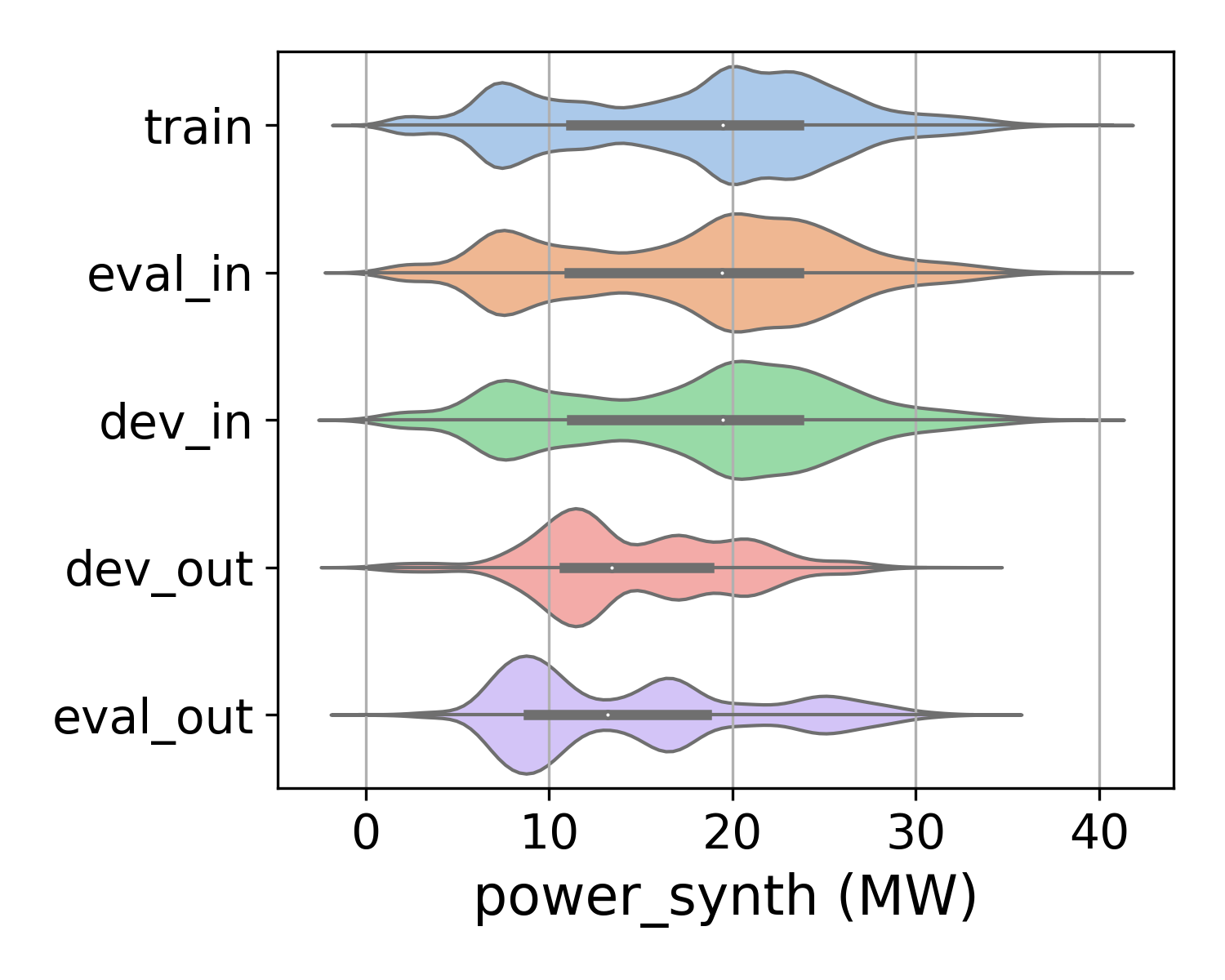}
     \end{subfigure}
     \hfill
     \begin{subfigure}
         \centering
         \includegraphics[width=0.3\textwidth]{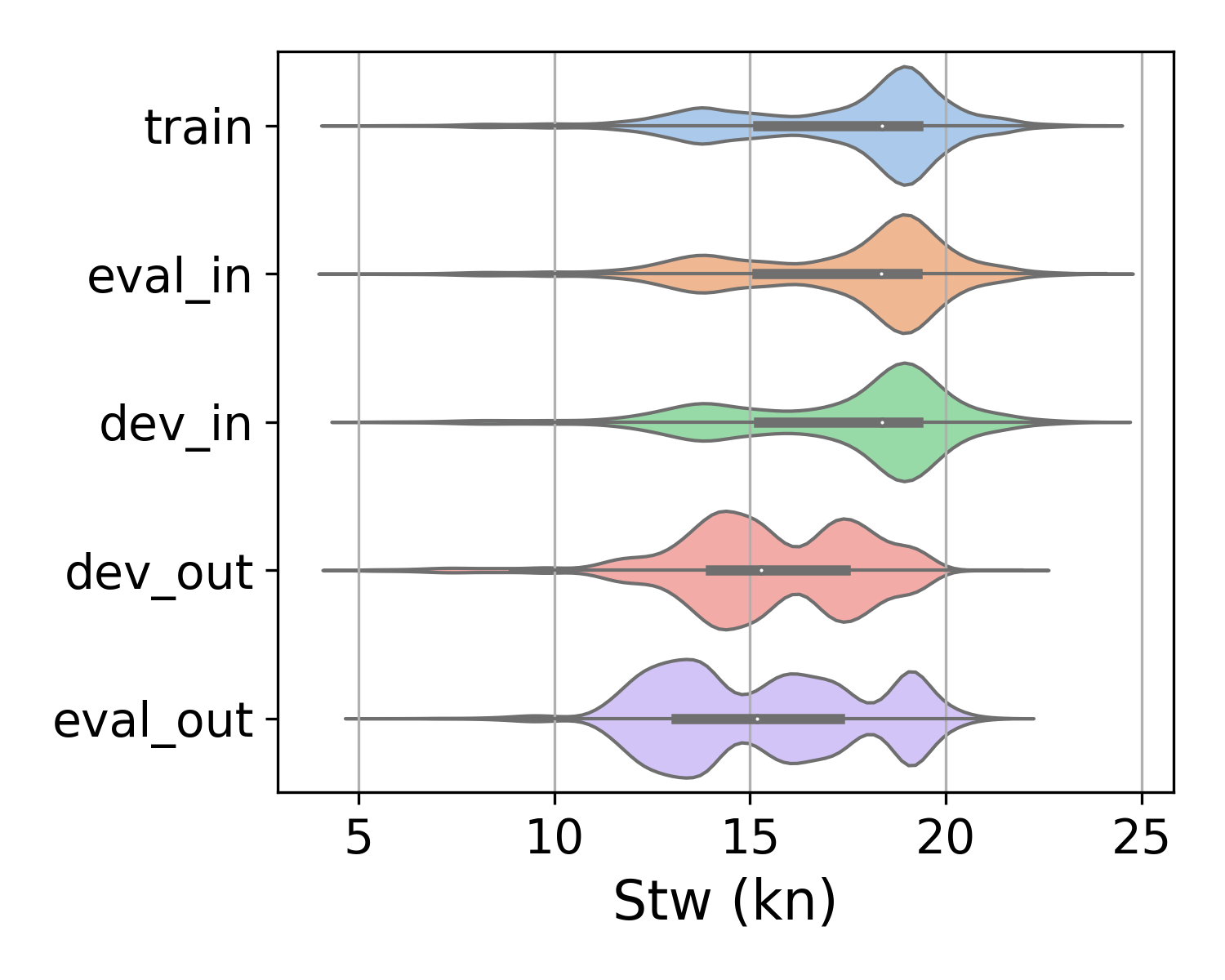}
     \end{subfigure}
     \hfill
     \begin{subfigure}
         \centering
         \includegraphics[width=0.3\textwidth]{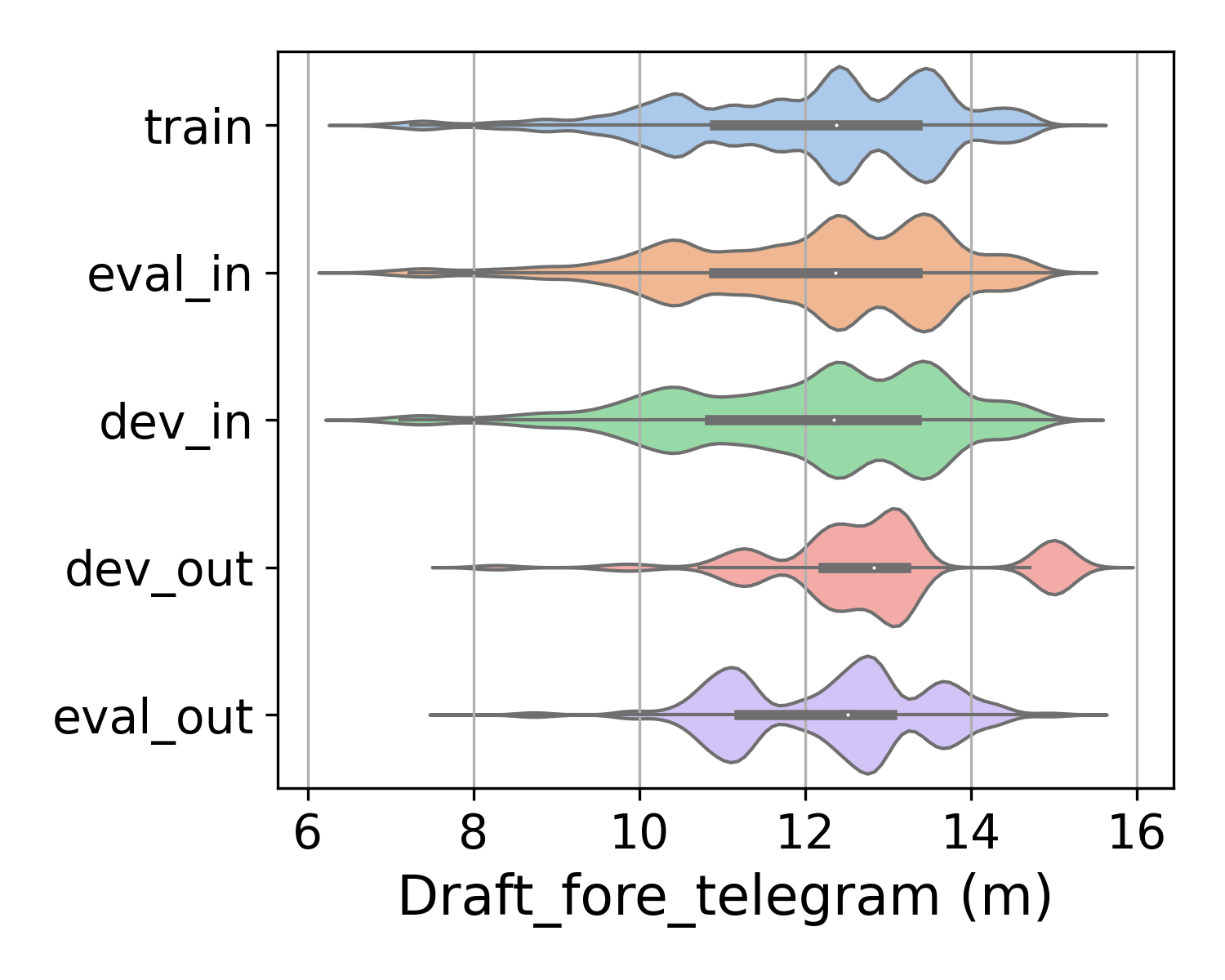}
     \end{subfigure}
     
     \begin{subfigure}
         \centering
         \includegraphics[width=0.3\textwidth]{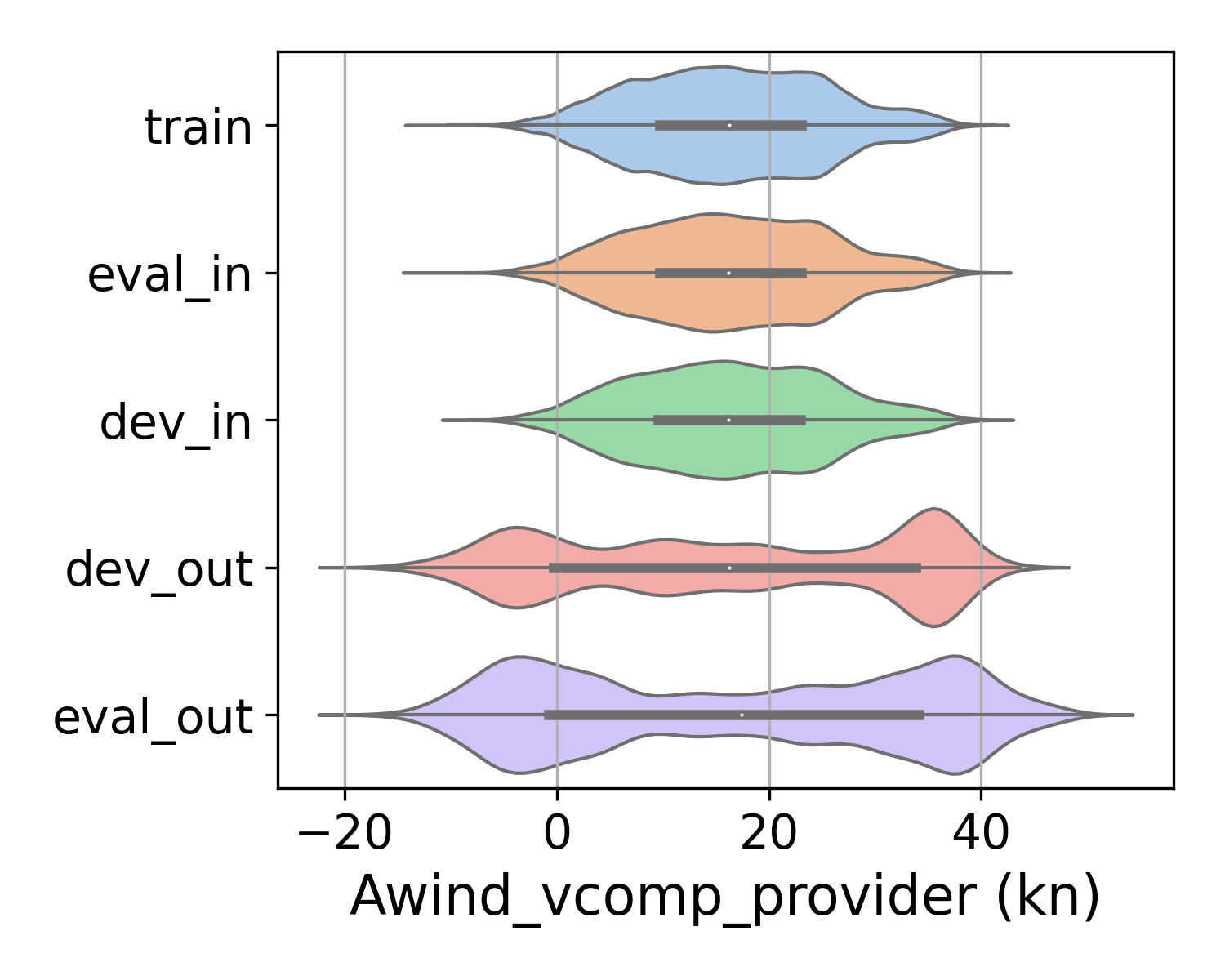}
     \end{subfigure}
     \hfill
     \begin{subfigure}
         \centering
         \includegraphics[width=0.3\textwidth]{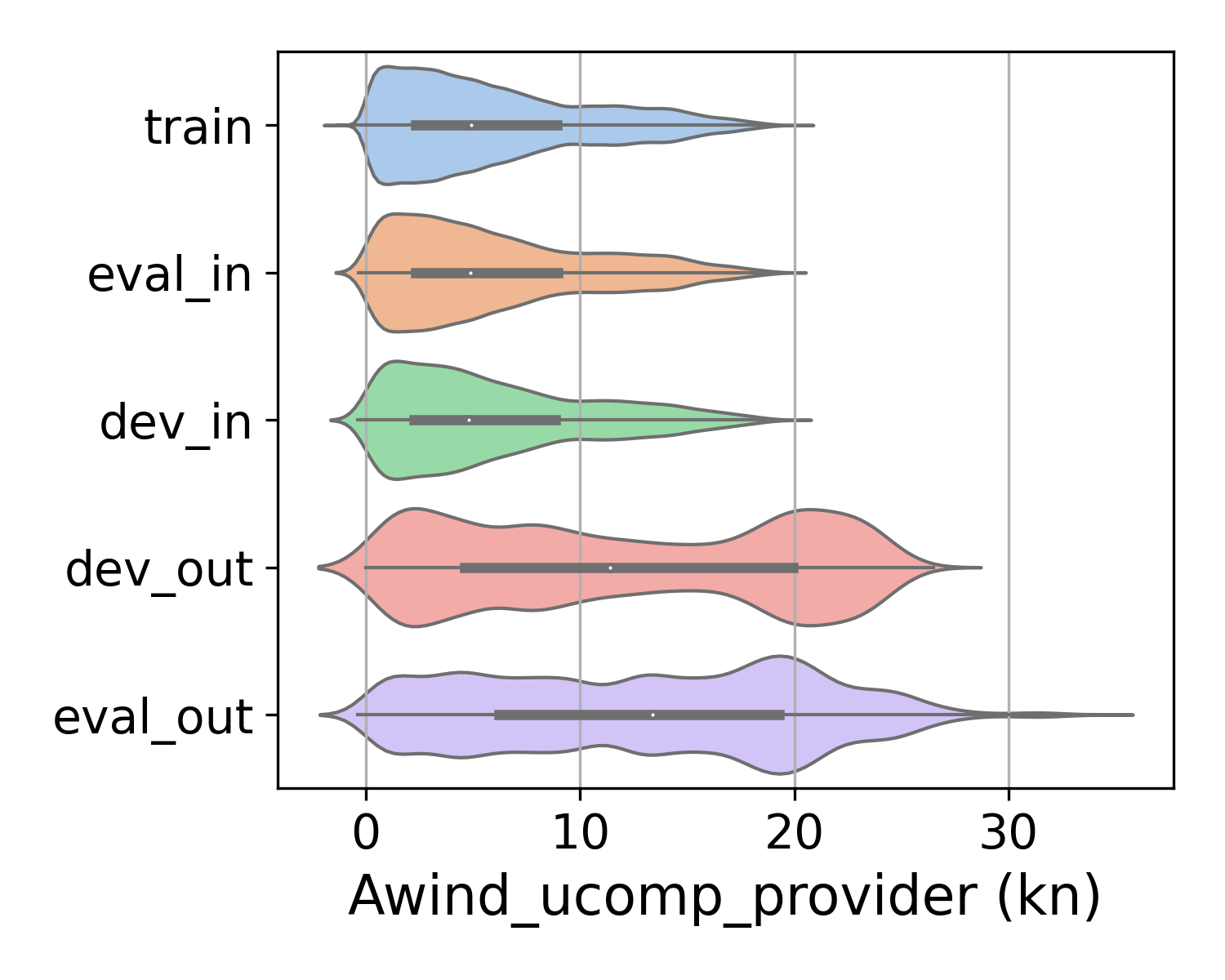}
     \end{subfigure}
     \hfill
     \begin{subfigure}
         \centering
         \includegraphics[width=0.3\textwidth]{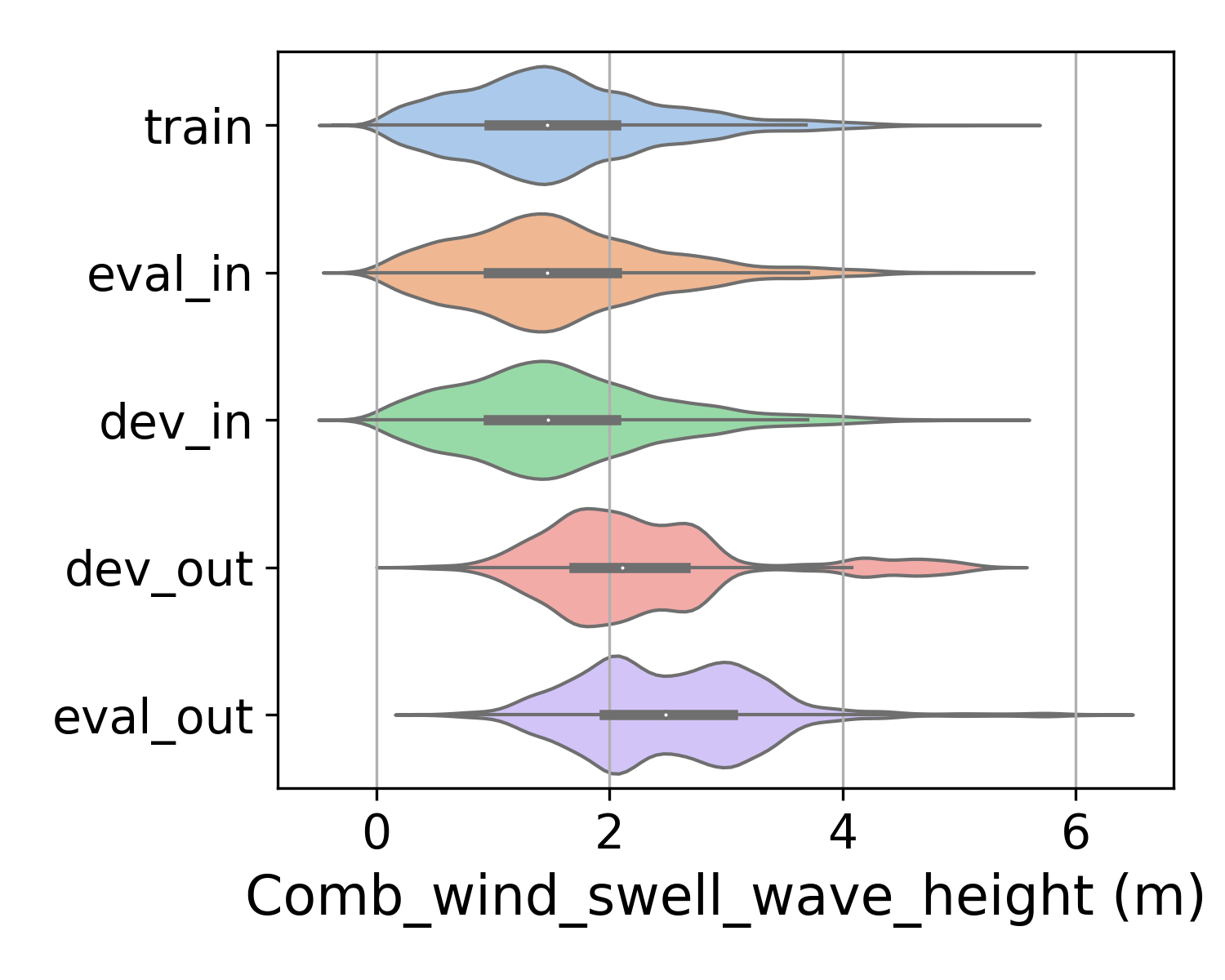}
     \end{subfigure}
\caption{Violin plots for the canonical partitions of the synthetic dataset after the noise injection (scaled to have the same width for better visualization).}
\label{fig:ds_violins_synthetic}
\end{figure}

\begin{figure}[htp]
 \centering
     \begin{subfigure}
         \centering
         \includegraphics[width=0.3\textwidth]{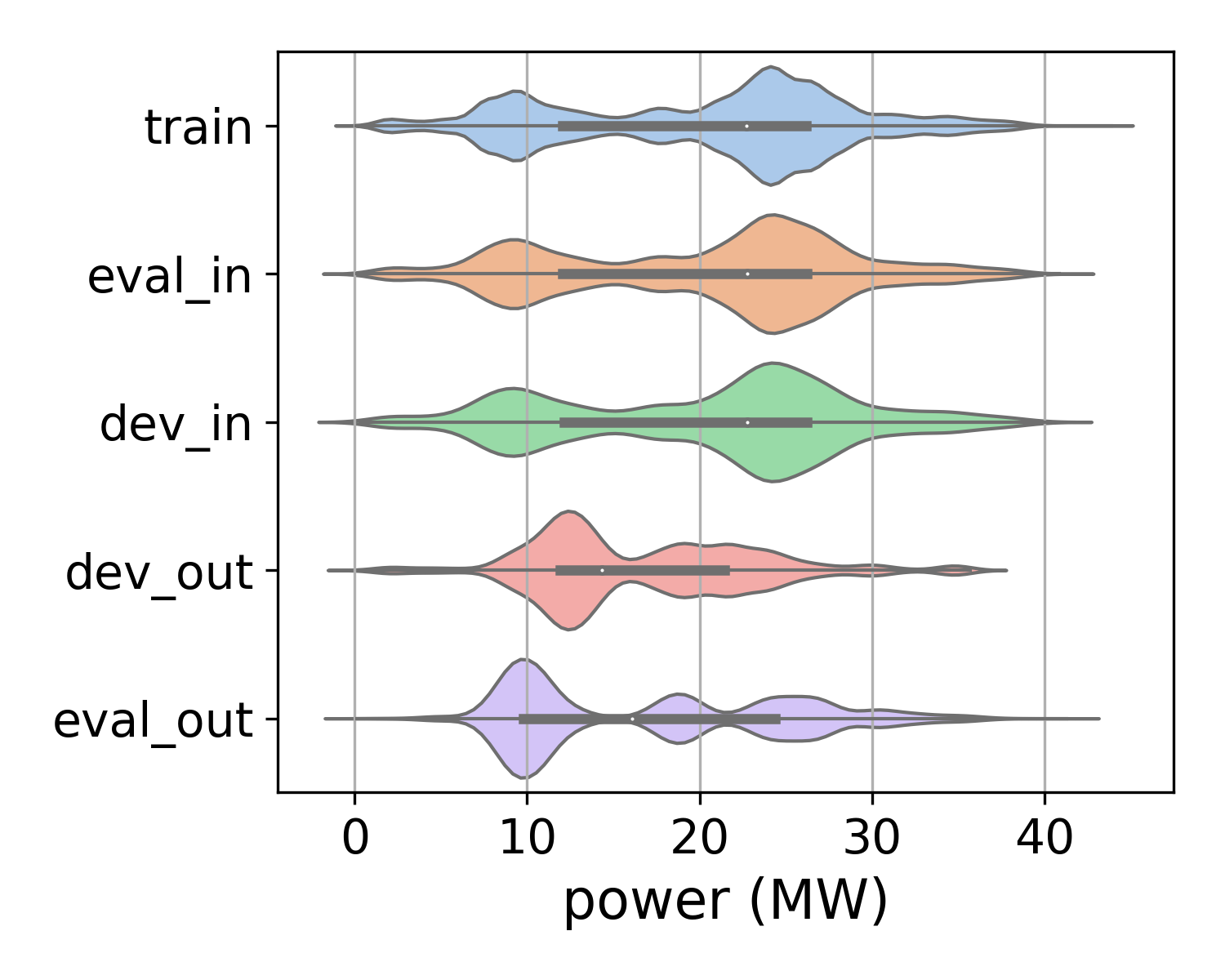}
     \end{subfigure}
     \hfill
     \begin{subfigure}
         \centering
         \includegraphics[width=0.3\textwidth]{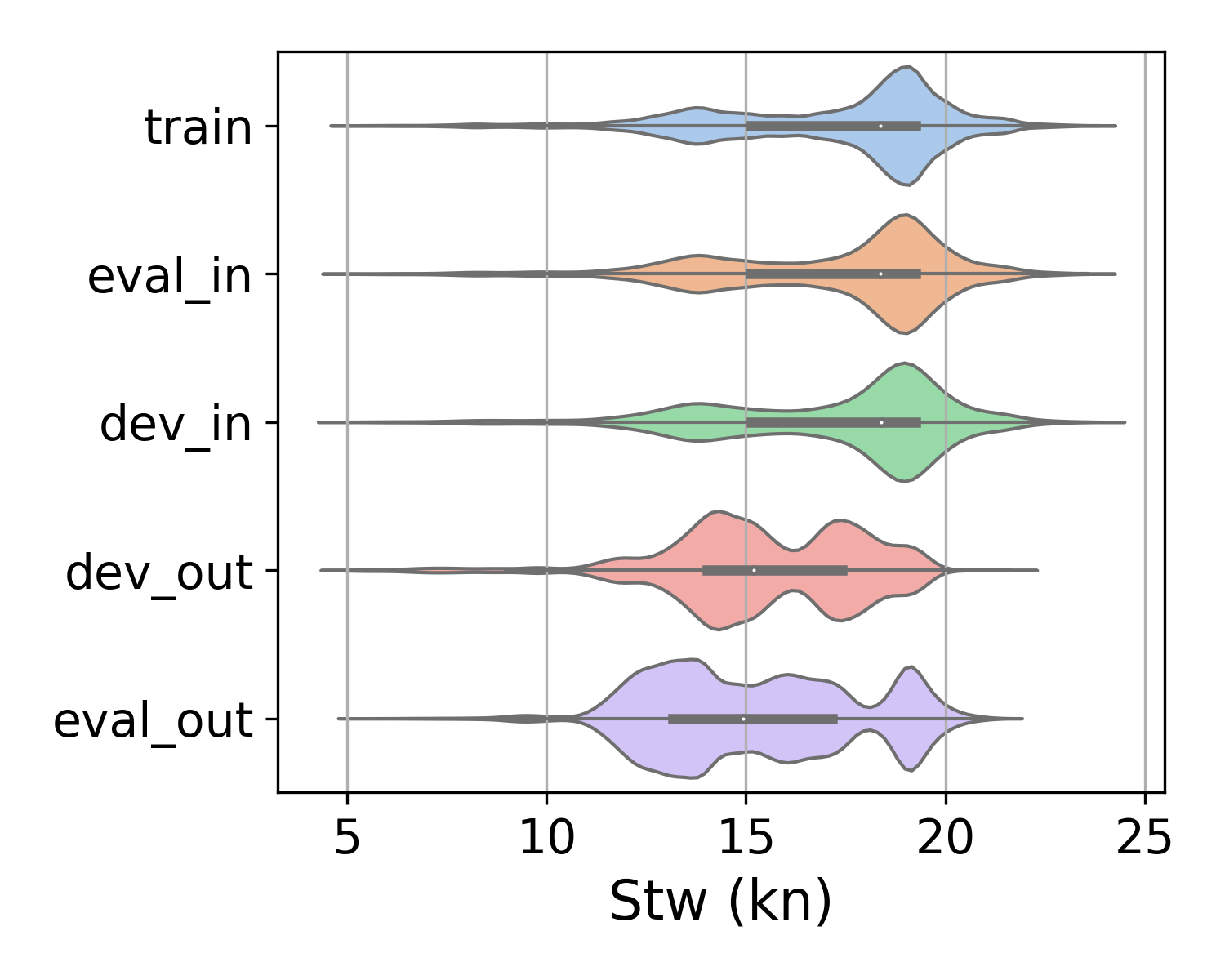}
     \end{subfigure}
     \hfill
     \begin{subfigure}
         \centering
         \includegraphics[width=0.3\textwidth]{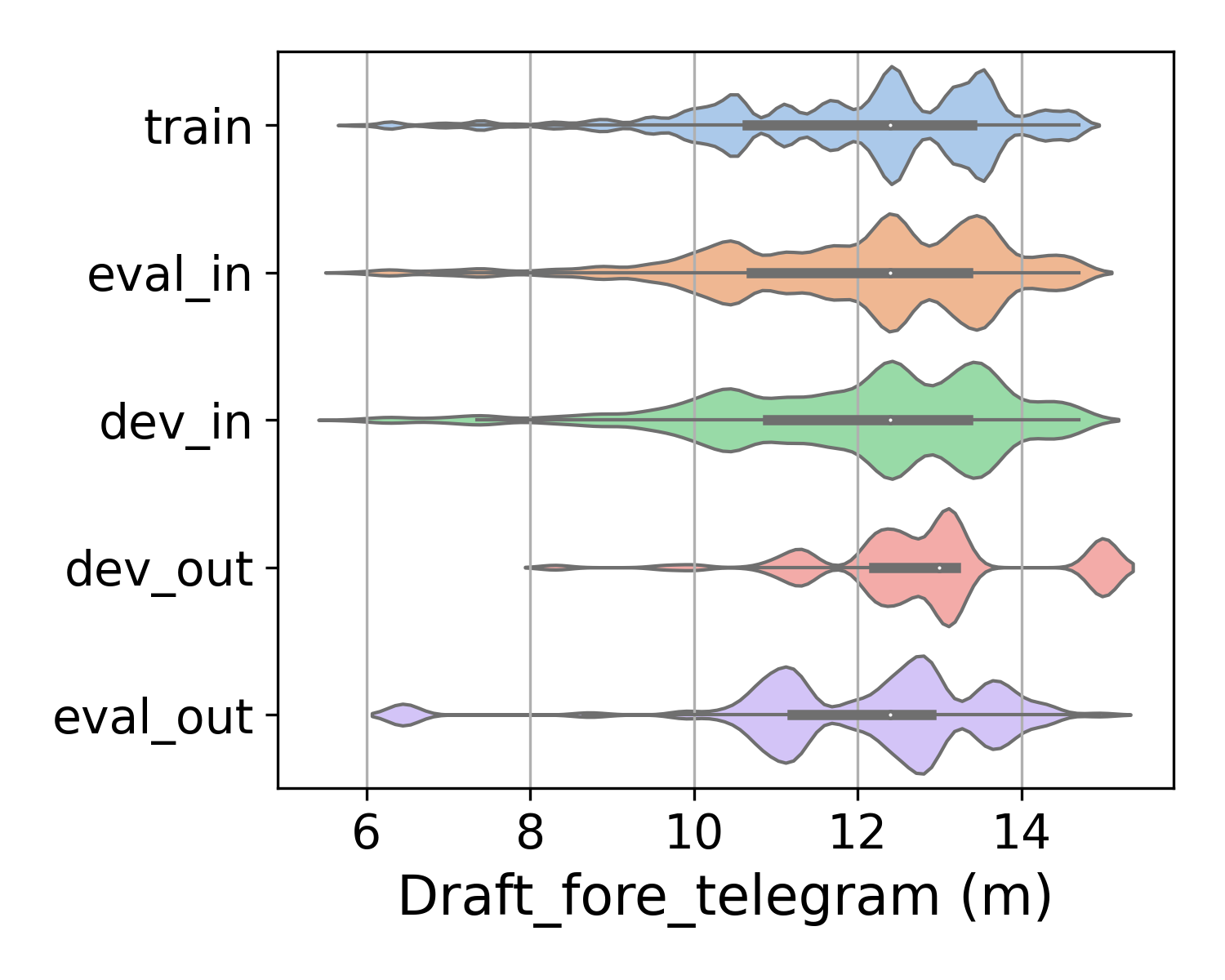}
     \end{subfigure}
     
     \begin{subfigure}
         \centering
         \includegraphics[width=0.3\textwidth]{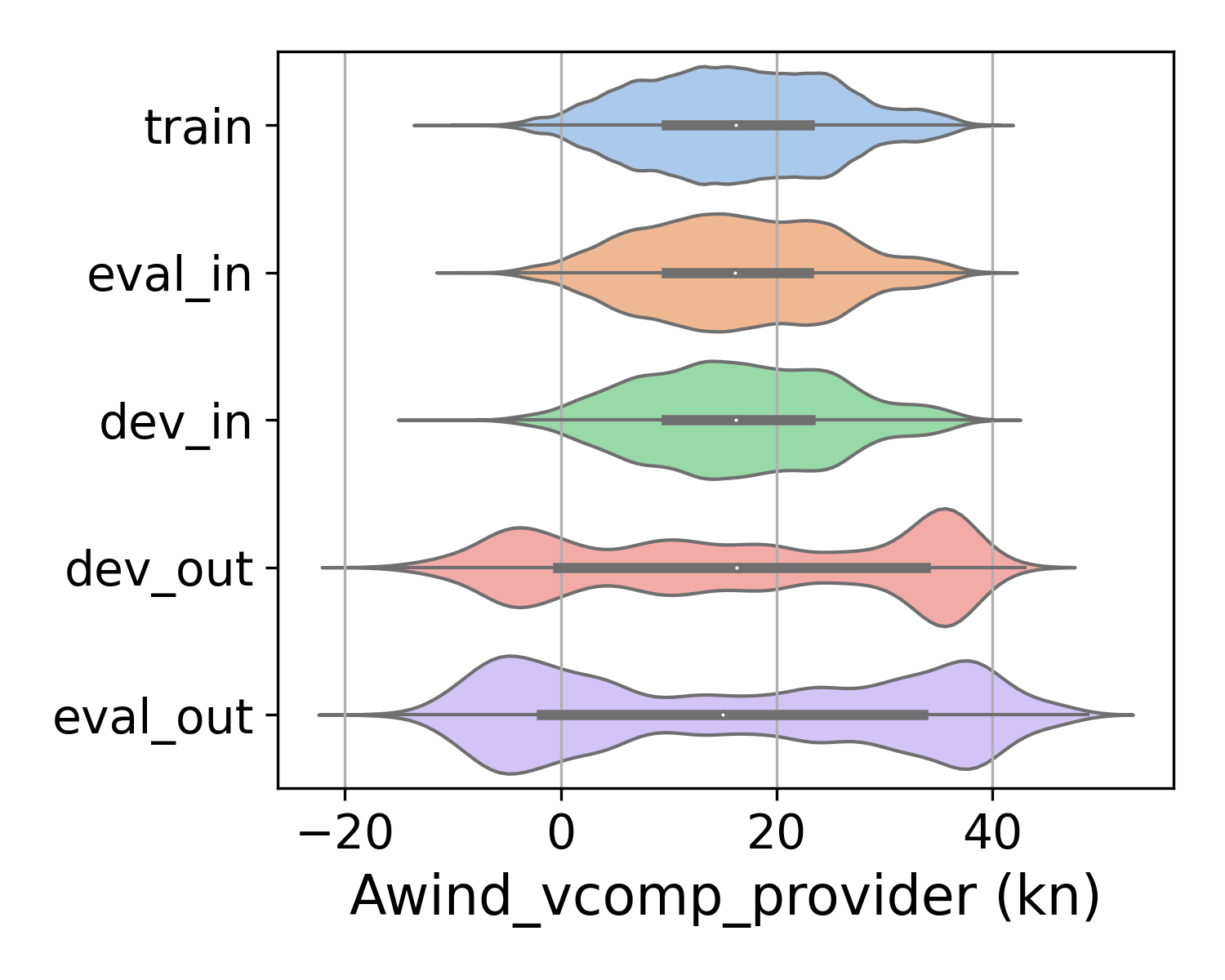}
     \end{subfigure}
     \hfill
     \begin{subfigure}
         \centering
         \includegraphics[width=0.3\textwidth]{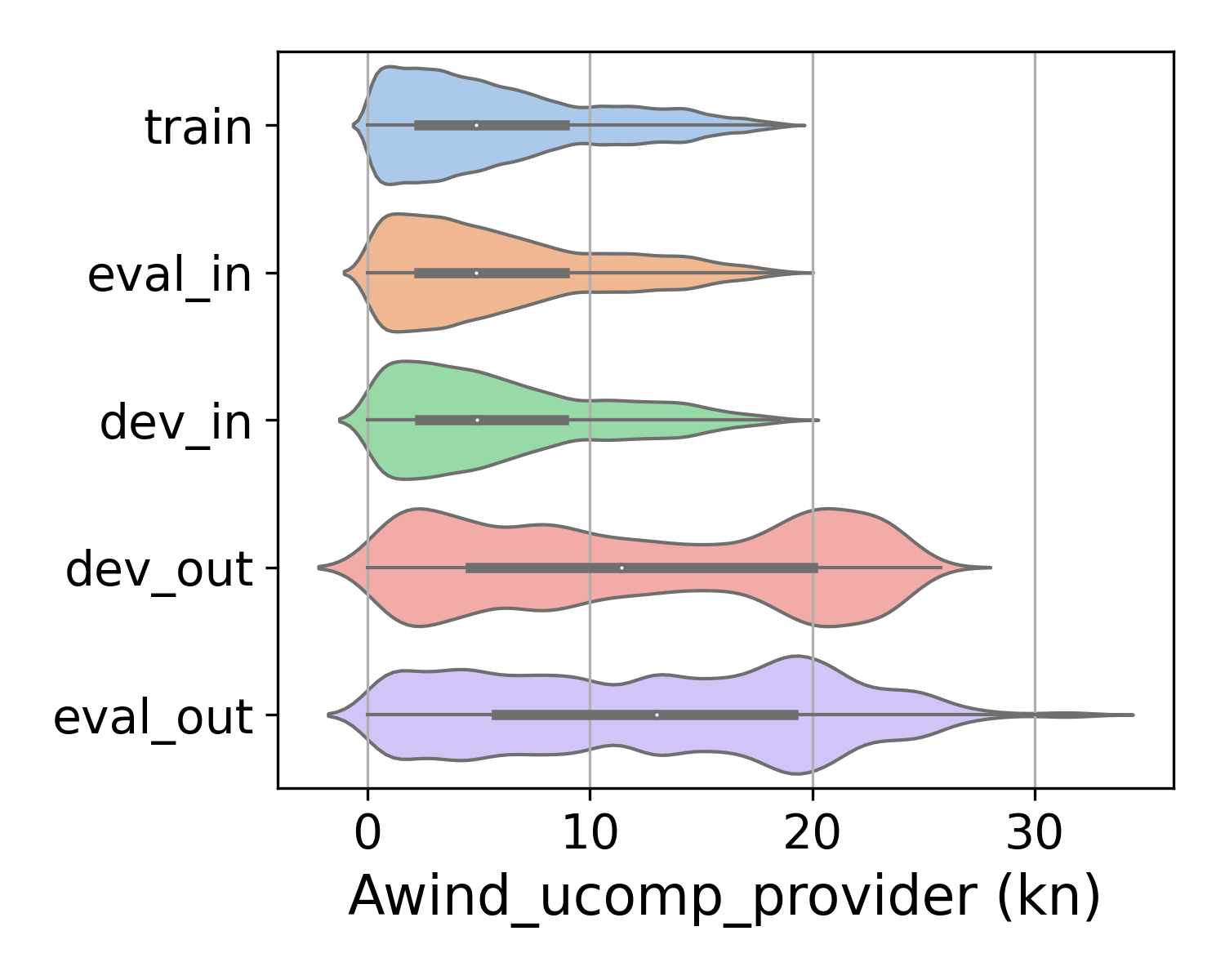}
     \end{subfigure}
     \hfill
     \begin{subfigure}
         \centering
         \includegraphics[width=0.3\textwidth]{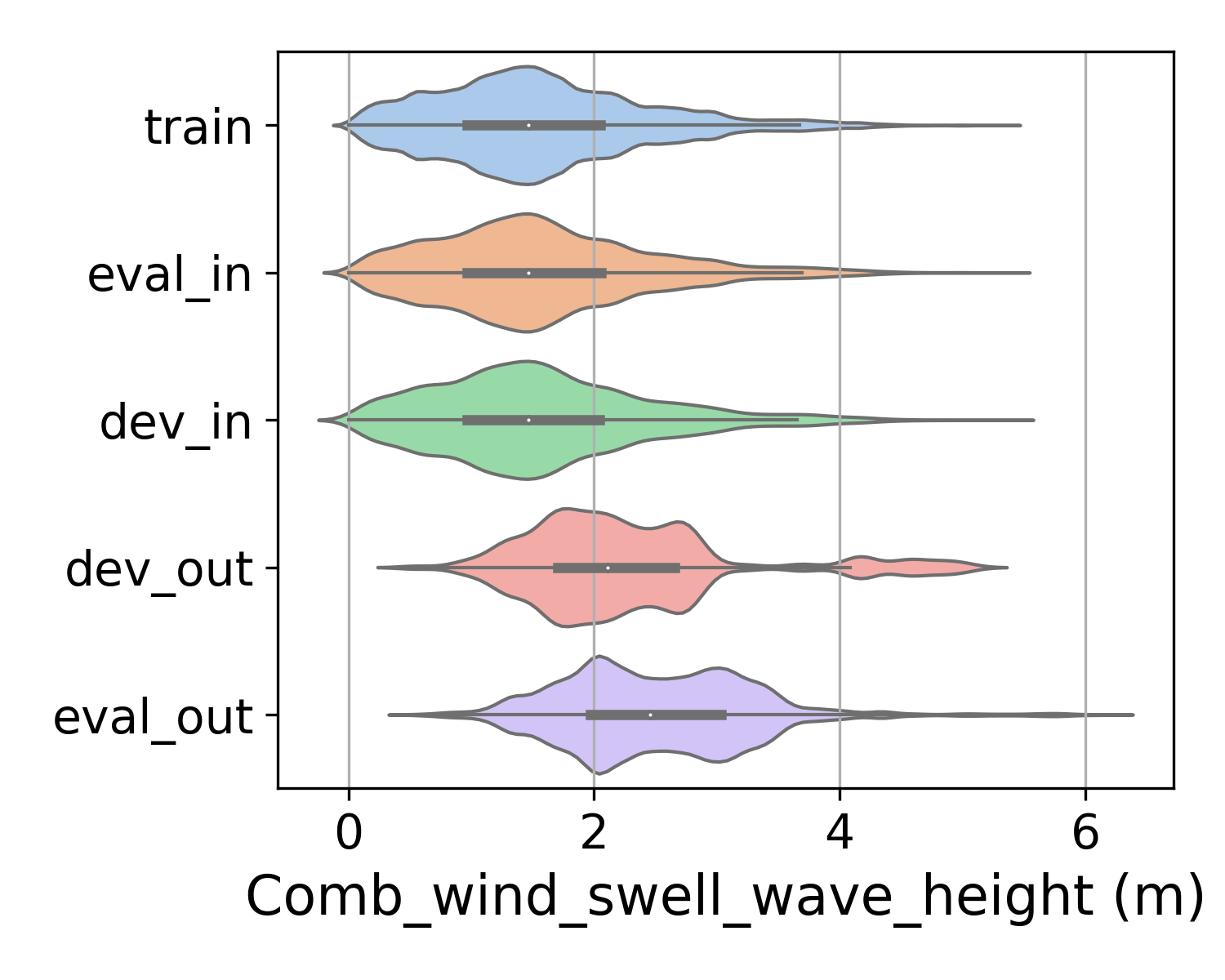}
     \end{subfigure}
\caption{Violin plots for the canonical partitions of the real dataset (scaled to have the same width for better visualization).}
\label{fig:ds_violins_real}
\end{figure}

\paragraph{Data analysis} The violin plots of the features for the canonical partitions for the synthetic dataset (Figure \ref{fig:ds_violins_synthetic}) and the real dataset (Figure \ref{fig:ds_violins_real}), demonstrate the comparability of the in-domain subsets and the distributional changes that are seen in the out-of-domain partitions, particularly for the target and wind related features.

\paragraph{Synthetic Data Generation}
For the synthetic dataset, real and sampled features are combined with power labels predicted a synthetic, physics-based model. The synthetic model  \cite{hullpic2022} is a generative function ($f_{synthetic}$) which takes as input a time-series of features (i.e. signals), as recorded from a real vessel, and calculates the power consumed by the vessel’s hull. This function finds the propeller cooperation point after calculating all the components of resistance (bare hull, appendages, wind, waves, fouling drag) for given speed, draft and trim. More specifically, for the generation of synthetic data, a non-linear solver script was created to find the operating point of a given propeller and hull resistance for each desired condition, as described by Bose \cite{bose2008marine}. The propeller curves (KT, KQ) can either be user defined or use the B-Series \cite{van1969wageningen}. For the resistance part, the calculation of each component can be described as follows: having the full hydrostatics table of the vessel for the whole range of drafts and trims, along with a series of geometric characteristics (bulb shape and size, transom, appendages etc), calm water resistance is calculated by employing the Holtrop method \cite{holtrop1982approximate} for slender ships (i.e. containers, RoRo, gas carriers) and Modified Holtrop \cite{nikolopoulos2019study} is used for bulkier ships like large Tankers and Bulk carriers. Following the ISO 15016 \cite{ships2015marine}, the weather added resistance is found by calculating the wind effect by using the regressions of Fujiwara \cite{fujiwara2006cruising}, while the wave effects are modelled according to STAwave1 and STAwave2 as also introduced by Tsujimoto \cite{tsujimoto2008practical}. Hull interaction factors are calculated depending on ship type, using empirical formulas, a summary of which can be found in Carlton \cite{carlton2018marine}. Scale effect corrections, cavitation criteria and corrections were also taken from Carlton \cite{carlton2018marine} and Bertram \cite{bertram2012chapter} . The effect of wake affecting energy saving devices can be modelled by adjusting the interaction factors. Fine-tuning of the method to fit a specific vessel (when there is not enough hydrostatic data, or discrepancies are observed), can be done by using sea trial data and/or detailed factors when available from a towing tank report, or actual measurements of well known conditions. Last but not least, the effect of fouling is modelled as the result of its manifestations (drag, propeller and interaction). The change in drag coefficient is modelled after Townsin \cite{townsin1981estimating}, the effect of fouling on the propeller performance is modelled as in Seo \cite{seo2016study} (increase in torque coefficient), as also described in Carlton \cite{carlton2018marine} and the change of interaction factors are modelled after Farkas \cite{farkas2020impact}. All the aforementioned models produce the effect of fouling on each component over time, which is measured from each drydock / cleaning event. While this allows for a sophisticated modelling of the interaction of features and power used, it still nevertheless a model which is simpler than reality and has fewer factors of variation. 

One of the key goals of this research is to look into the quality of uncertainty estimation both within and outside of domain areas. Working with a synthetic dataset allows for the insertion of well-controlled noise patterns, which should be reflected in the model's heteroscedastic predictive uncertainty~\cite{malinin-thesis}. To make the synthetic set realistic for this task, we apply two types of Gaussian noise with non-constant variance (heteroscedasticity) to the synthetic target $y_i$:
\begin{itemize}
    \item heteroscedastic Gaussian noise correlated with power,  $\varepsilon_{power, i}=N(0, a\cdot y_i)$. This type of noise simulates the scenario of linear deterioration of the torque meter accuracy as power increases,
    \item heteroscedastic Gaussian noise correlated with true wind speed, $\varepsilon_{wind,i} =N(0, b \cdot w_i)$.  Synthetic data is partitioned based on true wind speed, therefore adding the noise wind with variance linearly increasing with wind speed, simulates an increasing data uncertainty as we move from the in-domain partitions to out-of-domain ones. The goal of this approach is to capture the empirical observation that the most severe wind conditions encountered in the dataset are the most uncertain.
\end{itemize}

Here, $i = 1, \cdots, M$ stands for the i-th record, w is the true wind speed, $a=0.025$ (at power 40 MW the standard deviation of heteroscedastic power noise is 1MW) and $b=25$ (at wind speed 40 kn the standard deviation of heteroscedastic wind noise is 1MW). The synthetic power with noise is defined as: 
$$y'_i = y_i + \varepsilon_{power, i} + \varepsilon_{wind,i}$$     
\begin{table}[htp]
\centering
    \begin{small}
        \begin{tabular}{cc}
        \toprule
        Feature & $\sigma$ \\
        \midrule
        draft\_aft\_telegram & 0.15 m \\
        draft\_fore\_telegram & 0.15 m \\
        stw & 0.25 kn \\
        diff\_speed\_overground & 0.25 kn/3min \\
        awind\_speed\_provider & 0.5 kn \\
        rcurrent\_vcomp & 0.05 kn \\
        rcurrent\_ucomp & 0.05 kn \\
        comb\_wind\_swell\_wave\_height & 0.1 m \\
        \bottomrule
        \end{tabular}
    \end{small}
\caption{Standard deviation of the added Gaussian noise per input feature.}
\label{table:ds_input_noise}
\end{table}

\begin{figure}[htp]
\centering
    \includegraphics[scale=0.4]{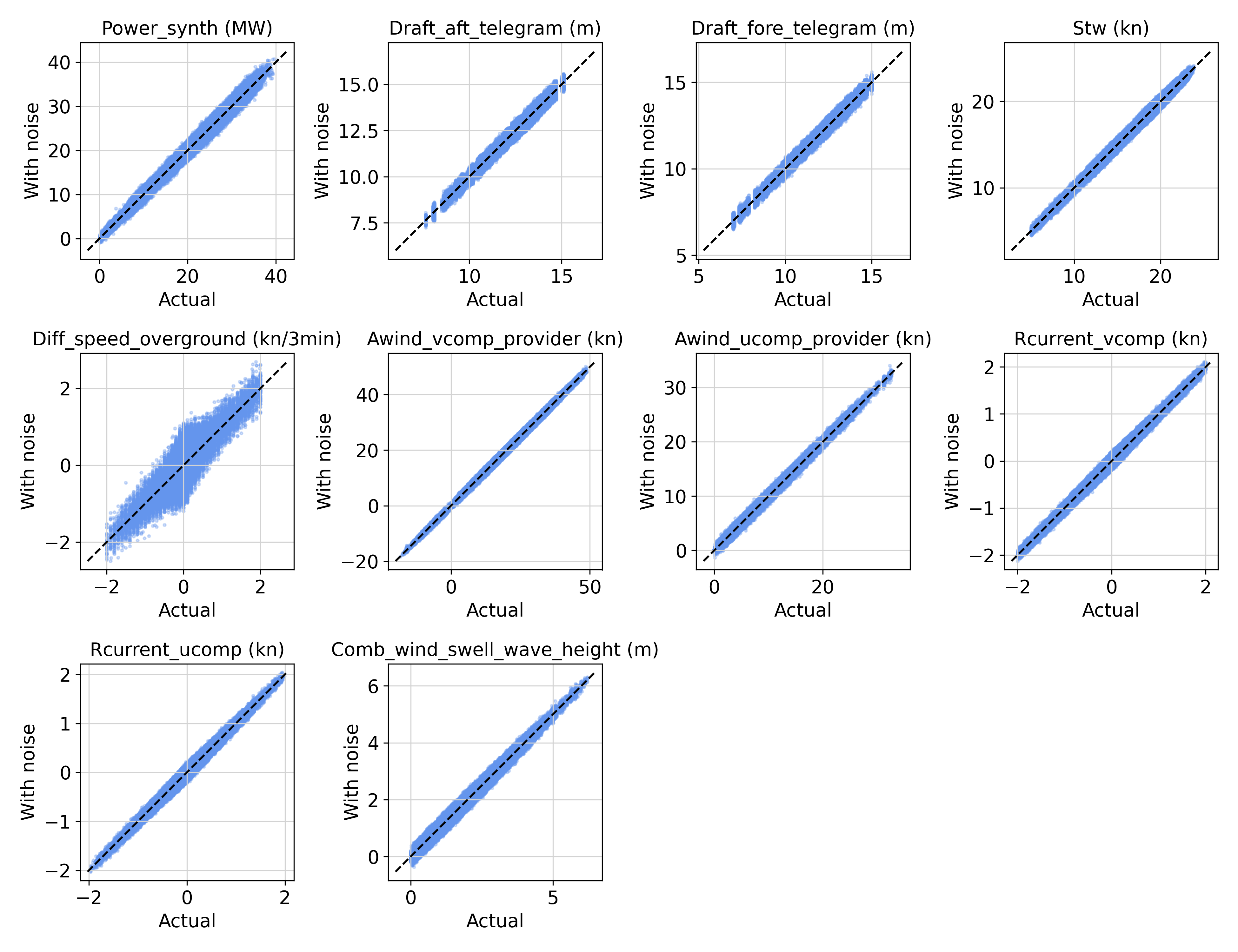}
\caption{Correlation plots illustrating the effect of injected noise per feature.}
\label{fig:ds_corr_plots}
\end{figure}

Furthermore, to emulate the effect of signal intrinsic noise coming from the data gathering process, we add Gaussian white noise $N(0,\sigma)$ to the training features (sensor noise, weather hindcast errors, transmission errors to name a few sources of inherent data variability). The standard deviation per feature (Table \ref{table:ds_input_noise}) is determined using the average expected noise magnitude of these signals. The effect of the injected noise on the data variance is illustrated in Figure \ref{fig:ds_corr_plots} via the correlation plots of the noisy data with the respective original signal per feature. 

\paragraph{Generalization set} In order to further evaluate the generalization capability of models under research in out-of-domain regions, we introduce for the first time, the notion of a generalization set as an augmented synthetic dataset (about 2.5 million records) by applying independently uniform sampling on input features from a predefined range of values shown in Table \ref{table:ds_generalization_sampling_range}. The idea of sampling from the convex hull of the full range of possible conditions (operational and weather) is a necessary condition to assure that the measure of performance is robust. For each sample point, power labels are generated by the synthetic model which by design can cover the complete feature space. The idea is depicted in Figures \ref{fig:ds_regular_data_split} and \ref{fig:ds_gen_set_idea}. Independently uniform sampling of features enables the creation of new records with feature vectors not regularly or even not physically possible to be met. For instance, in regular vessel operation, there is a low probability for a vessel to have high speed in extreme weather conditions. Beyond navigational preferences, confounding features such as apparent winds and waves  is another source of  spurious correlations in the dataset that could lead to  a biased model. 

In cases where the downstream task is implemented by a combinatorial optimization algorithm  (e.g weather routing) while the model produces the cost function, it is critical that the model is unbiased across the board, even under unlikely conditions. Because the optimization algorithm actively searches the space of all feasible states, it may select a poorly modeled one as optimal and drive the entire solution in an entirely wrong direction.

Having a biased dataset apart from training unreliable models also prevents from detecting such model as performance metrics also affected from the same biases. As the test set is practically following the same (biased) distribution with the rest of the dataset, a model could  show good performance while in reality would fail to generalize or even worse to properly disentagle all the causal factors received as input. Uniform sampling assures that all possible conditions are equally represented practically eliminating spurious correlation between input features. Also allows for measuring performance consistently even with the classic metrics (e.g mse, mae etc.) by providing an unbiased estimate of the performance of a model across the board.

The generalization set has two important characteristics making it suitable for model evaluation:
\begin{enumerate}
    \item There are no correlations between the input features (both causal and spurious correlations).
    \item The generalization set is suitable for evaluating model performance both in and out of domain, covering a wide range of operational conditions.
\end{enumerate}

\begin{table}[h!]
\centering
    \begin{small}
        \begin{tabular}{cc}
        \toprule
        Feature & Range \\
        \midrule
        speed over ground & [5, 23] kn \\
        draft aft & [8, 15] m  \\
        draft fore & [8, 15] m  \\
        true wind speed & [0, 40] kn \\
        relative wind angle & [0, 360] degrees \\
        current speed & [0, 2] kn \\
        relative current angle & [0, 360] degrees \\
        waves & [0, 6] m \\
        \bottomrule
        \end{tabular}
    \end{small}
\caption{Range of values of the input features used to create the generalization set by uniform
sampling.}
\label{table:ds_generalization_sampling_range}
\end{table}

\paragraph{Format} The data will be shared as several comma-separate value (CSV) files.

\subsection{Description of features and targets}\label{deepsea_features}

\begin{table}[H]
\centering
\scalebox{0.95}{
    \begin{tabular}{llm{4.7cm}l}
        \toprule
        Feature name & Units & \multicolumn{1}{l}{Description} & Source \\ 
        \midrule
        draft\_aft\_telegram & m & Draft at stern as reported by crew in daily reports & Telegrams \\
        draft\_fore\_telegram & m & Draft at bow as reported by crew in daily reports  & Telegrams\\      
        stw & kn & Speed through water (i.e. relative to any currents) of the vessel as measured by speed log & Onboard sensor\\
        diff\_speed\_overground & kn/3min & Acceleration of the vessel relative to ground & GPS \\
        awind\_vcomp\_provider & kn & Apparent wind speed component relative to the vessel along its direction of motion & Weather provider \\
        awind\_ucomp\_provider & kn & Apparent wind speed component relative to vessel perpendicular to its direction & Weather provider\\
        rcurrent\_vcomp & kn & Component of currents relative to the vessel along its direction of motion & Weather provider\\
        rcurrent\_ucomp & kn & Component of currents relative to vessel perpendicular to its direction& Weather provider \\
        comb\_wind\_swell\_wave\_height & m  & Combined wave height due to wind and sea swell & Weather provider\\
        timeSinceDryDock & minutes & Time since the last dry dock cleaning of the vessel & Calculated \\ 
        time\_id & $-$ & Run number representing time. It may be used as an index of the records & Calculated \\
        \bottomrule
    \end{tabular}
    }
\caption{Description of the input features.} 
\end{table}

\begin{table}[H]
\centering
% \scalebox{1.}{
    \begin{tabular}{llll}
        \toprule
        Feature name & Units & Description & Source \\ 
        \midrule
        power & kW & Propeller shaft power as measured by torquemeter & Onboard sensor \\
        power\_synth & kW & Synthetic power generated by the synthetic model & Estimated \\
        \bottomrule

    \end{tabular}
    % }
\caption{Description of the targets.} 
\end{table}

\subsection{Training details}\label{apn:ship-baselines}

To evaluate the proposed dataset partitioning through the prism of uncertainty, we use the following methods in the form of an ensemble, that are able to capture both epistemic and aleatoric uncertainty:  
\begin{itemize}
    \item Deep ensemble of 10 variational inference neural networks (Deep Ensemble VI)
    \item Deep ensemble of 10 Monte-Carlo (MC) dropout  \cite{Gal2016Dropout} neural networks (Deep Ensemble MC dropout)
    \item Ensemble of 10 deep neural networks (Ensemble DNN)
    \item A proprietary domain-constrained model is also introduced in the form of an ensemble of 10 dense neural-symbolic networks \cite{tsamoura2021neural} incorporating specific domain knowledge priors derived from the physics of the problem (Ensemble Symbolic). Domain specific knowledge is encoded via known relationships between input and output features, for example the cubic relationship between speed and power. Such physics priors are integrated with the rest of the network in a neural-symbolic fashion and as a result the model is still trained end-to-end like a normal deep regression model.
\end{itemize}

Each model outputs two parameters, the predicted mean and the predicted standard deviation of the conditional Normal distribution of the target (power) given the input. The variance of the predicted means across the members of the ensemble corresponds to the epistemic uncertainty and the mean of the predicted variances across the members is the measure of aleatoric uncertainty of the ensemble \cite{malinin2021shifts}). 

For all the methods except for the proprietary model (i.e Ensemble Symbolic) we use the same architecture: 2 hidden layers with 50 and 20 nodes and softplus activation function. The output layer has 2 nodes and a linear activation function. To satisfy the constraint of positive standard deviation the second output is fed through a softplus function and a constant $10^{-6}$ is added for numerical stability as proposed by \cite{deepensemble2017}. For the VI method we use Bayesian inference layers with Gaussian priors. They implement the Flipout estimator \cite{wen2018flipout} which performs a Monte Carlo approximation of the distribution. During inference for both the Deep Ensemble VI and Deep Ensemble MC dropout we sample 10 times each member of the ensemble (100 samples in total) to estimate the epistemic uncertainty. For the Ensemble DNN and Ensemble Symbolic model we use only the members of the ensemble to estimate the epistemic uncertainty.

Furthermore, we consider the single model version of the DNN and Symbolic methods. Both versions they only capture aleatoric uncertainty. For the VI  and MC dropout methods we also consider a simpler version of them by using a single seed model that is sampled 10 times during inference to capture the epistemic uncertainty.They referred as VI Ensemble (instead of Deep Ensemble) and MC dropout Ensemble respectively.

For optimization, we use the negative log likelihood loss function and the Adam optimizer with a learning rate of $10^{-4}$. The number of epochs is defined by early stopping, monitoring the mean absolute error (MAE) of the dev\_in set. The models are implemented in Tensorflow 2.

\subsection{Additional Results}\label{apn:ship-results}

\paragraph{Synthetic dataset} The performance metrics for the canonical partitions of the synthetic dataset and generalization set are presented in Tables  \ref{ds_errors_synthetic} and \ref{ds_retention_synthetic}. A single model metric $mean \pm \sigma$ is computed across the individual metric scores of all members. 

For the dev and eval sets, Ensemble DNN has the best predictive performance (Table \ref{ds_errors_synthetic}). Model ranking changes remarkably when considering the generalization set, with the Ensemble Symbolic having the best performance scores, showing percentage difference 18.5\% in terms of RMSE from the second best model that is the Ensemble DNN. Taking into account that the performance scores on the generalization set cover the whole feature space and are unbiased by construction of the set (i.e uniform sampling eliminates operational preferences and/or spurious correlations among features), the Ensemble Symbolic is expected to be the best candidate model deployed on unseen data, in terms of robustness. Another important observation is that the percentage differences of the scores between the models are significantly higher at the generalization set. This demonstrates that the generalization set can be an useful tool for model research and selection because it amplifies potentially insignificant variations in model performance when tested in a conventional dataset split.
 
Regarding the metrics that jointly assess robustness and predictive uncertainty (Table  \ref{ds_retention_synthetic}) it is observed that Ensemble DNN has the best scores in the generalization set. Ensemble DNN is not the best model in terms of robustness (Table \ref{ds_errors_synthetic}) and the fact that it takes the first place based on the retention metrics is an indication of considerable improvement of the quality of the uncertainty estimations (i.e better calibration of the predictive uncertainty with the error) in comparison to the Ensemble Symbolic. Same as before it is found that for the generalization set, the model ranking is well  defined  as there is a clear distinction of the scores across the models. This is not the case though for dev and eval sets, at which the top-2 models (Ensemble DNN and Ensemble VI) appear to have similar performance.

\begin{table}[!htp]
\centering
    \begin{small}
    \scalebox{0.7}{
    \setlength\tabcolsep{2pt}
        \begin{tabular}{l|l|l|llll|llll|llll}
            \toprule
            Dataset & Method & Model & \multicolumn{4}{c|}{RMSE (kW)} & \multicolumn{4}{c|}{MAE (kW)} & \multicolumn{4}{c}{MAPE (\%)} \\
            \multirow{7}{*}{ Dev } & & & In & Out & Full & Gen & In & Out & Full & Gen & In & Out & Full & Gen \\
            \midrule
            & MC dropout & Deep ensemble & 1082 & 1064 & \textbf{1073} & 1498 & 825 & 834 & \textbf{830} & 1000 & \textbf{6.16} & 7.63 & 6.91 & 23.00 \\
            & VI & Deep ensemble & \textbf{1081} & 1068 & 1075 & 1446 & \textbf{823} & 838 & \textbf{830} & 975 & 6.38 & 7.70 & 7.04 & 23.19 \\
            & DNN & Ensemble & 1088 & \textbf{1062} & 1075 & 1427 & 827 & \textbf{832} & \textbf{830} & 953 & 6.24 & \textbf{7.49} & \textbf{6.87} & \textcolor{red}{\textbf{21.21}} \\
            & Symbolic & Ensemble & 1132 & 1126 & 1129 & \textcolor{red}{\textbf{1204}} & 851 & 864 & 858 & \textcolor{red}{\textbf{873}} & 7.32 & 8.94 & 8.13 & 27.33 \\
            & MC dropout & Ensemble & $1091_{\pm 3}$ & $1074_{\pm 4}$ & $1082_{\pm 3}$ & $1526_{\pm 84}$ & $832_{\pm 2}$ & $842_{\pm 3}$ & $837_{\pm 3}$ & $1023_{\pm 35}$ & $6.35_{\pm 0.20}$ & $7.69_{\pm 0.09}$ & $7.03_{\pm 0.13}$ & $24.54_{\pm 1.04}$ \\
            & VI & Ensemble & $1085_{\pm 4}$ & $1074_{\pm 6}$ & $1079_{\pm 4}$ & $1458_{\pm 38}$ & $825_{\pm 3}$ & $842_{\pm 5}$ & $833_{\pm 2}$ & $985_{\pm 22}$ & $6.42_{\pm 0.18}$ & $7.75_{\pm 0.07}$ & $7.08_{\pm 0.12}$ & $23.72_{\pm 1.26}$ \\
            & DNN & Single & $1096_{\pm 7}$ & $1081_{\pm 10}$ & $1089_{\pm 8}$ & $1487_{\pm 52}$ & $834_{\pm 5}$ & $846_{\pm 6}$ & $840_{\pm 6}$ & $1008_{\pm 29}$ & $6.34_{\pm 0.26}$ & $7.67_{\pm 0.10}$ & $7.01_{\pm 0.15}$ & $23.66_{\pm 1.62}$ \\
            & Symbolic & Single & $1134_{\pm 2}$ & $1129_{\pm 5}$ & $1132_{\pm 3}$ & $1213_{\pm 27}$ & $853_{\pm 1}$ & $866_{\pm 4}$ & $860_{\pm 2}$ & $879_{\pm 19}$ & $7.33_{\pm 0.05}$ & $8.96_{\pm 0.06}$ & $8.15_{\pm 0.05}$ & $27.54_{\pm 1.73}$ \\
            \midrule
            \multirow{4}{*}{ Eval }& MC dropout & Deep ensemble & \textbf{1069} & 1111 & 1090 & 1498 & 814 & 859 & 837 & 1000 & 6.26 & 6.95 & 6.59 & 23.00 \\
            & VI & Deep ensemble & \textbf{1069} & 1104 & \textbf{1086} & 1446 & \textbf{813} & 854 & \textbf{834} & 975 & 6.24 & 6.92 & 6.58 & 23.19 \\
            & DNN & Ensemble & 1076 & \textbf{1099} & 1087 & 1427 & 818 & \textbf{851} & \textbf{834} & 953 & \textbf{6.13} & \textbf{6.91} & \textbf{6.52} & \textcolor{red}{\textbf{21.21}} \\
            & Symbolic & Ensemble & 1117 & 1133 & 1125 & \textcolor{red}{\textbf{1204}} & 841 & 866 & 854 & \textcolor{red}{\textbf{873}} & 7.25 & 7.29 & 7.27 & 27.33 \\
            & MC dropout & Ensemble & $1078_{\pm4}$ & $1122_{\pm6}$ & $1100_{\pm5}$ & $1526_{\pm84}$ & $822_{\pm3}$ & $868_{\pm5}$ & $845_{\pm4}$ & $1023_{\pm35}$ & $6.34_{\pm0.08}$ & $7.05_{\pm0.08}$ & $6.70_{\pm0.07}$ & $24.54_{\pm1.04}$ \\
            & VI & Ensemble & $1072_{\pm3}$ & $1109_{\pm6}$ & $1090_{\pm4}$ & $1458_{\pm38}$ & $815_{\pm2}$ & $858_{\pm4}$ & $837_{\pm3}$ & $985_{\pm22}$ & $6.28_{\pm0.13}$ & $6.96_{\pm0.06}$ & $6.62_{\pm0.09}$ & $23.72_{\pm1.26}$ \\
            & DNN & Single & $1084_{\pm6}$ & $1116_{\pm18}$ & $1100_{\pm12}$ & $1487_{\pm52}$ & $825_{\pm5}$ & $864_{\pm14}$ & $844_{\pm10}$ & $1008_{\pm29}$ & $6.24_{\pm0.13}$ & $7.04_{\pm0.17}$ & $6.64_{\pm0.14}$ & $23.66_{\pm1.62}$ \\
            & Symbolic & Single & $1120_{\pm2}$ & $1137_{\pm5}$ & $1128_{\pm3}$ & $1213_{\pm27}$ & $843_{\pm1}$ & $869_{\pm4}$ & $856_{\pm2}$ & $879_{\pm19}$ & $7.26_{\pm0.04}$ & $7.30_{\pm0.03}$ & $7.28_{\pm0.03}$ & $27.54_{\pm1.73}$ \\
            \midrule
        \end{tabular}
        }
    \end{small}
\caption{Predictive performance for the canonical partitions of the synthetic dataset and the generalization set. One standard deviation is quoted for the single seed results.}
\label{ds_errors_synthetic}
\end{table}

\begin{table}[!htp]
\centering
    \begin{small}
    \scalebox{0.6}{
    \setlength\tabcolsep{2pt}
        \begin{tabular}{l|l|l|llll|llll|llll}
            \toprule
            Dataset & Method & Model & \multicolumn{4}{c|}{R-AUC $* 10^{5}$} & \multicolumn{4}{c|}{F1-AUC} & \multicolumn{4}{c}{F1@95\%} \\
            \multirow{7}{*}{ Dev } & & & In & Out & Full & Gen & In & Out & Full & Gen & In & Out & Full & Gen \\
            \midrule
            & MC dropout & Deep ensemble & 4.17 & 4.66 & 4.40 & 4.97 & 0.479 & 0.427 & 0.454 & 0.477 & 0.576 & 0.545 & 0.561 & 0.576 \\
            & VI & Deep ensemble & \textbf{4.03} & 4.54 & \textbf{4.26} & 4.32 & 0.491 & \textbf{0.433} & \textbf{0.465} & 0.506 & 0.579 & 0.544 & 0.563 & 0.582 \\
            & DNN & Ensemble & 4.08 & \textbf{4.50} & \textbf{4.26} & \textcolor{red}{\textbf{4.20}} & \textbf{0.492} & \textbf{0.433} & \textbf{0.465} &  \textcolor{red}{\textbf{0.509}} & \textbf{0.581} & \textbf{0.549} & \textbf{0.565} & 0.595 \\
            & Symbolic & Ensemble & 4.90 & 5.49 & 5.17 & 4.41 & 0.475 & 0.423 & 0.452 & 0.494 & 0.571 & 0.539 & 0.555 & \textcolor{red}{\textbf{0.596}} \\
            & MC dropout & Ensemble & $4.23_{\pm 0.05}$ & $4.70_{\pm 0.04}$ & $4.45_{\pm 0.04}$ & $5.39_{\pm 0.45}$ & $0.485_{\pm 0.002}$ & $0.428_{\pm 0.002}$ & $0.459_{\pm 0.001}$ & $0.481_{\pm 0.013}$ & $0.573_{\pm 0.002}$ & $0.541_{\pm 0.003}$ & $0.557_{\pm 0.003}$ & $0.560_{\pm 0.012}$ \\
            & VI & Ensemble & $4.05_{\pm 0.02}$ & $4.58_{\pm 0.04}$ & $4.29_{\pm 0.02}$ & $4.53_{\pm 0.31}$ & $0.490_{\pm 0.001}$ & $0.432_{\pm 0.001}$ & $0.464_{\pm 0.002}$ & $0.499_{\pm 0.011}$ & $0.578_{\pm 0.001}$ & $0.544_{\pm 0.002}$ & $0.562_{\pm 0.001}$ & $0.578_{\pm 0.012}$ \\
            & DNN & Single & $4.16_{\pm 0.06}$ & $4.68_{\pm 0.08}$ & $4.41_{\pm 0.07}$ & $5.27_{\pm 0.63}$ & $0.488_{\pm 0.003}$ & $0.430_{\pm 0.002}$ & $0.461_{\pm 0.003}$ & $0.471_{\pm 0.019}$ & $0.576_{\pm 0.003}$ & $0.544_{\pm 0.002}$ & $0.560_{\pm 0.003}$ & $0.568_{\pm0.012}$ \\
            & Symbolic & Single & $4.92_{\pm0.03}$ & $5.52_{\pm0.06}$ & $5.20_{\pm0.04}$ & $4.55_{\pm0.26}$ & $0.475_{\pm0.001}$ & $0.423_{\pm0.001}$ & $0.452_{\pm0.001}$ & $0.490_{\pm0.003}$ & $0.570_{\pm0.001}$ & $0.538_{\pm0.001}$ & $0.554_{\pm0.001}$ & $0.594_{\pm0.006}$ \\
            \midrule
            \multirow{4}{*}{ Eval } & MC dropout & Deep ensemble & 4.11 & 4.80 & 4.47 & 4.97 & 0.487 & 0.432 & 0.459 & 0.477 & 0.587 & 0.548 & 0.568 & 0.576 \\
            & VI & Deep ensemble & \textbf{3.97} & \textbf{4.59} & \textbf{4.29} & 4.32 & \textbf{0.497} & \textbf{0.441} & \textbf{0.470} & 0.506 & \textbf{0.589} & \textbf{0.549} & \textbf{0.570} & 0.582 \\
            & DNN & Ensemble & 4.02 & 4.60 & 4.32 & \textcolor{red}{\textbf{4.20}} & \textbf{0.497} & 0.439 & 0.469 & \textcolor{red}{\textbf{0.509}} & 0.588 & \textbf{0.549} & 0.569 & 0.595 \\
            & Symbolic & Ensemble & 4.82 & 5.42 & 5.09 & 4.41 & 0.484 & 0.426 & 0.458 & 0.494 & 0.579 & 0.548 & 0.564 & \textcolor{red}{\textbf{0.596}} \\
            & MC dropout & Ensemble & $4.17_{\pm0.05}$ & $4.87_{\pm0.06}$ & $4.54_{\pm0.06}$ & $5.39_{\pm0.45}$ & $0.491_{\pm0.001}$ & $0.435_{\pm0.001}$ & $0.463_{\pm0.001}$ & $0.481_{\pm0.013}$ & $0.582_{\pm0.002}$ & $0.542_{\pm0.003}$ & $0.561_{\pm0.003}$ & $0.560_{\pm0.012}$ \\
            & VI & Ensemble & $4.00_{\pm0.02}$ & $4.64_{\pm0.06}$ & $4.33_{\pm0.04}$ & $4.53_{\pm0.31}$ & $0.496_{\pm0.002}$ & $0.440_{\pm0.002}$ & $\boldsymbol{0.470_{\pm0.002}}$ & $0.499_{\pm0.011}$ & $0.588_{\pm0.002}$ & $0.546_{\pm0.002}$ & $0.568_{\pm0.002}$ & $0.578_{\pm0.012}$ \\
            & DNN & Single & $4.09_{\pm0.05}$ & $4.84_{\pm0.19}$ & $4.49_{\pm0.14}$ & $5.27_{\pm0.63}$ & $0.493_{\pm0.004}$ & $0.433_{\pm0.005}$ & $0.464_{\pm0.005}$ & $0.471_{\pm0.019}$ & $0.584_{\pm0.004}$ & $0.543_{\pm0.007}$ & $0.564_{\pm0.005}$ & $0.568_{\pm0.012}$ \\
            & Symbolic & Single & $4.84_{\pm0.02}$ & $5.46_{\pm0.05}$ & $5.13_{\pm0.03}$ & $4.55_{\pm0.26}$ & $0.483_{\pm0.001}$ & $0.425_{\pm0.001}$ & $0.457_{\pm0.0}$ & $0.490_{\pm0.003}$ & $0.579_{\pm0.0}$ & $0.547_{\pm0.002}$ & $0.563_{\pm0.001}$ & $0.594_{\pm0.006}$ \\
            \midrule
        \end{tabular}
        }
    \end{small}
\caption{Retention performance for the canonical partitions of the synthetic dataset and the generalization set. One standard deviation is quoted for the single seed results.}
\label{ds_retention_synthetic}
\end{table}

\begin{figure}[!htp]
\centering
    \begin{tabular}{c@{ }c@{ }c@{ }}
        Development & Evaluation & Generalization \\
        \includegraphics[width=.32\linewidth]{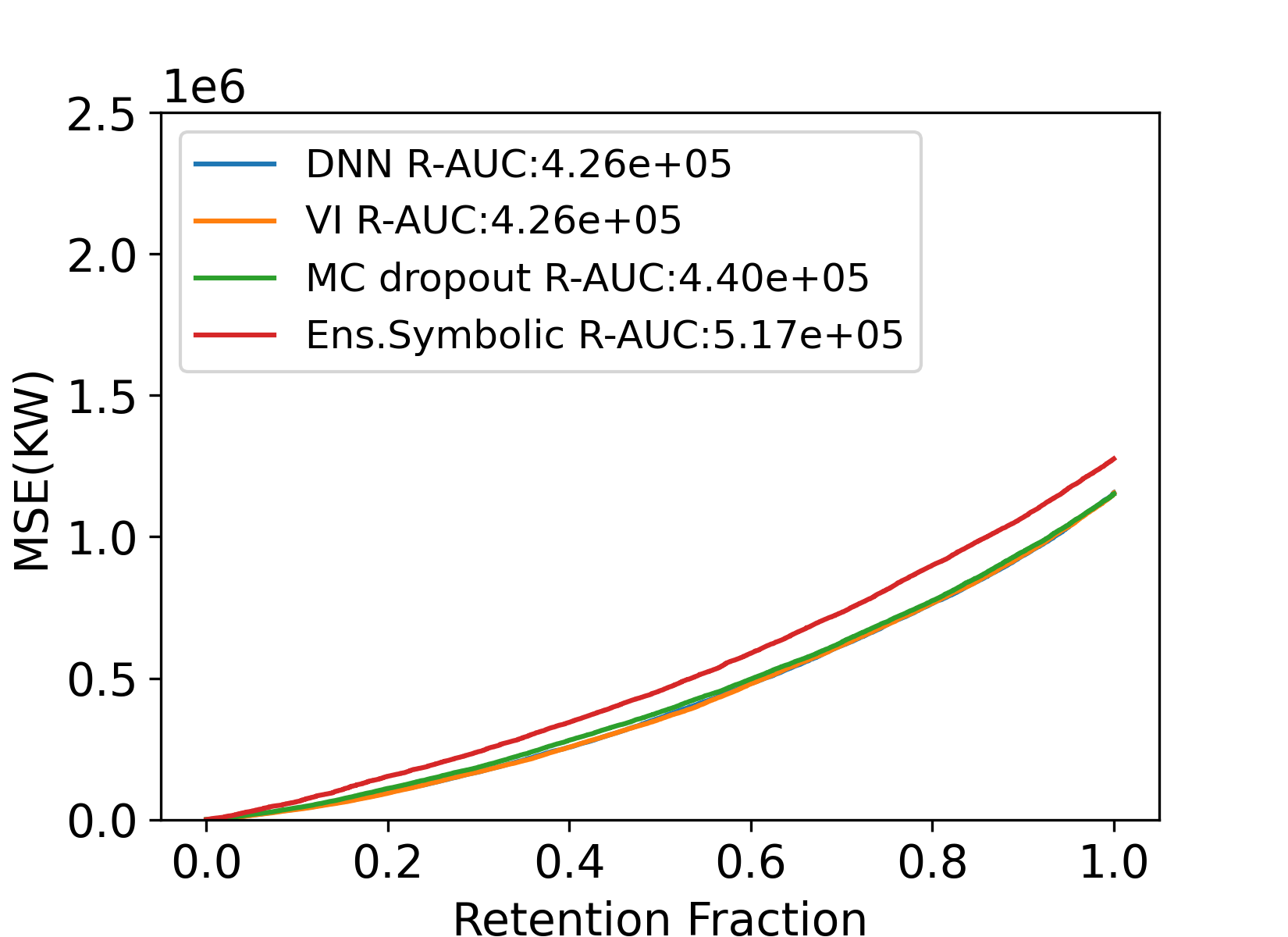}&
        \includegraphics[width=.32\linewidth]{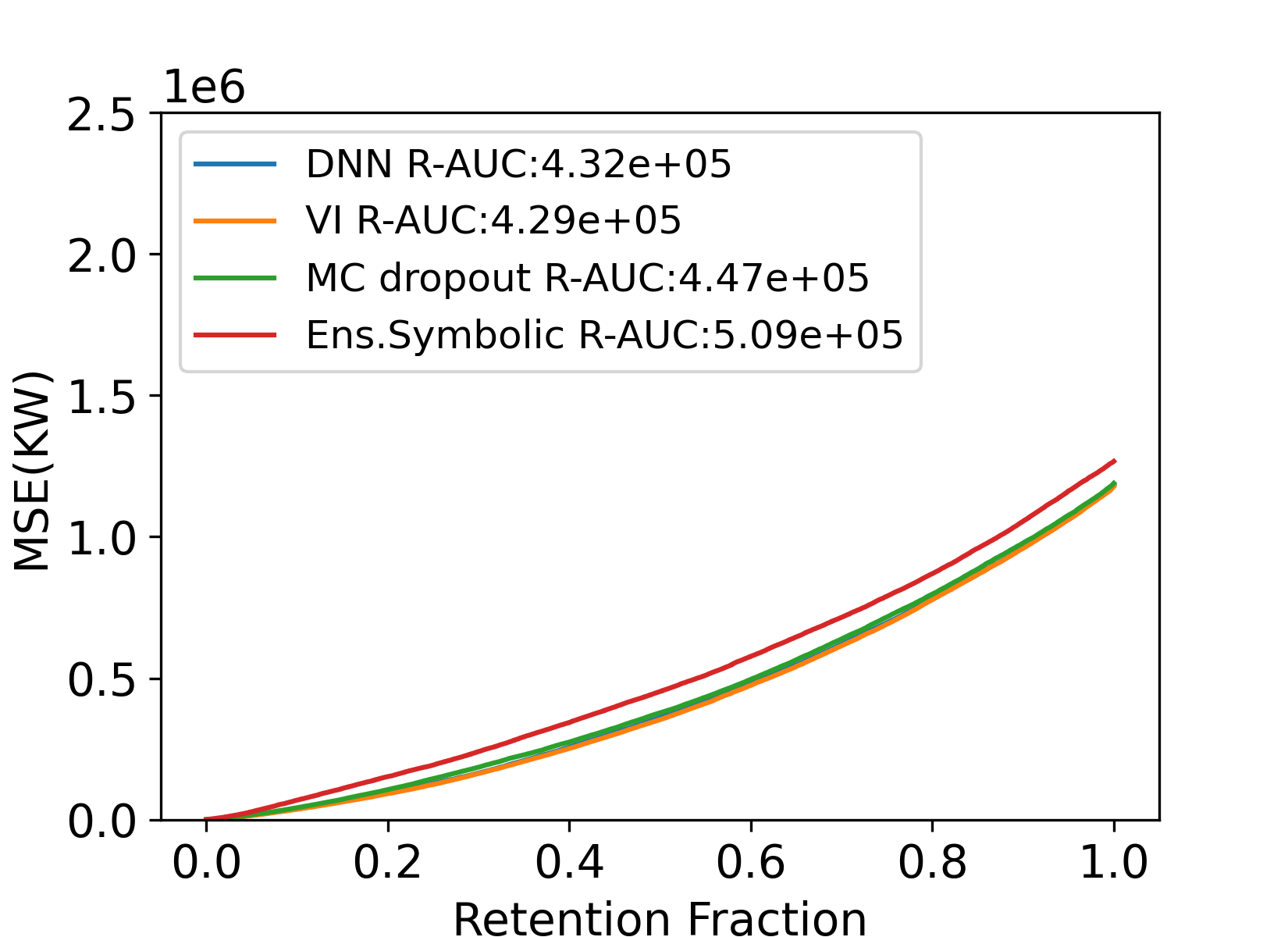}&
        \includegraphics[width=.32\linewidth]{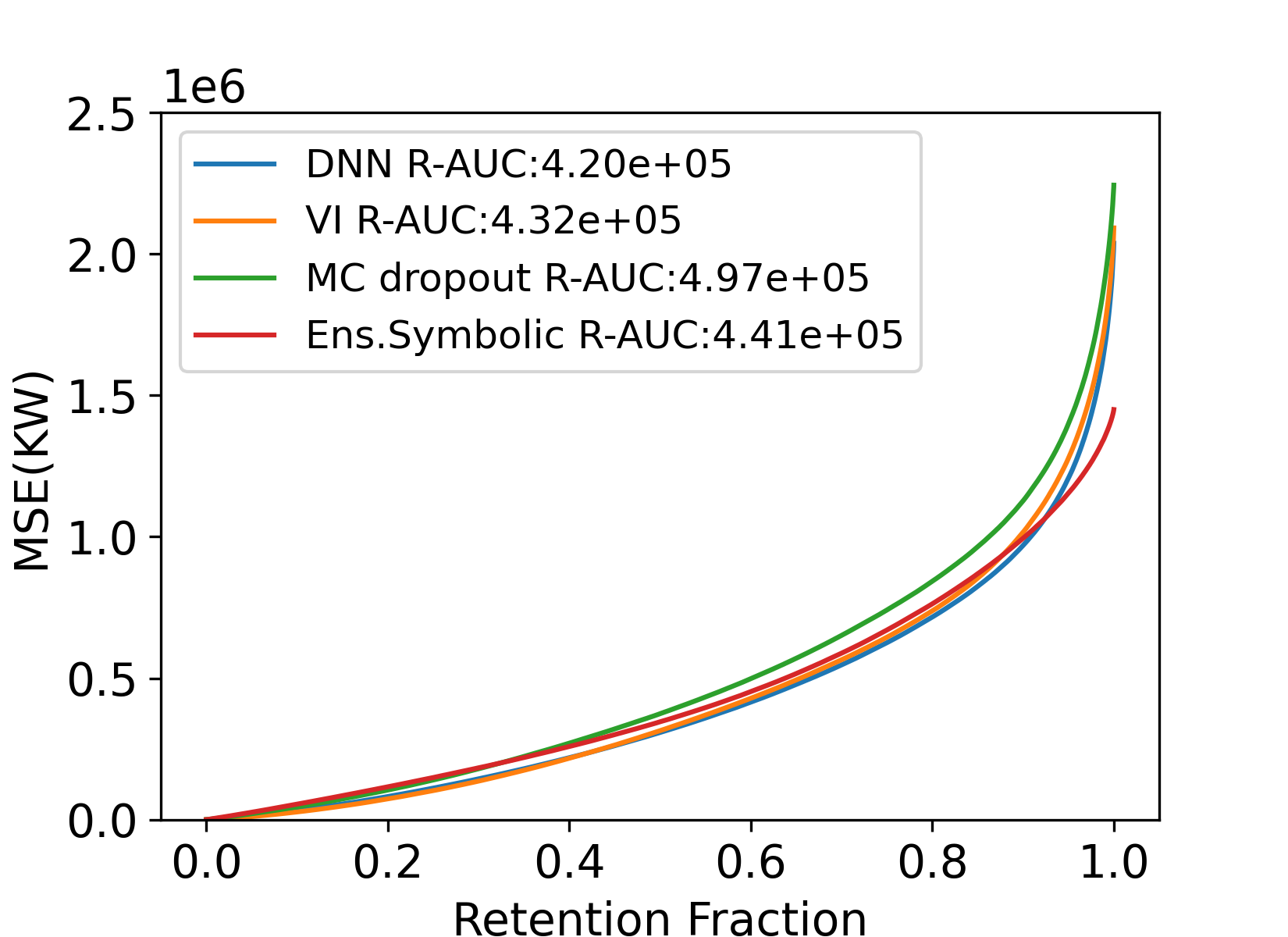}\\
        \includegraphics[width=.32\linewidth]{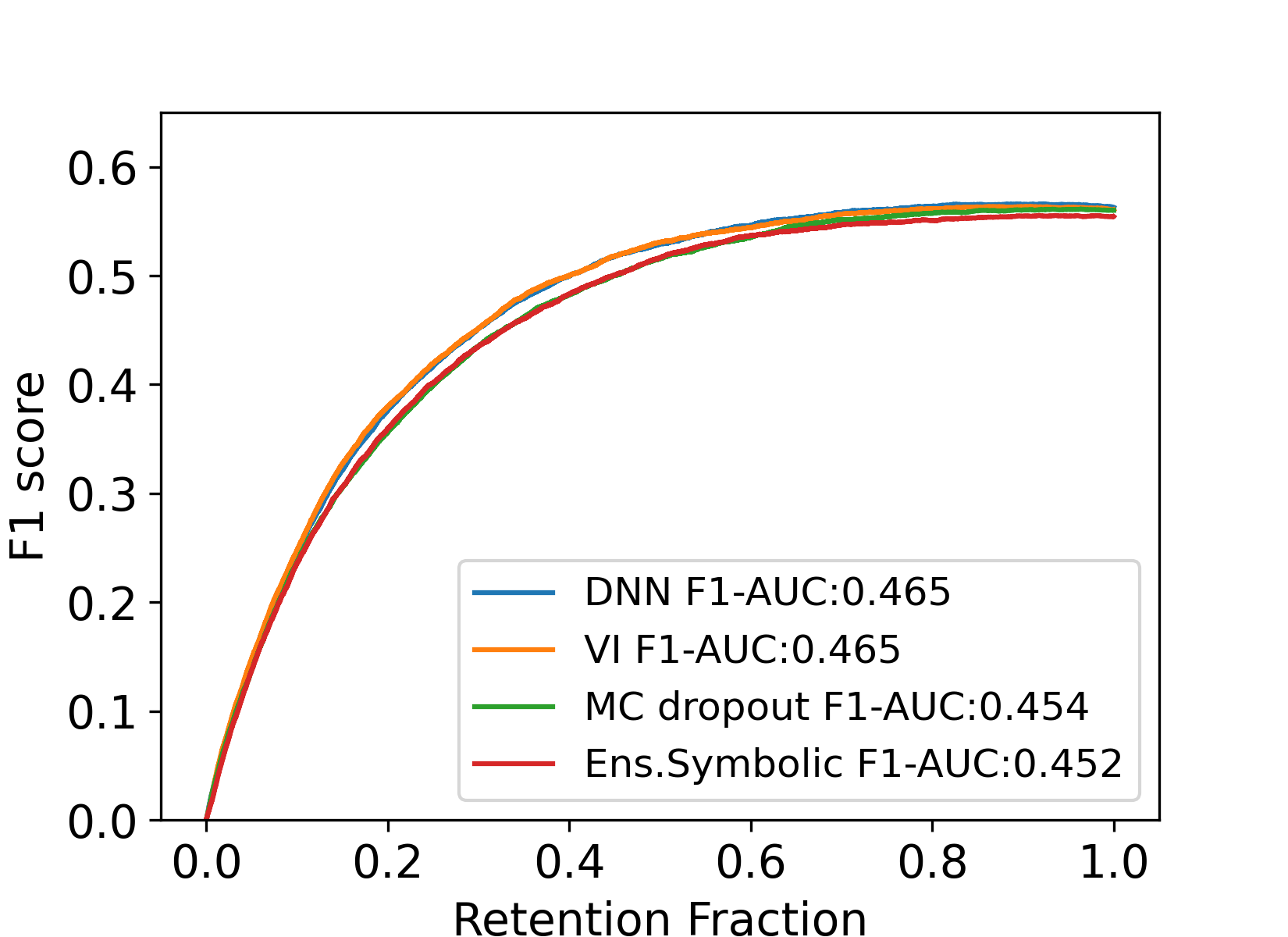}&
        \includegraphics[width=.32\linewidth]{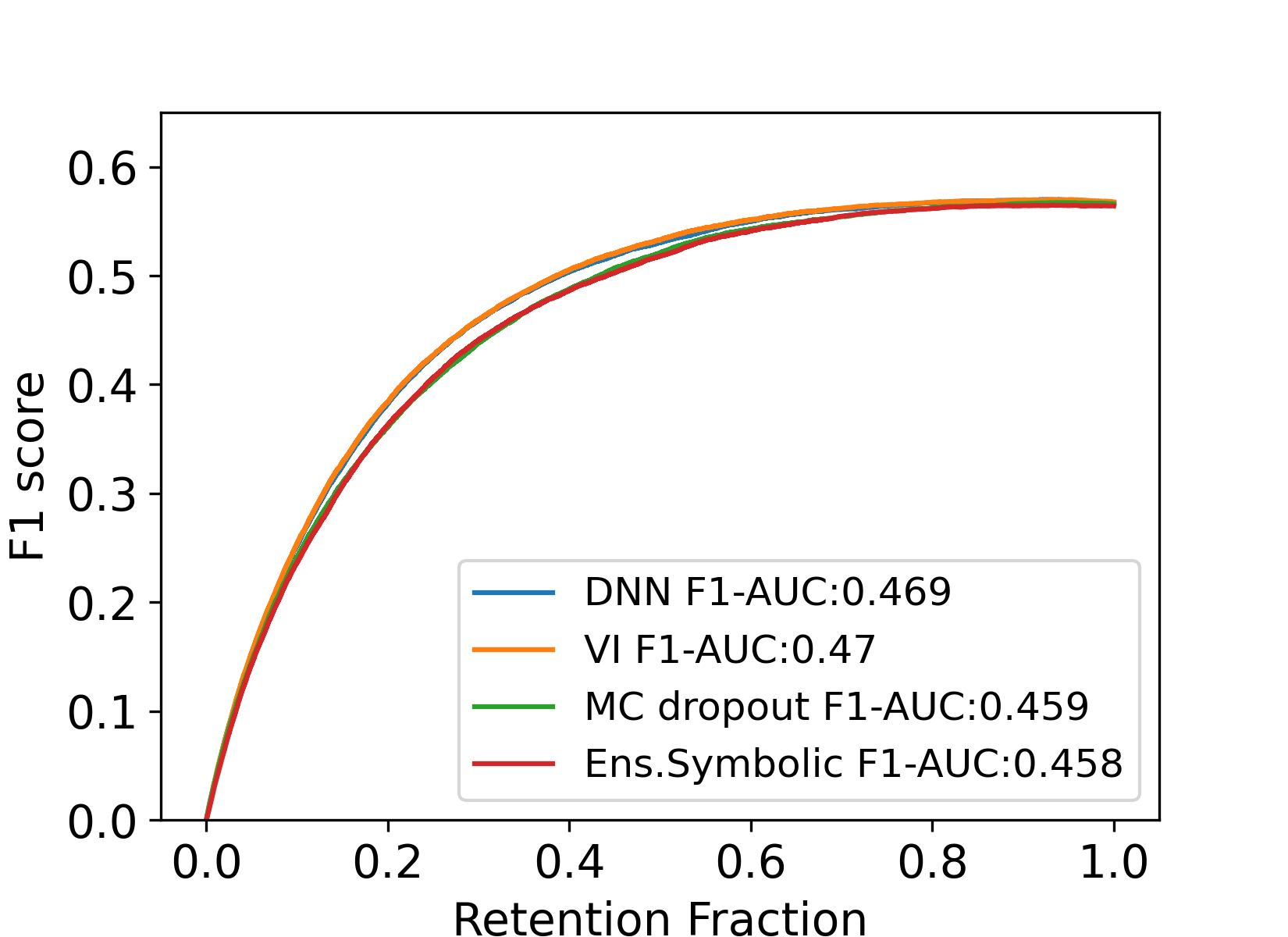}&
        \includegraphics[width=.32\linewidth]{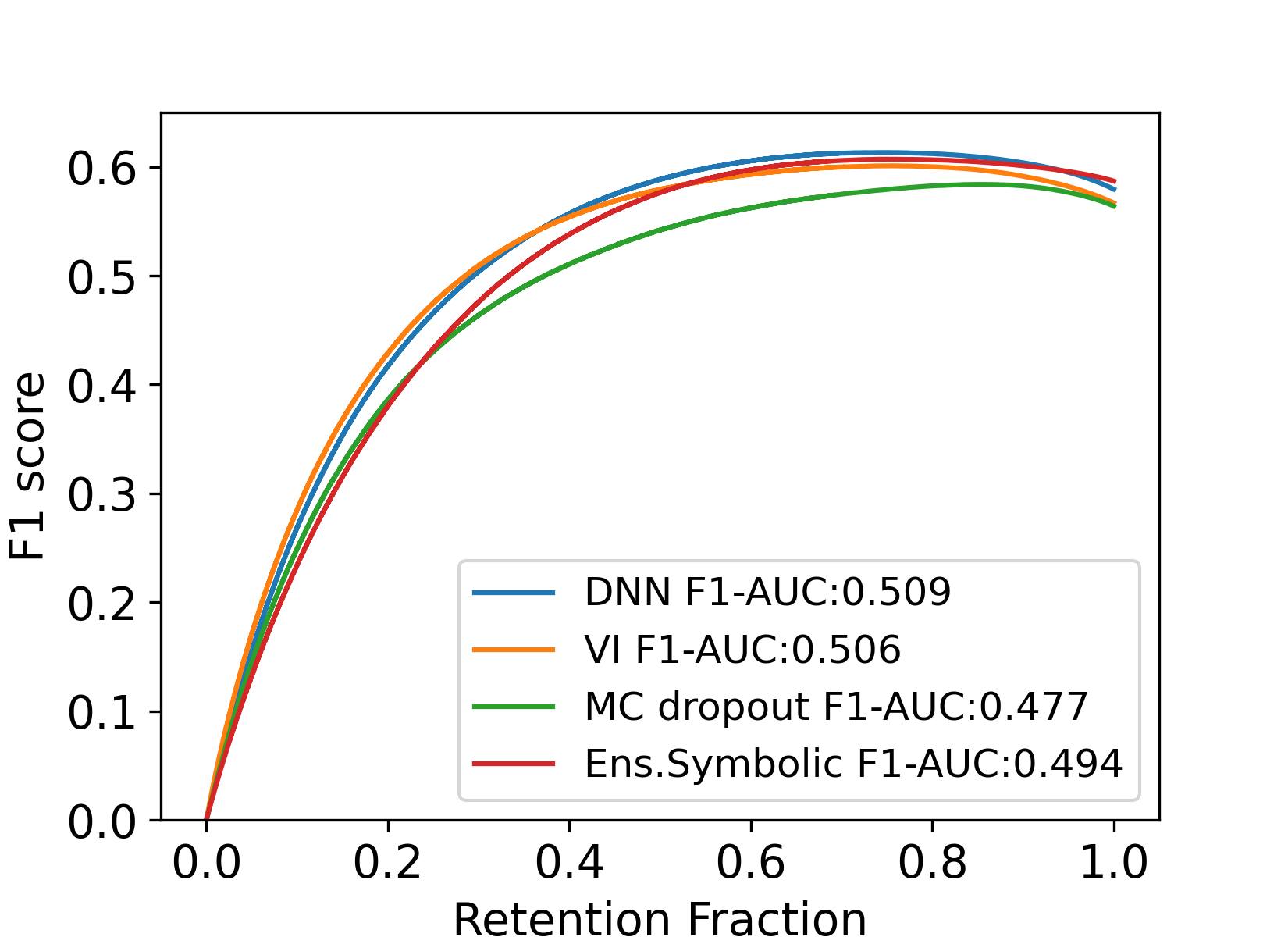}\\
    \end{tabular}
\caption{Retention curves for the synthetic development, evaluation and generalization sets. VI and MC dropout refer to the deep ensemble technique while DNN and Symbolic correspond to the ensemble setting.}
\label{fig:ds_retention_curves_synthetic}
\end{figure}

\paragraph{Real dataset} Due to a lack of knowledge of the actual data generation process, it is not possible to generate an analogous test set to the generalization set, which is only relevant in  the 'synthetic world.' As a result, model evaluation is limited to canonical partitions. Furthermore, we compare the results for the two datasets, namely synthetic and real ones, and we provide an interpretation of the observed behaviors.

The performance scores are shown in the Tables \ref{ds_errors_real} and \ref{ds_retention_real}. Deep Ensemble VI has the best performance across all metrics. This is not the case for the synthetic dataset at which Ensemble DNN has the best scores overall. This outcome is of no surprise, as both methods (along with Deep Ensemble MC dropout) are similar therefore deviations on their ranking are to be expected when working with different datasets.   

For the Ensemble Symbolic, it is worth noting that it has the lowest predictive performance in the dev and eval sets, which is consistent with the results in the synthetic data. Furthermore, when compared with the corresponding synthetic partitions, it is discovered to have a larger performance drop compared to the best model. This outcome is attributed to the limited expressivity of the Ensemble Symbolic, resulting in a more pronounced performance degradation in the real data. 
On the other hand, the synthetic generalization set revealed that the Ensemble Symbolic is the best candidate in terms of robustness across all possible operational conditions. Such domain constrained models have the advantage of unbiased performance across all possible conditions (operational or weather) making them good candidates for active performance optimization tasks (such as vessel-specific weather routing).

\begin{table}[!htp]
\centering
    \begin{small}
    \scalebox{0.75}{
    \setlength\tabcolsep{2pt}
        \begin{tabular}{l|l|l|lll|lll|lll}
            \toprule
            Dataset & Method & Model & \multicolumn{3}{c|}{RMSE (kW)} & \multicolumn{3}{c|}{MAE (kW)} & \multicolumn{3}{c}{MAPE (\%)} \\
            \multirow{7}{*}{ Dev } & & & In & Out & Full  & In & Out & Full & In & Out & Full \\
            \midrule

            & MC dropout & Deep ensemble & 1269 & 1501 & 1389 & 855 & 1067 & \textbf{961} & 5.42 & \textbf{7.39} & 6.40 \\
            & VI & Deep ensemble & \textbf{1264} & 1514 & 1395 & \textbf{848} & 1074 & \textbf{961} & \textbf{5.29} & 7.44 & \textbf{6.37} \\
            & DNN & Ensemble & 1285 & \textbf{1484} & \textbf{1388} & 868 & \textbf{1066} & 967 & 5.43 & 7.49 & 6.46 \\
            & Symbolic & Ensemble & 1405 & 1630 & 1522 & 971 & 1181 & 1076 & 6.41 & 8.98 & 7.69 \\
            & MC dropout & Ensemble & $1291_{\pm21}$ & $1540_{\pm56}$ & $1422_{\pm30}$ & $874_{\pm17}$ & $1098_{\pm39}$ & $986_{\pm17}$ & $5.63_{\pm0.22}$ & $7.80_{\pm0.29}$ & $6.72_{\pm0.14}$ \\
            & VI & Ensemble & $1276_{\pm12}$ & $1537_{\pm41}$ & $1413_{\pm24}$ & $858_{\pm9}$ & $1093_{\pm26}$ & $975_{\pm14}$ & $5.39_{\pm0.07}$ & $7.65_{\pm0.19}$ & $6.53_{\pm0.11}$ \\
            & DNN & Single & $1318_{\pm51}$ & $1547_{\pm63}$ & $1438_{\pm44}$ & $893_{\pm41}$ & $1113_{\pm46}$ & $1003_{\pm30}$ & $5.74_{\pm0.32}$ & $8.03_{\pm0.42}$ & $6.88_{\pm0.27}$ \\
            & Symbolic & Single & $1416_{\pm8}$ & $1654_{\pm53}$ & $1540_{\pm25}$ & $980_{\pm11}$ & $1199_{\pm54}$ & $1089_{\pm23}$ & $6.46_{\pm0.10}$ & $9.09_{\pm0.38}$ & $7.77_{\pm0.18}$ \\
            \midrule
            \multirow{4}{*}{ Eval } & MC dropout & Deep ensemble & 1248 & 1925 & 1622 & 850 & 1389 & 1119 & 5.54 & 8.29 & 6.91 \\
            & VI & Deep ensemble & \textbf{1243} & \textbf{1895} & \textbf{1602} & \textbf{842} & \textbf{1356} & \textbf{1098} & \textbf{5.38} & \textbf{8.09} & \textbf{6.73} \\
            & DNN & Ensemble & 1264 & 1928 & 1630 & 863 & 1414 & 1138 & 5.48 & 8.75 & 7.12 \\
            & Symbolic & Ensemble & 1393 & 2341 & 1926 & 964 & 1744 & 1354 & 6.37 & 10.66 & 8.52 \\
            & MC dropout & Ensemble & $1271_{\pm18}$ & $1954_{\pm47}$ & $1649_{\pm26}$ & $868_{\pm16}$ & $1416_{\pm39}$ & $1142_{\pm20}$ & $5.72_{\pm0.20}$ & $8.54_{\pm0.33}$ & $7.13_{\pm0.20}$ \\
            & VI & Ensemble & $1255_{\pm11}$ & $1916_{\pm32}$ & $1620_{\pm20}$ & $852_{\pm9}$ & $1377_{\pm39}$ & $1114_{\pm22}$ & $5.46_{\pm0.06}$ & $8.26_{\pm0.33}$ & $6.86_{\pm0.18}$ \\
            & DNN & Single & $1296_{\pm47}$ & $1985_{\pm111}$ & $1677_{\pm61}$ & $887_{\pm39}$ & $1462_{\pm91}$ & $1175_{\pm47}$ & $5.74_{\pm0.28}$ & $9.15_{\pm0.62}$ & $7.44_{\pm0.38}$ \\
            & Symbolic & Single & $1403_{\pm8}$ & $2366_{\pm304}$ & $1948_{\pm179}$ & $973_{\pm10}$ & $1770_{\pm253}$ & $1371_{\pm123}$ & $6.42_{\pm0.10}$ & $10.84_{\pm1.42}$ & $8.63_{\pm0.70}$ \\
            \midrule
        \end{tabular}}
    \end{small}
\caption{Predictive performance for the canonical partitions of the real dataset. One standard deviation is quoted for the single seed results.}
\label{ds_errors_real}
\end{table}

\begin{table}[!htp]
\centering
    \begin{small}
    \scalebox{0.7}{
    \setlength\tabcolsep{2pt}
        \begin{tabular}{l|l|l|lll|lll|lll}
            \toprule
            Dataset & Method & Model & \multicolumn{3}{c|}{R-AUC $* 10^{5}$} & \multicolumn{3}{c|}{F1-AUC} & \multicolumn{3}{c}{F1@95\%} \\
            \multirow{7}{*}{ Dev } & & & In & Out & Full & In & Out & Full & In & Out & Full \\
            \midrule
            & MC dropout & Deep ensemble & 4.52 & 7.21 & 6.05 & 0.510 & \textbf{0.469} & 0.486 & 0.618 & 0.536 & 0.577 \\
            & VI & Deep ensemble & \textbf{4.29} & 7.22 & \textbf{5.81} & \textbf{0.521} & 0.460 & \textbf{0.493} & \textbf{0.625} & \textbf{0.541} & \textbf{0.584} \\
            & DNN & Ensemble & 4.71 & \textbf{6.80} & 5.84 & 0.514 & 0.441 & 0.477 & 0.616 & 0.521 & 0.570 \\
            & Symbolic & Ensemble & 6.55 & 10.80 & 8.48 & 0.453 & 0.362 & 0.419 & 0.557 & 0.472 & 0.511 \\
            & MC dropout & Ensemble & $4.75_{\pm0.32}$ & $8.58_{\pm0.99}$ & $6.84_{\pm0.48}$ & $0.510_{\pm0.009}$ & $0.450_{\pm0.014}$ & $0.477_{\pm0.006}$ & $0.610_{\pm0.009}$ & $0.525_{\pm0.012}$ & $0.568_{\pm0.005}$ \\
            & VI & Ensemble & $4.40_{\pm0.17}$ & $7.85_{\pm0.52}$ & $6.18_{\pm0.24}$ & $0.518_{\pm0.007}$ & $0.448_{\pm0.012}$ & $0.484_{\pm0.004}$ & $0.620_{\pm0.005}$ & $0.535_{\pm0.009}$ & $0.579_{\pm0.003}$ \\
            & DNN & Single & $5.00_{\pm0.57}$ & $7.84_{\pm0.62}$ & $6.61_{\pm0.51}$ & $0.505_{\pm0.016}$ & $0.416_{\pm0.020}$ & $0.459_{\pm0.012}$ & $0.605_{\pm0.020}$ & $0.503_{\pm0.019}$ & $0.555_{\pm0.012}$ \\
            & Symbolic & Single & $6.99_{\pm0.22}$ & $11.33_{\pm1.14}$ & $8.99_{\pm0.53}$ & $0.449_{\pm0.006}$ & $0.355_{\pm0.019}$ & $0.412_{\pm0.009}$ & $0.552_{\pm0.010}$ & $0.465_{\pm0.019}$ & $0.509_{\pm0.009}$ \\ 
            \midrule
            \multirow{4}{*}{ Eval } & MC dropout & Deep ensemble & 4.34 & \textbf{13.59} & 9.28 & 0.513 & 0.394 & 0.451 & 0.621 & 0.459 & 0.544 \\
            & VI & Deep ensemble & \textbf{4.15} & 14.07 & \textbf{9.13} & \textbf{0.525} & \textbf{0.398} & \textbf{0.467} & \textbf{0.627} & \textbf{0.477} & \textbf{0.557} \\
            & DNN & Ensemble & 4.50 & 14.00 & 9.52 & 0.517 & 0.387 & 0.451 & 0.616 & 0.428 & 0.528 \\
            & Symbolic & Ensemble & 6.39 & 22.56 & 13.56 & 0.455 & 0.267 & 0.383 & 0.558 & 0.320 & 0.447 \\
            & MC dropout & Ensemble & $4.57_{\pm0.27}$ & $14.46_{\pm1.28}$ & $10.00_{\pm0.59}$ & $0.511_{\pm0.010}$ & $0.383_{\pm0.022}$ & $0.441_{\pm0.013}$ & $0.610_{\pm0.010}$ & $0.441_{\pm0.026}$ & $0.530_{\pm0.015}$ \\
            & VI & Ensemble & $4.26_{\pm0.16}$ & $14.58_{\pm1.02}$ & $9.57_{\pm0.56}$ & $0.521_{\pm0.007}$ & $0.383_{\pm0.017}$ & $0.455_{\pm0.008}$ & $0.621_{\pm0.006}$ & $0.467_{\pm0.027}$ & $0.549_{\pm0.014}$ \\
            & DNN & Single & $4.78_{\pm0.52}$ & $15.68_{\pm2.16}$ & $10.97_{\pm1.21}$ & $0.506_{\pm0.017}$ & $0.364_{\pm0.020}$ & $0.425_{\pm0.024}$ & $0.603_{\pm0.021}$ & $0.416_{\pm0.033}$ & $0.515_{\pm0.023}$ \\
            & Symbolic & Single & $6.81_{\pm0.23}$ & $27.85_{\pm7.33}$ & $17.51_{\pm4.27}$ & $0.449_{\pm0.006}$ & $0.254_{\pm0.037}$ & $0.360_{\pm0.027}$ & $0.553_{\pm0.009}$ & $0.322_{\pm0.068}$ & $0.447_{\pm0.028}$ \\
            \midrule
        \end{tabular}}
    \end{small}
\caption{Retention performance for the canonical partitions of the real dataset. One standard deviation is quoted for the single seed results.}
\label{ds_retention_real}
\end{table}

\begin{figure}[!htp]
\centering
    \begin{tabular}{c@{ }c@{ }}
        Development & Evaluation \\
        \includegraphics[width=.32\linewidth]{figures/error_retention_all_models_dev.png}&
        \includegraphics[width=.32\linewidth]{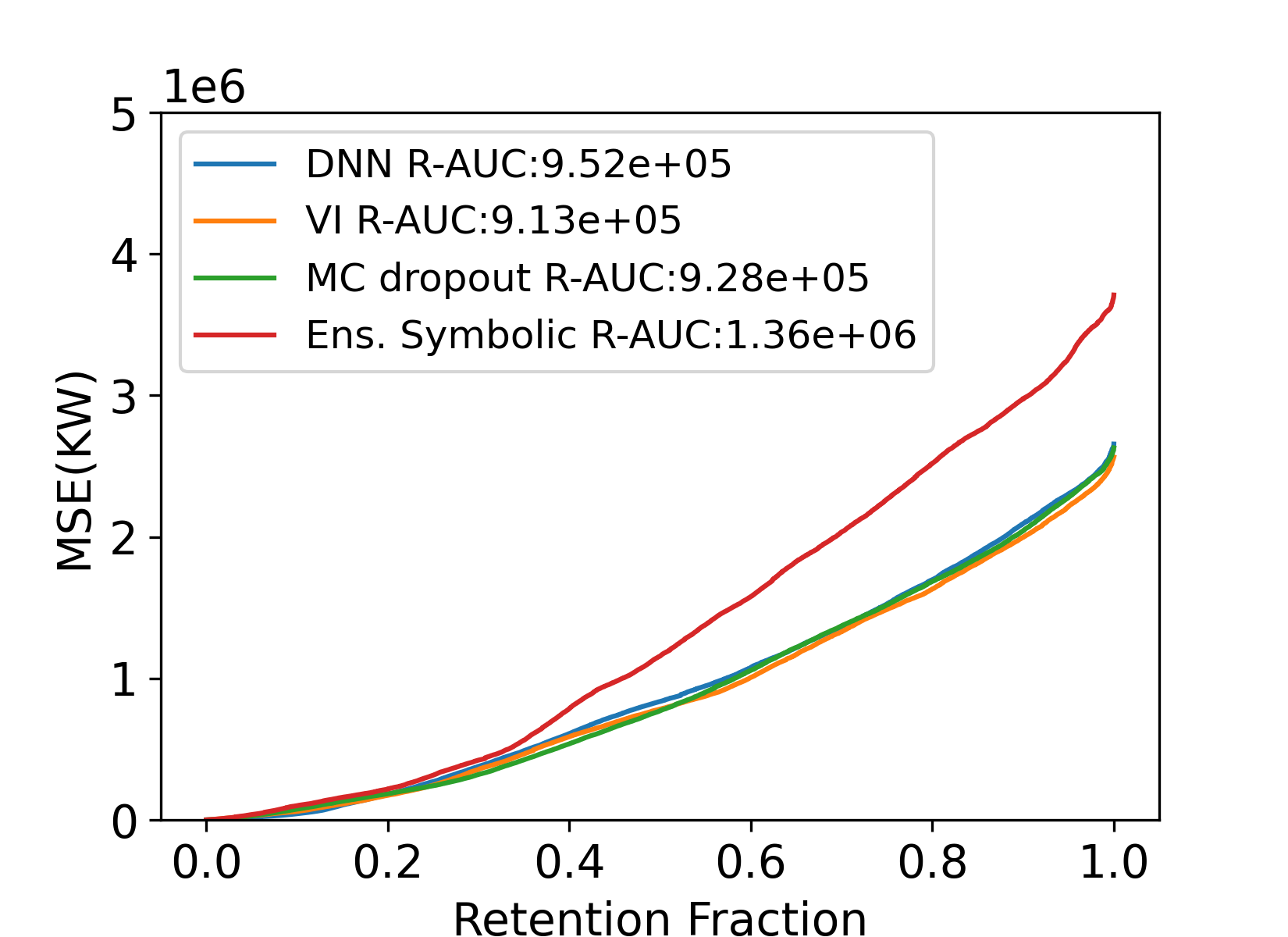}\\
        \includegraphics[width=.32\linewidth]{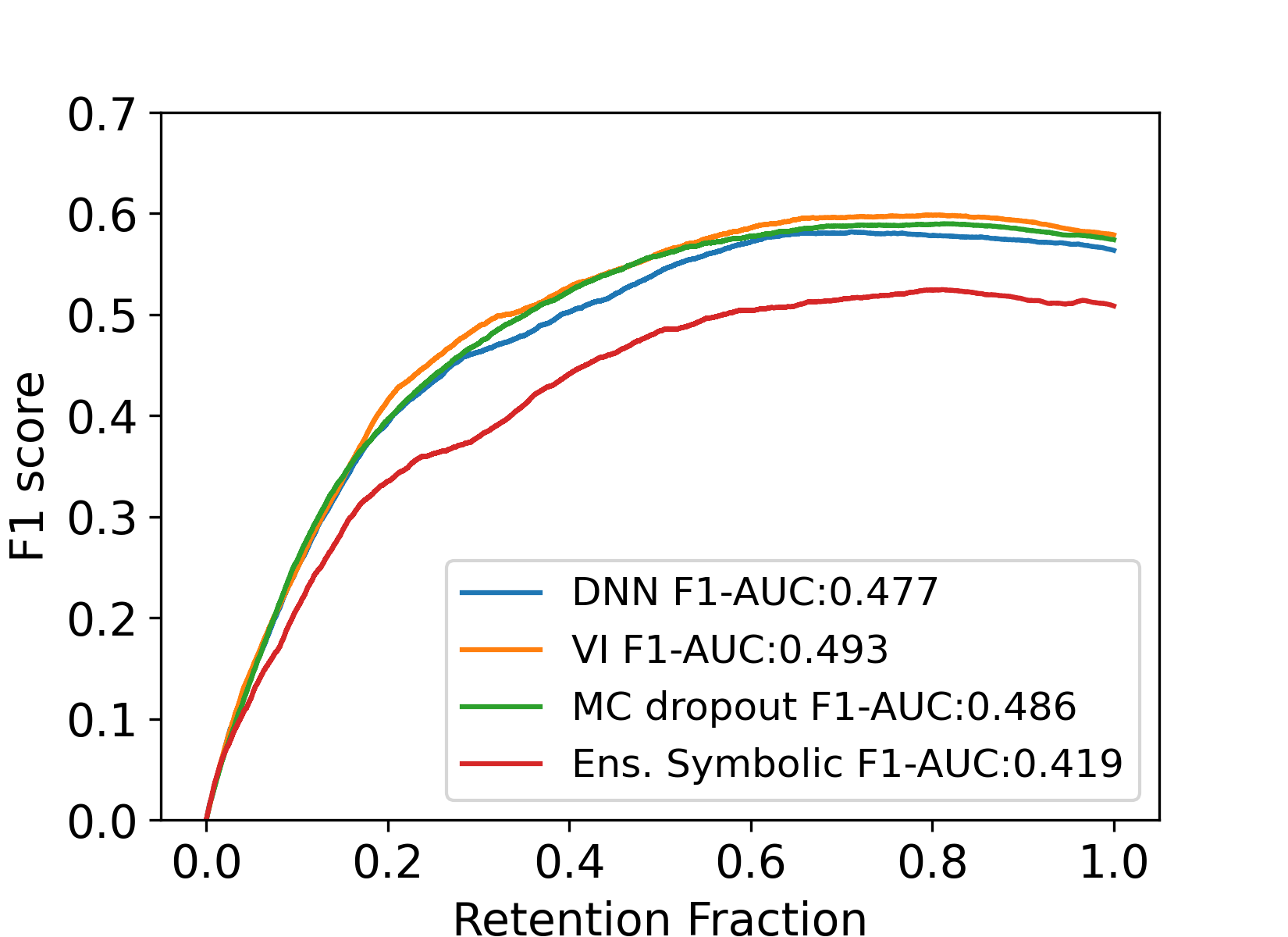}&
        \includegraphics[width=.32\linewidth]{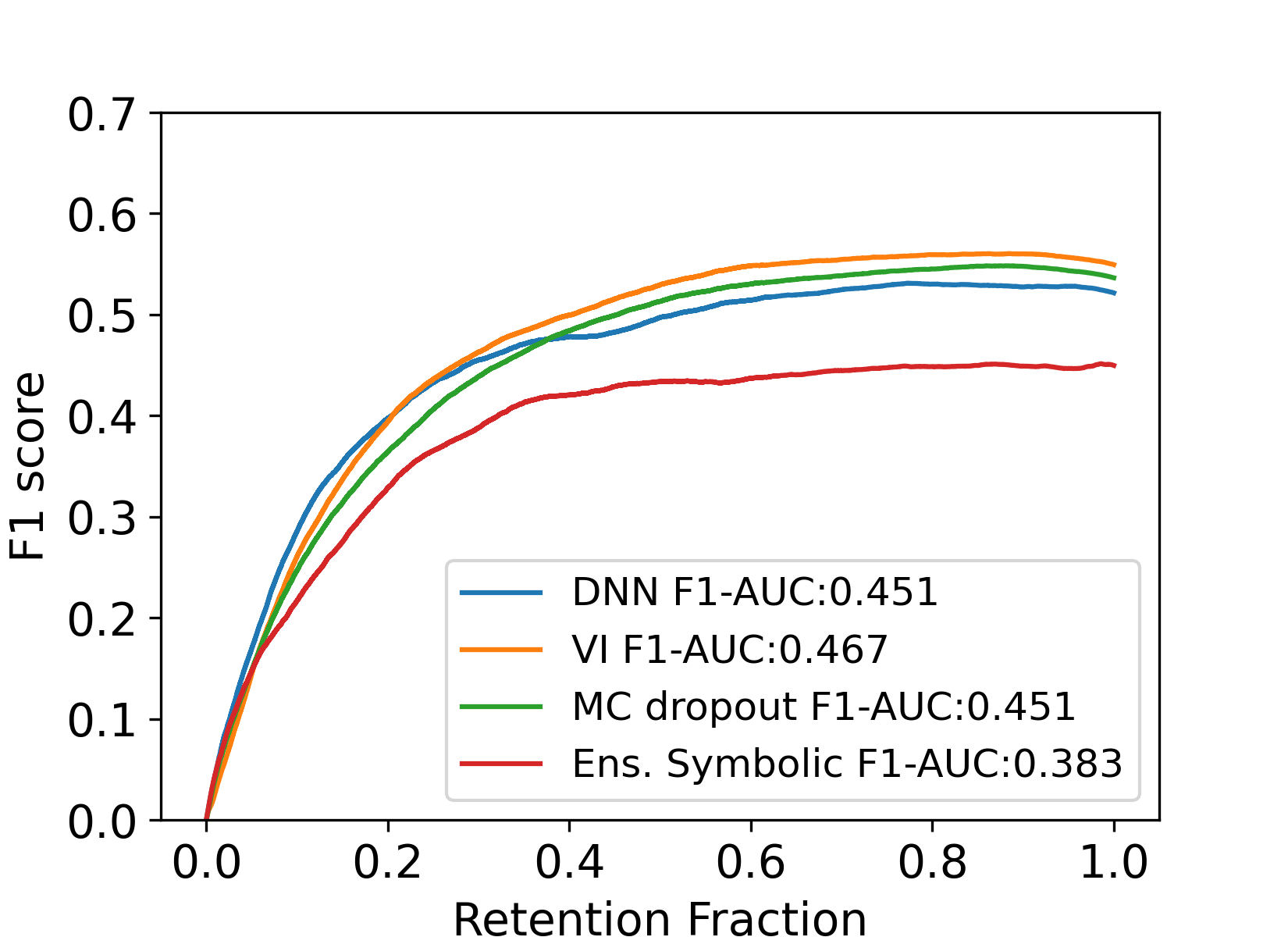}\\
    \end{tabular}
\caption{Retention curves for the real development, evaluation sets. VI and MC dropout refer to the deep ensemble technique while DNN and Symbolic correspond to the ensemble setting.}
\label{fig:ds_retention_curves_real}
\end{figure}

%\section{Rules}\label{apn:rules}

\end{document}